\documentclass{article}

\PassOptionsToPackage{numbers, compress}{natbib}

 \usepackage[main, final]{neurips_2026}

\usepackage[utf8]{inputenc} 
\usepackage[T1]{fontenc}    
\usepackage{hyperref}       
\usepackage{url}            
\usepackage{booktabs}       
\usepackage{amsfonts}       
\usepackage{nicefrac}       
\usepackage{microtype}      
\usepackage{xcolor}         
\usepackage{graphicx}
\usepackage{subcaption}
\usepackage{array}
\usepackage{amsmath}
\usepackage{amssymb}
\usepackage{mathtools}
\usepackage{amsthm}
\usepackage{algorithm}
\usepackage{algorithmic}
\usepackage[table]{xcolor}
\usepackage{colortbl}
\usepackage{wrapfig}
\usepackage{makecell}

\hypersetup{
    colorlinks=true,
    linkcolor=black!65,
    citecolor=black!65,
    urlcolor=blue!50!black,
    pdftitle={GLOVE: Global Verifier for LLM Memory-Environment Realignment},
    pdfauthor={Xingkun Yin, Hongyang Du},
    pdfsubject={Conference on Neural Information Processing Systems (NeurIPS) 2026},
    pdfkeywords={LLM agents, long-term memory, environmental drift, memory verification, active probing}
}

\usepackage{fancyhdr}
\fancypagestyle{nicelogo}{%
  \fancyhf{}%
  \fancyhead[C]{\includegraphics[width=4.2in]{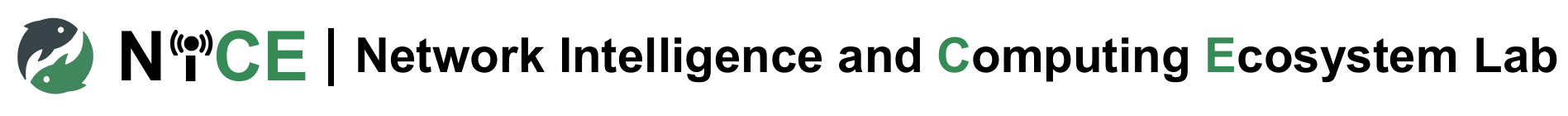}}%
}
\fancypagestyle{plain}{%
  \fancyhf{}%
  \fancyfoot[C]{\thepage}%
}

\title{GLOVE: Global Verifier for LLM Memory-Environment Realignment}

\author{%
  Xingkun Yin, Hongyang Du \\
  Department of Electrical and Computer Engineering\\
  The University of Hong Kong\\
  Hong Kong SAR, China\\
  \texttt{\{yinxingkun@connect., duyu@\}hku.hk} \\
}

\newcommand{\improv}[1]{\textcolor{cyan!60!green}{\scriptsize($\uparrow$#1)}}
\newcommand{\decline}[1]{\textcolor{orange}{\scriptsize($\downarrow$#1)}}

\theoremstyle{plain}
\newtheorem{theorem}{Theorem}[section]

\theoremstyle{definition}

\theoremstyle{remark}

\begin{document}
\setlength{\headheight}{28pt}
\addtolength{\topmargin}{-6pt}
\addtolength{\headsep}{-10pt}

\maketitle
\thispagestyle{nicelogo}   

\begin{abstract}
Most existing memory-enhanced Large Language Model (LLM) approaches implicitly assume that memory validity can be established either through external evaluators that provide task-specific success signals or through internal model cognition, such as reflection, for editing memory entries. However, these assumptions often break down in practical environments with dynamic drifts. We propose the \underline{Glo}bal \underline{Ve}rifier (GLOVE), a plug-and-play module for LLM memory systems that establishes a relative notion of truth to achieve memory-environment realignment in the face of environmental drifts.
Through active probing to detect inconsistencies between retrieved memories and fresh observations, GLOVE enables memory-environment realignment by verifying and updating memory without access to task-specific ground-truth supervision or strong reliance on model introspection. 
We evaluate GLOVE across a spectrum of tasks ranging from web navigation and discrete planning to continuous and embodied control, covering both simulated benchmarks and real-world robotic deployment.
Across all domains, GLOVE improves adaptation across various LLM memory designs under both explicit and implicit drift, often recovering performance from near-failure to high success rates.
These results suggest a practical pathway to memory-environment realignment for self-evolving cognitive agents.

\par\smallskip
\noindent\textbf{Code and Experiment Settings:}~\url{https://github.com/NICE-HKU/GLOVE}.
\end{abstract}

\section{Introduction}
Large Language Models (LLMs) have evolved from static conversational interfaces to autonomous agents that execute long-horizon tasks, including web navigation and embodied control~\cite{yao2022react, driess2023palme}. This shift toward general-purpose autonomy increases the importance of experience accumulation, with recent evidence suggesting that future capability gains will rely on agents learning from their own interaction histories~\cite{silver2025era}. External memory is therefore becoming a central component of LLM agents, as it allows them to reuse past observations, action traces, and task outcomes across long time horizons. However, this paradigm also introduces a fundamental weakness, as stored experience is often treated as reliable once it has been written into memory. Without continuous validation against the current environment, memory can grow into a mixture of useful experience, outdated facts, and internally plausible but invalid assumptions. This problem becomes especially critical in dynamic settings, where an experience may be correct when it is stored but false when it is later retrieved. For example, a web interface may change its button layout, a navigation route may become blocked, or a user preference may shift over time. In these cases, stale memory can persist even after it no longer reflects the current environment, causing the agent to reuse obsolete decisions with high confidence.

Existing LLM memory systems provide useful mechanisms for storing, organizing, and retrieving experience, but they do not fully solve this deployment-time validation problem.
We characterize current approaches to maintaining memory validity along two broad axes.
The first axis is memory-internal adaptation.
MemoryBank uses passive decay~\cite{zhong2024memorybank}, while Generative Agents synthesize high-level insights through reflection~\cite{park2023generative}.
Another line reduces the cognitive burden on the LLM by reorganizing memory into structured forms, such as graph-based or hierarchical representations, as in MemGPT and G-Memory~\cite{packer2023memgpt, zhang2025gmemory}.
These approaches improve memory usability, retrieval quality, and reasoning coherence.
However, they still rely on the agent to judge memory validity mainly from internal consistency.
This is insufficient under environmental drift, because a stale memory can be logically consistent with prior experience while being inconsistent with the current environment.
Passive decay, reflection, or organization may make such memories easier to forget, retrieve, or justify, but they do not by themselves test whether the stored experience remains valid.
The second axis is feedback-based validation, where memory or behavior is checked using explicit rewards, verifiers, task-success feedback, or other external signals~\cite{shinn2023reflexion, cobbe2021training}. Representative examples include Memory-R1 and Voyager with an evaluator~\cite{yan2025memory, wang2024voyager}. Such feedback can be effective when it is dense, timely, and aligned with the memory entries that need to be edited. In open-ended deployments, however, supervision is often sparse, delayed, or absent. Even when partial feedback exists, it usually evaluates terminal task outcomes rather than the intermediate state-action experiences that populate memory. This creates a localization gap. For example, in navigation, a previously safe route may become blocked due to environmental changes~\cite{ditzler2015learning}. A reflection-based agent may repeatedly justify a plan using an obsolete map stored in memory, while a feedback-based approach may only observe failure at the end of a long trajectory. In both cases, the agent lacks a direct mechanism for identifying which intermediate memory entries became invalid. As a result, it can retrieve and reuse stale steps indiscriminately, and may treat hallucinated or outdated premises as reliable evidence~\cite{huanglarge}.

Dynamic environments further expose this gap by changing the response pattern associated with a familiar context. Under such changes, the same retrieved memory can lead to different consequences at different deployment times. The key issue is therefore not merely whether an agent can store useful experience, but whether it can detect when live interaction contradicts retrieved experience and then realign its memory store before the stale experience continues to affect future decisions. A practical memory system should treat memory as an editable hypothesis about current environment behavior, rather than as a static record of past truth. This requires an online realignment mechanism that monitors inconsistencies between stored experience and fresh interaction evidence, suppresses or revises contradicted memories, and updates the memory store as new evidence accumulates.
To address this issue, we introduce a relative-truth-based approach for memory-environment realignment. Relative truth refers to the current validity of memory as supported by observable interaction evidence, including immediate state transitions, reward or score changes, user responses, and delayed task feedback. This formulation covers both explicit drift, where misalignment appears through changed state-action-outcome transitions, and implicit drift, where the visible transition structure may remain similar while downstream feedback changes. Thus, memory validity is determined by its consistency with current interaction evidence, rather than by internal coherence alone.

We instantiate this design with \underline{Glo}bal \underline{Ve}rifier~(GLOVE), a plug-and-play module that augments existing memory systems without modifying their underlying architecture. GLOVE monitors retrieved memory during online interaction, detects persistent inconsistencies between stored experience and fresh evidence, and realigns the memory store by removing obsolete experience and incorporating verified outcomes. This design enables memory systems to maintain alignment with dynamic environments while remaining compatible with a wide range of existing agent frameworks.
Our contributions are summarized as follows:
\begin{itemize}
  \item \textbf{Memory-Environment Misalignment Analysis.} We formalize memory-environment misalignment in dynamic environments, identifying an ``epistemic gap'' in which LLM agents rely on static historical experience despite unobserved changes in environment dynamics and the absence of immediate ground-truth feedback.
  \item \textbf{Technical Solution.} We propose \textbf{GLOVE}, 
  a plug-and-play memory-validation module that closes this epistemic gap through active probing. GLOVE detects cognitive dissonance between retrieved memories and fresh interaction evidence, constructs relative truth from observable consequences of current interaction, and edits the memory store by replacing obsolete experience.
  \item \textbf{Empirical Validation.} We evaluate GLOVE across web navigation, discrete planning, continuous control, and real-world robot navigation under both explicit and implicit environmental drift. Experiments across diverse memory architectures and LLM backbones show that GLOVE improves adaptation after drift, remains effective when exact replay is unavailable, and concentrates interaction overhead around drift events.
\end{itemize}

\section{Related Works}

\textbf{Long-Term Agentic Memory.}
Recent LLM agents have improved long-term memory through scalable storage, structured retrieval, and reflective summarization, including external memory expansion~\cite{wangmplus}, graph- or hierarchy-based organization~\cite{lightrag,zhang2025gmemory}, and reflection-based consolidation of episodic experience~\cite{park2023generative,shinn2023reflexion,zhang2025agent}. 
However, most of these systems still operate in a RAG-style regime that treats retrieved memory as valid once stored~\cite{lewis2020retrieval}. Under environmental drift, this assumption can turn stale experience into a persistent source of error. GLOVE targets this gap by introducing explicit verification between retrieved memory and current environment feedback.


\textbf{Self-Evolving Agent and Environment Adaptation.}
A related line of work studies self-evolving agents through introspective self-correction and interaction-driven adaptation. Semantic critics and self-refinement loops can improve reasoning when reliable feedback or reference criteria are available~\cite{shinn2023reflexion,NEURIPS2023_91edff07,lan2024criticeval}. Other approaches interleave reasoning and acting or improve behavior from ongoing interaction~\cite{yao2022react,wang2024voyager}, while lifelong adaptation methods address changing environments mainly through failure feedback or stationary-update assumptions~\cite{gama2014survey,padakandla2021survey,wen2025real}. Yet these methods typically update behavior additively and do not explicitly verify whether previously stored memories remain valid after drift. GLOVE focuses on this missing step by detecting memory-environment misalignment and actively realigning stale memory.

\section{Problem Setup and Global Verifier}
\label{gv}
\subsection{Memory-Environment Misalignment Objective}
We consider a memory-augmented LLM agent interacting with a dynamic environment over discrete time steps. At time $t$, the agent receives a raw observation $o_t$ and maps it to a task-relevant state representation $s_t=\phi(o_t)$, where $s_t\in\mathcal{S}$ and $\phi$ is an environment-specific state formation procedure. Examples are provided in Appendix~\ref{app:state_formation}. Given $s_t$, the agent retrieves a memory context $C_t$ from an external experience bank $\mathcal{D}_t$ and selects an action $a_t\in\mathcal{A}$. After executing $a_t$, the environment returns an observable outcome $s'_t$. 
We characterize the local behavior of the environment at time $t$ by a response distribution $\mathcal{Q}_t(\cdot\mid s,a)$ over observable outcomes for a decision context $(s,a)$. The agent stores past interaction records in an experience bank
\begin{equation}
    \mathcal{D}_t=\biguplus_{i=1}^{N_t}\{e_i\}, \qquad e_i=(s_i,a_i,s'_i),
\end{equation}
where $\mathcal{D}_t$ is an indexed multiset of records collected before time $t$, duplicate outcomes are preserved as separate samples, while auxiliary metadata is omitted for clarity.

An \emph{environmental drift} is relevant to memory validation when the response distribution that generated a stored experience at \(e_i\)'s collection time \(t_i<t\) differs from the current one:
\begin{equation}
    \exists e_i=(s_i,a_i,s'_i)\in\mathcal{D}_t
    \quad \mathrm{s.t.}\quad
    \mathcal{Q}_{t_i}(\cdot\mid s_i,a_i)\neq
    \mathcal{Q}_{t}(\cdot\mid s_i,a_i),
    \label{eq:drift_def}
\end{equation}
This definition covers both explicit and implicit drifts. In explicit drift, the observable transition outcome changes, such as after an interface update, topology change, or dynamics shift. In implicit drift, the visible transition may remain similar while the downstream consequence changes, e.g., through reward reversal, score change, or shifted user preference. Thus, a memory entry can become obsolete either because the environment now produces a different outcome or because the same apparent outcome has a different task consequence.

We define \emph{memory-environment misalignment} as the resulting discrepancy between the response pattern implied by retrieved memory and the current environment response. For a decision context $(s,a)$, let $\widehat{\mathcal{Q}}_{\mathcal{D}_t}(\cdot\mid s,a)$ denote the empirical outcome distribution induced by relevant entries in the current experience bank. The objective of \emph{Global Verification} is to realign the memory bank when current interaction evidence indicates that this memory-induced distribution no longer matches the environment:
\begin{equation}
    \mathcal{D}_{t+1}
    =
    \mathrm{Realign}(\mathcal{D}_t,\mathcal{E}_t),
    \qquad
    \widehat{\mathcal{Q}}_{\mathcal{D}_{t+1}}(\cdot\mid s,a)
    \approx
    \mathcal{Q}_{t}(\cdot\mid s,a),
    \label{eq:global_verification_obj}
\end{equation}
where $\mathcal{E}_t$ denotes fresh evidence collected from current or future interactions. The challenge is to decide when such realignment is needed and how much evidence is sufficient to revise memory reliably. GLOVE addresses these two questions through cognitive dissonance detection and relative truth construction, described next.

\subsection{GLOVE Framework Overview}
\label{gv-gloveFrameworkOverview}
\begin{figure*}[t]
    \centering
    \includegraphics[width=\textwidth]{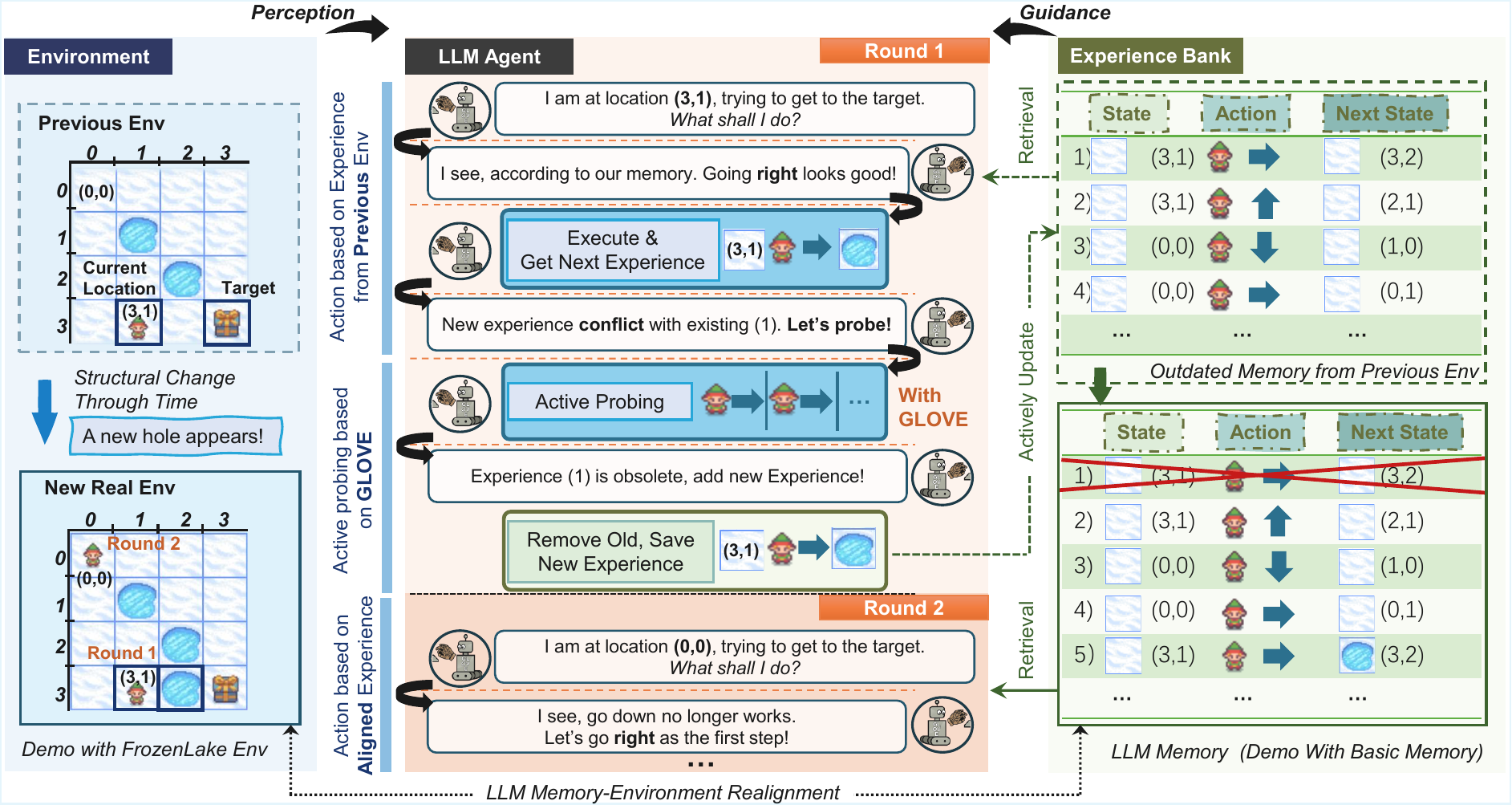}
    \caption{The overview of the GLOVE-augmented LLM workflow.}
    \label{fig:framework}
\end{figure*}
Unlike retrieval-augmented agents that treat retrieved experience as fixed ground truth~\cite{packer2023memgpt}, GLOVE treats memory as a \textbf{hypothesis} about environment behavior and verifies it through interaction. As illustrated in Fig.~\ref{fig:framework}, GLOVE is integrated into the agent--environment loop. At each step, the agent retrieves relevant experiences from the experience bank $\mathcal{D}$ to guide action selection. After the action is executed, GLOVE compares the observed outcome with stored outcomes associated with the same state-action precondition. When a discrepancy is detected, GLOVE triggers an \textbf{active verification} procedure. In \emph{synchronous verification}, the agent re-executes the queried action under the same state context to collect fresh outcomes immediately. When exact replay is unavailable, GLOVE uses \emph{replay-relaxed asynchronous verification}: it suppresses the contradicted memory from future retrieval, marks the queried precondition as pending verification, and accumulates evidence when counterpart states are re-encountered through natural interaction. These two modes differ operationally, but both construct relative truth from observable interaction evidence, without relying on memory-validity labels or internal introspection. Algorithm~\ref{alg:glove} summarizes the complete procedure, with the three phases detailed below.

\begin{algorithm}[t]
\caption{Memory-enhanced LLM Agent with GLOVE}
\label{alg:glove}
\begin{algorithmic}[1]
\item[] {\bfseries Input:} Task $t$, LLM agent $\pi$, environment $\mathit{Env}$, experience bank $\mathcal D$, probing budget $\alpha$, persistence threshold $p_{\rm th}$
\item[] {\bfseries Output:} Task trajectory $\tau$, updated experience bank $\mathcal D$
\STATE Initialize pending verification queue $\mathcal P \leftarrow \emptyset$ and track consecutive surprise counts for each $(s,a)$

\item[] \textbf{\# Main Execution Loop}
\WHILE{Task $t$ not finished}
\STATE Observe current state $s_t$ and retrieve related information $C_t$ from $\mathcal D$
\STATE Form action \(a_t \sim \mathrm{LLM}(s_t, C_t)\)
\STATE Execute $a_t$ in $\mathit{Env}$, observe outcome $s'_t$
\STATE Record transition $e_t$, $(s_t, a_t,  s_t')$, in trajectory $\tau$
\item[] \textbf{\# GLOVE Process}
\item[] \textit{\# Phase I: Cognitive Dissonance Detection}
\STATE Retrieve local counterpart experiences
\(\mathcal{N} \) according to \eqref{eqcollect}
\IF{$\mathcal N \neq \emptyset$ \textbf{and} $\Phi_{\mathrm{surp}}(e_t)$ holds for $p_{\rm th}$ consecutive encounters of $(s_t,a_t)$}
    \item[] \textit{\# Phase II: Relative Truth Formation}
    \IF{sync mode}
        \STATE Recollect $(s_t,a_t)$ data points for $\alpha$ trials
        \STATE Collect fresh outcomes $\mathcal V=\{s'_{t,1},\ldots,s'_{t,\alpha}\}$
        \STATE Construct verified transition summary $\hat{Q}_{t}(\cdot \mid s_t,a_t)$ from $\mathcal V$
        \item[] \textit{\# Phase III: Memory--Environment Realignment}
        \STATE Remove obsolete counterparts: $\mathcal D \leftarrow \mathcal D \setminus \mathcal N$
        \STATE {Insert verified outcomes: $\mathcal D \leftarrow \mathcal D \cup \{(s_t,a_t,s') \mid s' \in (\hat{Q}_{t}(\cdot \mid s_t,a_t))\}$}
    \ELSE
        \STATE {Suppress contradicted counterparts and add $e_t$ to pending verification queue $\mathcal P$}
    \ENDIF
\ENDIF
\ENDWHILE
\STATE \textbf{Return} $\tau,\mathcal D$
\end{algorithmic}
\end{algorithm}

\subsection{When to Realign: Cognitive Dissonance Detection}
\label{gv-activeProbingAndRelativeTruth}
Cognitive dissonance occurs when a newly observed outcome is inconsistent with historical experience under the same effective state-action precondition. Since the environment response $\mathcal{Q}_t(\cdot\mid s,a)$ may be stochastic, a single unseen outcome does not necessarily imply environmental drift. GLOVE therefore detects dissonance through distributional consistency. Let the current interaction at time $t$ be $e_t=(s_t,a_t,s'_t)$. GLOVE retrieves counterpart experiences by matching the current precondition $(s_t,a_t)$ against historical records in the experience bank:
\begin{equation}\label{eqcollect}
\mathcal{N}(e_t) = \{ e_k \in \mathcal{D} \mid s_k \sim s_t \land a_k = a_t \},
\end{equation}
where $s_k \sim s_t$ denotes that two state representations are considered equivalent under a task-dependent matching rule. This relation may be exact in deterministic settings or approximate in open-ended environments, such as nearest-neighbor matching in a representation space. It only defines local comparability for verification and does not constrain the memory format used for action generation. Environment-specific examples are provided in Appendix~\ref{app:state_formation}. The set $\mathcal{N}(e_t)$ can be viewed as historical samples from the response distribution under the same effective precondition $(s_t,a_t)$. If $\mathcal{N}(e_t)=\emptyset$, the transition is treated as novel exploration. Otherwise, GLOVE constructs an empirical distribution $\hat{\mathcal{Q}}_{\mathrm{hist}}(\cdot\mid s_t,a_t)$, where $\hat{\mathcal{Q}}_{\mathrm{hist}}(s')$ is the empirical frequency of outcome $s'$ among experiences in $\mathcal{N}(e_t)$.
Cognitive dissonance is detected when the new outcome is unlikely under this distribution. We define the \textit{surprise predicate} as
\begin{equation}
\label{eq:surprise}
\Phi_{\mathrm{surp}}(e_t) \iff 
\hat{\mathcal{Q}}_{\mathrm{hist}}(s'_t \mid s_t,a_t) < \epsilon,
\end{equation}
where $\epsilon$ controls the tolerated probability mass of rare outcomes. Intuitively, $\Phi_{\mathrm{surp}}$ triggers when the observed transition falls outside the typical behavior of the environment under the same precondition.
To reduce sensitivity to transient noise, GLOVE maintains a counter for each state-action pair and initiates verification when $\Phi_{\mathrm{surp}}$ holds for $p_{\rm th}$ consecutive encounters of the same $(s_t,a_t)$. A non-surprising observation for that pair resets the counter.

\textbf{Detection Guarantee.}
Assume the environment remains locally stationary over a time window and the agent repeatedly encounters the same $(s,a)$. The historical outcomes can be treated as independent samples from a fixed response distribution $\mathcal{Q}_t(\cdot\mid s,a)$, and $\hat{\mathcal{Q}}_{\mathrm{hist}}(\cdot\mid s,a)$ estimates the typical outcomes under this precondition. We bound the finite-sample deviation of this estimate as:
\begin{theorem}[Finite-Sample Detection Bound]
\label{thm:detection}
For any outcome $s'$ and confidence level $\delta \in (0,1)$, with probability at least $1-\delta$,
\begin{equation}
\label{eq:detection_bound}
\left| \hat{\mathcal{Q}}_{\mathrm{hist}}(s' \mid s,a) - \mathcal{Q}_t(s' \mid s,a) \right|
\le \sqrt{\frac{\ln(2/\delta)}{2n}}.
\end{equation}
\end{theorem}


\textbf{Interpretation.}
Theorem~\ref{thm:detection} characterizes the finite-sample uncertainty of the empirical response estimate under locally stationary dynamics. Let $r_n(\delta)=\sqrt{\ln(2/\delta)/(2n)}$. For any outcome $s'$ whose true stationary probability satisfies $\mathcal{Q}_t(s'\mid s,a) \ge \epsilon + r_n(\delta)$, the empirical estimate can fall below the surprise threshold $\epsilon$ only when the estimation error exceeds the concentration margin. Thus, such a typical outcome is falsely marked as surprising with probability at most $\delta$. The persistence threshold $p_{\rm th}$ further reduces transient false alarms by requiring repeated surprise events under the same $(s,a)$ before verification is triggered, trading detection sensitivity for robustness to noise. The proof is provided in Appendix~\ref{app:proof_thm_4_1}.

\textbf{Deterministic Special Case.}
If the environment is deterministic, $\mathcal{Q}_t(\cdot\mid s,a)$ is supported on a single outcome $s^\star$, with $\mathcal{Q}_t(s^\star\mid s,a)=1$.
Under the local stationarity assumption, once at least one historical counterpart sample is observed, $\hat{\mathcal{Q}}_{\mathrm{hist}}(\cdot\mid s,a)$ also concentrates on $s^\star$, so the actual empirical deviation is zero.
The distribution-free Hoeffding bound in~\eqref{eq:detection_bound} remains positive for finite $n$, but the surprise predicate in~\eqref{eq:surprise} becomes an exact check of whether the newly observed outcome $s'_t$ equals $s^\star$.
Thus, exact-match detection is a deterministic special case of the distributional criterion.

\subsection{What to Trust: Relative Truth Construction}
\label{sec:conflict}
When cognitive dissonance is detected, the agent must resolve uncertainty about the current environment response distribution $\mathcal{Q}_t(\cdot\mid s_t,a_t)$. GLOVE addresses this by active verification, which re-estimates the local response distribution under the current environment. Specifically, GLOVE re-executes action $a_t$ from state $s_t$ for $\alpha$ trials, where $\alpha$ is the verification budget for this state-action pair. This produces fresh outcomes
\begin{equation}
\mathcal{V} = \{ s'_{t,1}, \ldots, s'_{t,\alpha} \},
\end{equation}
which are treated as i.i.d. samples from $\mathcal{Q}_t(\cdot\mid s_t,a_t)$. GLOVE aggregates these outcomes into an empirical distribution $\hat{\mathcal{Q}}_t(\cdot \mid s_t, a_t)$, referred to as the \emph{relative truth}. This distribution is the agent's current estimate of $\mathcal{Q}_t$ from observable interaction evidence, without requiring oracle labels of transition correctness or memory validity. The term relative indicates that correctness is defined with respect to current environment behavior.

\textbf{Theoretical Guarantee.}
We next characterize the verification budget needed to construct a reliable relative truth. Consider a discrete outcome setting where, for any state-action pair $(s,a)$ at time $t$, the response distribution $\mathcal{Q}_t(\cdot\mid s,a)$ is supported on $K$ outcome modes. Re-executing $(s_t,a_t)$ for $\alpha$ trials gives $\alpha$ i.i.d. samples from $\mathcal{Q}_t(\cdot\mid s_t,a_t)$, and GLOVE forms $\hat{\mathcal{Q}}_t$ as their empirical distribution. Using the concentration inequality for discrete distributions~\cite{weissman2003inequalities}, we obtain:
\begin{theorem}[Verification Budget Requirement]
\label{thm:convergence}
Let $\mathcal{Q}_t$ be the environment response distribution supported on $K$ outcome modes, and let $\hat{\mathcal{Q}}_t$ be the relative truth estimated from $\alpha$ verification samples. For any target accuracy $\varepsilon > 0$ and confidence level $\delta \in (0,1)$, if
\begin{equation}
\label{eq:alpha_requirement}
\alpha \;\ge\; \frac{2\bigl(K \ln 2 + \ln(1/\delta)\bigr)}{\varepsilon^2},
\end{equation}
then with probability at least $1-\delta$,
\begin{equation}
\label{eq:verification_bound}
\| \hat{\mathcal{Q}}_t - \mathcal{Q}_t \|_1 \le \varepsilon,
\end{equation}
where $\|\cdot\|_1$ denotes the $L_1$ distance.
\end{theorem}

\textbf{Interpretation.}
Theorem~\ref{thm:convergence} quantifies how $\alpha$ should scale with the outcome space and the desired reliability. Larger outcome spaces $K$ or stricter requirements $\varepsilon,\delta$ require more verification samples, while stable environments can use smaller budgets. Together with the detection mechanism in Section~\ref{gv-activeProbingAndRelativeTruth}, this gives a closed-loop design: detection decides \emph{when} verification is needed, and Eq.~\ref{eq:alpha_requirement} specifies \emph{how much} interaction is sufficient to estimate the current response distribution. The proof is provided in Appendix~\ref{app:proof_thm_4_2}.

\textbf{Deterministic Special Case.}
If the queried precondition $(s,a)$ has a deterministic response, then $\mathcal{Q}_t(\cdot\mid s,a)$ is supported on a single outcome $s^\star$, with $\mathcal{Q}_t(s^\star\mid s,a)=1$. A single re-execution identifies this outcome, so the empirical distribution $\hat{\mathcal{Q}}_t$ constructed with $\alpha=1$ satisfies $\hat{\mathcal{Q}}_t=\mathcal{Q}_t$ almost surely. Equivalently, $\| \hat{\mathcal{Q}}_t - \mathcal{Q}_t \|_1 = 0$ in~\eqref{eq:verification_bound}, so the accuracy requirement holds for any $\varepsilon>0$ with probability $1$. Thus, one probe is sufficient in deterministic settings, and the verification budget can be set to $\alpha=1$. In practice, $\alpha$ controls the trade-off between verification cost and reliability, so GLOVE starts with a small $\alpha$ and increases it only when inconsistency persists.

\subsection{How to Update: Memory--Environment Realignment}
\label{sec:memory_realign}

After relative truth construction, GLOVE updates the experience bank so that future retrieval no longer depends on obsolete counterparts.
Let $\mathcal{N}(e_t)$ be the counterpart collection in Eq.~\eqref{eqcollect}, and let $\mathcal{V}=\{s'_{t,1},\ldots,s'_{t,\alpha}\}$ be the fresh outcomes collected for $(s_t,a_t)$.
In synchronous verification, GLOVE removes contradicted counterparts and inserts verified outcomes:
\begin{equation}
    \mathcal{D} \leftarrow
    \bigl(\mathcal{D}\setminus_{\mathrm m} \mathcal{N}(e_t)\bigr)
    \uplus
    \biguplus_{j=1}^{\alpha}\{(s_t,a_t,s'_{t,j})\}.
    \label{eq:memory_realign_sync}
\end{equation}
Here $\setminus_{\mathrm m}$ and $\uplus$ denote multiset removal and union, preserving duplicate outcomes as separate samples that induce $\hat{\mathcal{Q}}_t(\cdot\mid s_t,a_t)$ without changing the agent's memory interface.
When exact replay is unavailable, GLOVE applies replay-relaxed asynchronous realignment. It suppresses contradicted counterparts from retrieval, adds the queried precondition to a pending verification queue $\mathcal{P}$, and accumulates evidence when comparable contexts are later encountered through natural interaction. Once sufficient evidence is collected, GLOVE applies the same replacement rule as in Eq.~\eqref{eq:memory_realign_sync}. This fallback avoids repeated execution of irreversible, costly, or safety-critical actions while preventing contradicted memories from continuing to guide future decisions.

\section{Experiments}
\label{experiments}
\begin{table*}[t]
\centering
\caption{\textbf{GLOVE's robustness to environmental drift (Explicit).} Success rate in percent. The shaded rows show the performance of GLOVE-augmented agents. The LLM is GPT-4o. We annotate the performance gap relative to the baseline (e.g., \textcolor{cyan!60!green}{\scriptsize($\uparrow$56.3)} indicates improvement).}
\label{table-gpt4o_explicit}
\footnotesize
\setlength{\tabcolsep}{3.5pt} 
\renewcommand{\arraystretch}{1.1}

\resizebox{1\textwidth}{!}{%
\begin{tabular}{l|ccc|ccc|ccc}
\toprule
\multicolumn{10}{c}{\textbf{\textsc{Verifier-visible structural drift}}} \\
\midrule
& \multicolumn{3}{c|}{\textbf{WebShop} (Semantic Drift)}
& \multicolumn{3}{c|}{\textbf{FrozenLake} (Topology Drift)}
& \multicolumn{3}{c}{\textbf{MountainCar} (Dynamics Drift)} \\
\cmidrule(lr){2-4}\cmidrule(lr){5-7}\cmidrule(lr){8-10}
\textbf{Method} & \textbf{Source} & \textbf{$\to$ Drift I} & \textbf{$\to$ Drift II}
& \textbf{Source} & \textbf{$\to$ Drift I} & \textbf{$\to$ Drift II}
& \textbf{Source} & \textbf{$\to$ Drift I} & \textbf{$\to$ Drift II} \\
\midrule

\text{No Memory (Plain)} & 15   & 5   & 0   & 0   & 5   & 0   & 85   & 75   & 55   \\
\midrule
\text{Vanilla} & 85   & 0   & 0   & 85   & 0   & 0   & 100   & 100   & 90   \\
\rowcolor{gray!10}
\hspace{0.9em}\textbf{+GLOVE} & 85   & 85   \improv{85  } & 95   \improv{95  } & 80   & 75   \improv{75  } & 80   \improv{80  } & 100   & 100   & 100   \improv{10  } \\
\addlinespace[2pt]
\text{MemoryBank} & 85   & 20   & 20   & 80   & 0   & 45   & 95   & 100   & 100   \\
\rowcolor{gray!10}
\hspace{0.9em}\textbf{+GLOVE} & 100   & 90   \improv{70  } & 85   \improv{65  } & 80   & 70   \improv{70  } & 65   \improv{20  } & 95   & 100   & 100   \\
\addlinespace[2pt]
\text{Voyager} & 85   & 0   & 0   & 85   & 0   & 0   & 100   & 90   & 100   \\
\rowcolor{gray!10}
\hspace{0.9em}\textbf{+GLOVE} & 80   & 90   \improv{90  } & 95   \improv{95  } & 85   & 85   \improv{85  } & 75   \improv{75  } & 100   & 100   \improv{10  } & 100   \\
\addlinespace[2pt]
\text{Generative Agent} & 80   & 0   & 0   & 80   & 15   & 0   & 100   & 95   & 70   \\
\rowcolor{gray!10}
\hspace{0.9em}\textbf{+GLOVE} & 75   & 95   \improv{95  } & 95   \improv{95  } & 80   & 80   \improv{65  } & 85   \improv{85  } & 95   & 100   \improv{5  } & 100   \improv{30  } \\

\bottomrule
\end{tabular}%
}
\end{table*}

\subsection{Experiment Setup}
\label{experiments-setup}

\textbf{Environments.}
We evaluate GLOVE on three benchmarks and one real-world robotic deployment. The benchmarks include \textbf{WebShop}~\cite{yao2022webshop} for web navigation, \textbf{FrozenLake}~\cite{brockman2016openai} for discrete planning, and \textbf{MountainCar}~\cite{brockman2016openai} for continuous control. To test deployment under physical noise, we also evaluate GLOVE on a \textbf{JetRover} mobile robot, where the agent must navigate a real maze and update stale target-location memory after semantic drift. 
These environments cover semantic reasoning, discrete planning, continuous dynamics, and embodied memory realignment.
We introduce controlled environmental drifts as difficulty-enhanced evaluation settings. 
Agents first build memories in the source environment and then continue solving the same task after drift, allowing us to measure adaptation under stale memory. Specifically, \textit{Explicit Drift} refers to observable structural changes, including altered layouts, transition dynamics, or semantic mappings. We report success rate over twenty rounds. \textit{Implicit Drift} refers to changes in hidden task logic, such as reward reversals or hidden transition rules, that are not directly observable from immediate state descriptions alone. We report the score obtained over twenty rounds. All drift settings are implemented as systematic benchmark-level modifications and applied uniformly across methods without algorithm-specific tuning. Detailed specifications are provided in Appendix~\ref{subsec:explicit_env_drifts} and Appendix~\ref{subsec:implicit_env_drifts}.

\textbf{Baselines.}
We evaluate GLOVE across representative memory architectures to test whether it functions as a general augmentation module. The baselines include (1) \textbf{No Memory}, a zero-shot agent without external context; (2) \textbf{Vanilla}, a basic RAG agent; and (3) agentic memory architectures, including \textbf{Voyager}~\cite{wang2024voyager}, \textbf{MemoryBank}~\cite{zhong2024memorybank}, and \textbf{Generative Agents}~\cite{park2023generative}. Each architecture is evaluated in its original form and with GLOVE, isolating the effect of memory verification and realignment. These baselines cover plain retrieval, passive decay, reflective memory, and iterative-refinement memory backends. Details are provided in Appendix~\ref{appendix-baseline_setup}.
We focus on memory-augmented LLM agents as the main comparison class, while Appendix~\ref{app:dynaq_positioning} provides an additional positioning check against adaptive model-based RL methods.

\textbf{LLM Backbones.}
We evaluate GLOVE across diverse LLM backbones to test architecture-agnostic performance, including open-weights models including \textbf{Llama-3.1-8B}, \textbf{Llama-3.3-70B}, \textbf{Qwen2.5-7B}, \textbf{Qwen3-30B}, \textbf{DeepSeek-V3.2}, as well as proprietary models, such as \mbox{\textbf{GPT-4o}} and \textbf{Grok-3}.

\subsection{Memory Realignment under Environmental Drift}
\paragraph{Verifier-visible drift.}\label{exp-explicit}
We evaluate agents under explicit structural drift by transferring memories collected in a source environment to drifted variants with altered semantics, topology, or dynamics. Table~\ref{table-gpt4o_explicit} reports success rates before and after drift using GPT-4o as the backbone. Static memory retrieval leads to severe degradation once the environment changes: under WebShop semantic drift (Drift I), Voyager drops from 85\% to 0\%, and under FrozenLake topology drift (Drift II), Generative Agent drops from 80\% to 0\%. These failures show that memory systems performing well in the source environment can still lack mechanisms to realign stored experience after structural change. MemoryBank provides limited robustness through time-based forgetting, but this passive adaptation is slow and can retain high-confidence invalid memories; for example, it still reaches 0\% success in FrozenLake Drift I. Adding GLOVE yields consistent improvements across explicit drift settings by detecting conflicts between stored memories and observed outcomes and then realigning the experience bank. In WebShop, GLOVE-augmented agents recover to around 90\% success after semantic drift, compared with 20\% for MemoryBank. In FrozenLake Drift II, GLOVE improves performance by an average of 65\% over non-augmented agents, indicating effective correction of outdated topological memories. In MountainCar, GLOVE maintains near-perfect performance, showing adaptation to changed continuous dynamics. These gains hold across multiple LLM backbones, as summarized in Table~\ref{tab:appendix_backbone_sweep} in Appendix~\ref{appendix-exp_result_explicit}. Appendix~\ref{app:fulltraj_ablation} further shows that broader episodic context alone does not remove stale-memory failure under drift, supporting that the observed gains come from explicit memory verification and realignment rather than a specific backbone, memory format, or context length.

\begin{wraptable}{r}{0.6\linewidth}
\vspace{-10pt}
\centering
\caption{GPT-4o: GLOVE's robustness to implicit environmental drift.}
\label{table-gpt4o_implicit}
\footnotesize
\setlength{\tabcolsep}{1pt}
\renewcommand{\arraystretch}{1}
\begin{tabular}{l|ccccc|cccc}
\toprule
\multicolumn{10}{c}{\textbf{\textsc{Verifier-hidden drift}}} \\
\midrule
& \multicolumn{5}{c|}{\makecell{\textbf{WebShop} \\ \footnotesize (Semantic Drift)}}
& \multicolumn{4}{c}{\makecell{\textbf{FrozenLake} \\ \footnotesize (Reward Reversal)}} \\
\cmidrule(lr){2-6}\cmidrule(lr){7-10}
\textbf{Method}
& \multicolumn{2}{c}{\textbf{Source}} & \multicolumn{3}{c|}{$\to$ \textbf{Hidden drift}}
& \multicolumn{2}{c}{\textbf{Source}} & \multicolumn{2}{c}{$\to$ \textbf{Hidden drift}} \\
\midrule
\text{No Memory}
& \multicolumn{2}{c}{82.5} & \multicolumn{3}{c|}{87.5} & \multicolumn{2}{c}{0  } & \multicolumn{2}{c}{0  } \\
\midrule
\text{Vanilla}
& \multicolumn{2}{c}{100  } & \multicolumn{3}{c|}{75  } & \multicolumn{2}{c}{47.5} & \multicolumn{2}{c}{50  } \\
\rowcolor{gray!10}
\hspace{0.9em}\textbf{+GLOVE}
& \multicolumn{2}{c}{97.5} & \multicolumn{3}{c|}{93.8 \improv{18.8}} & \multicolumn{2}{c}{35  } & \multicolumn{2}{c}{97.5 \improv{47.5}} \\
\addlinespace[2pt]
\text{MemoryBank}
& \multicolumn{2}{c}{100  } & \multicolumn{3}{c|}{81.2} & \multicolumn{2}{c}{62.5} & \multicolumn{2}{c}{57.5} \\
\rowcolor{gray!10}
\hspace{0.9em}\textbf{+GLOVE}
& \multicolumn{2}{c}{98.8} & \multicolumn{3}{c|}{98.8 \improv{17.5}} & \multicolumn{2}{c}{67.5} & \multicolumn{2}{c}{97.5 \improv{40  }} \\
\addlinespace[2pt]
\text{Voyager}
& \multicolumn{2}{c}{98.8} & \multicolumn{3}{c|}{75  } & \multicolumn{2}{c}{67.5} & \multicolumn{2}{c}{50  } \\
\rowcolor{gray!10}
\hspace{0.9em}\textbf{+GLOVE}
& \multicolumn{2}{c}{98.8} & \multicolumn{3}{c|}{98.8 \improv{23.8}} & \multicolumn{2}{c}{40  } & \multicolumn{2}{c}{97.5 \improv{47.5}} \\
\addlinespace[2pt]
\text{Generative Agent}
& \multicolumn{2}{c}{98.8} & \multicolumn{3}{c|}{75  } & \multicolumn{2}{c}{62.5} & \multicolumn{2}{c}{50  } \\
\rowcolor{gray!10}
\hspace{0.9em}\textbf{+GLOVE}
& \multicolumn{2}{c}{98.8} & \multicolumn{3}{c|}{75  } & \multicolumn{2}{c}{62.5} & \multicolumn{2}{c}{92.5 \improv{42.5}} \\
\bottomrule
\end{tabular}
\end{wraptable}
\textbf{Verifier-hidden drift.}
We evaluate GLOVE under implicit drift, including reward reversals and changes in hidden transition logic. Unlike explicit drift, implicit drift may leave the immediate observation space unchanged, causing memory-based agents to trust stale experience that remains superficially consistent with the current state. As shown in Table~\ref{table-gpt4o_implicit}, this can make stored experience harmful after drift: MemoryBank, despite its forgetting mechanism, suffers an 18.8\% score decrease under the hidden WebShop semantic change. GLOVE addresses this failure by checking observable interaction consequences, including both outcome $s'$ and associated reward dynamics. In Table~\ref{table-gpt4o_implicit}, adding GLOVE improves the Vanilla agent from 47.5\% to 97.5\% in FrozenLake hidden drift. In the Grok-3 benchmark, Voyager with GLOVE reaches 100\% success, compared with 62.5\% for Voyager alone, as reported in Table~\ref{tab:plugplay_memdec_style_Grok_3_implicit}. These results show that GLOVE can realign memory even when the drift is not visible from state descriptions alone. More generally, explicit drift can be detected from changes in state-action-outcome transitions, while implicit drift requires observable downstream consequences such as reward, score, user response, or delayed task feedback. GLOVE uses these signals as interaction evidence rather than oracle labels of memory validity. If both the interaction process and all feedback remain unchanged, the drift is observationally indistinguishable from stationarity for any experience-based method.

\subsection{Statistical Reliability}
\label{sec:statistical_reliability}

We further assess the statistical reliability of the main recovery trend
through 10 paired runs on the Qwen3-30B FrozenLake explicit-drift setting.
For each run and memory backend, the baseline and its GLOVE variant use
byte-equivalent source-memory snapshots and identical task instances.
In Drift II, GLOVE improves Vanilla from $0.0\pm0.0\%$ to
$75.0\pm0.0\%$ and MemoryBank from $18.5\pm7.5\%$ to
$75.0\pm0.0\%$. Both improvements are positive in all 10 paired runs and
remain significant after Bonferroni correction
(adjusted $p=0.0078$ for both; Cohen's $h=2.094$ and $1.205$).
For settings without repeated paired runs, we do not make
statistical-significance claims.
Full counts, confidence intervals, and paired tests are reported in
Appendix~\ref{app:adaptive_statistical_analysis}.

\subsection{Comparison with Direct Online Memory Adaptation}
\label{sec:adaptive_memory_baselines}
To separate GLOVE's gains from simple online updating and from the
extra-interaction confound, we compare against six direct adaptive rules
on Qwen3-30B FrozenLake explicit drift:
\textsc{LatestOverwrite}, \textsc{RecentK-K3},
\textsc{SlidingWindow-W5}, \textsc{RecencyWeighted-H5},
\textsc{PageHinkleyReset}, and \textsc{PeriodicRefresh-S30}.
In the natural protocol, each baseline updates only from its own trajectory,
while GLOVE retains standard verification.
We also impose a strict $B=310$ post-drift interaction horizon:
GLOVE uses 280 task interactions plus 30 verification interactions,
whereas each baseline may spend all 310 on task execution.

\begin{table}[t]
\centering
\caption{\textbf{Direct adaptive-memory comparison on Qwen3-30B FrozenLake
explicit drift.}
Natural columns use 20 episodes per drift phase; all methods achieve
20/20 in Source.
The last column reports completed successes under the same $B=310$
post-drift interaction budget.}
\label{tab:adaptive_equal_budget_main}
\small
\setlength{\tabcolsep}{6pt}
\renewcommand{\arraystretch}{1.05}
\begin{tabular}{lccc}
\toprule
& \multicolumn{2}{c}{\textbf{Natural adaptation}}
& \textbf{Fixed budget} \\
\cmidrule(lr){2-3}\cmidrule(lr){4-4}
\textbf{Method}
& \textbf{Drift I}
& \textbf{Drift II}
& \textbf{Successes @ $B=310$} \\
\midrule
\textbf{GLOVE}
& \textbf{16/20}
& \textbf{15/20}
& \textbf{31} \\
LatestOverwrite
& 16/20
& 0/20
& 16 \\
RecentK-K3
& 16/20
& 0/20
& 16 \\
SlidingWindow-W5
& 14/20
& 0/20
& 14 \\
RecencyWeighted-H5
& 12/20
& 0/20
& 12 \\
PageHinkleyReset
& 14/20
& 0/20
& 14 \\
PeriodicRefresh-S30
& 5/20
& 7/20
& \underline{18} \\
\bottomrule
\end{tabular}
\end{table}

The pair-local rules expose a \emph{self-confirming exploration deadlock}.
After Drift I, the stale \texttt{Down$\rightarrow$Hole} record discourages
the policy from selecting Down even when it becomes safe again in Drift II;
LatestOverwrite selects Down zero times across 381 Drift-II actions and
therefore never observes the corrective transition.
PeriodicRefresh can restore exploration through blind forgetting, but also
removes valid memories.
GLOVE instead targets suspected stale records for verification, allowing
correction even when the current policy avoids the corresponding action.
Under the same $B=310$ horizon, GLOVE achieves 31 successes versus 18 for
the strongest adaptive baseline.
Full protocols are in Appendix~\ref{app:adaptive_statistical_analysis}.

\subsection{Verification Assumptions and Cost}
\label{sec:replay_relaxed}

\textbf{Replay-relaxed verification.}
We evaluate whether GLOVE remains effective when exact replay is unavailable. In the replay-feasible setting, the agent can re-execute a matched state-action pair to estimate the current environment response. This assumption can fail in continuous, irreversible, or high-stakes environments where the agent cannot safely reset and repeat an action. We therefore evaluate a replay-relaxed asynchronous variant, which treats exact replay as the strongest verification mode and asynchronous verification as a safer fallback.

\begin{wrapfigure}{r}{0.5\linewidth}
\vspace{-10pt}
    \centering
    \includegraphics[width=\linewidth]{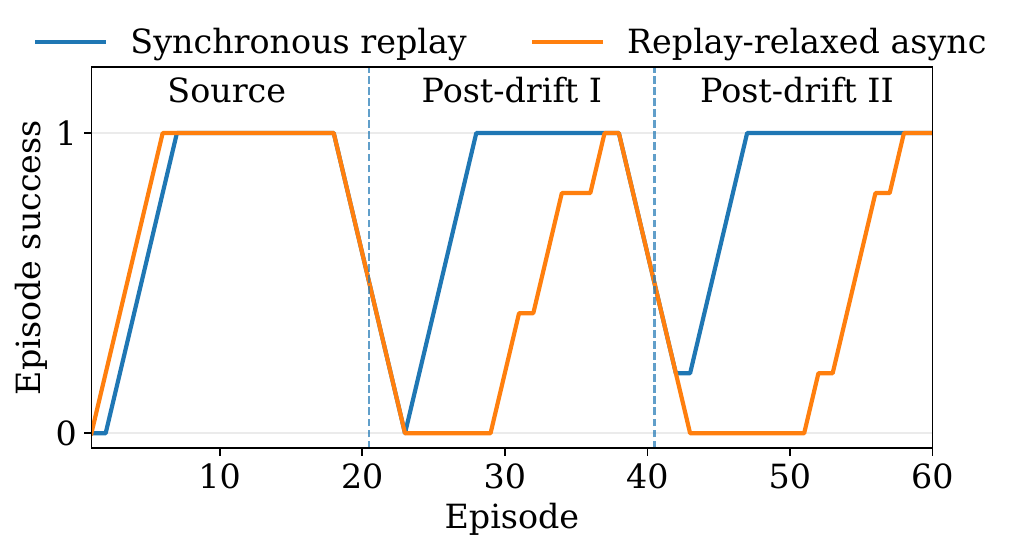}
    \caption{Replay-feasible versus replay-relaxed verification in FrozenLake.}
    \label{fig:async_recovery}
\vspace{-7pt}
\end{wrapfigure}
 Instead of re-executing a consequential action, GLOVE suppresses the contradicted memory and waits for non-destructive confirmation from future natural encounters, passive observations, sandboxed checks, or task feedback. As shown in Fig.~\ref{fig:async_recovery}, asynchronous verification recovers at Episode~32 and Episode~54 under the two drift blocks, compared with Episode~26 and Episode~45 for synchronous replay. Thus, removing exact replay delays recovery but does not prevent memory realignment, showing that GLOVE remains effective under weaker deployment assumptions.

\textbf{Budget sensitivity.}
Active verification adds environment interactions, but the overhead is selective. Appendix~\ref{app:cost_analysis} shows that verification steps concentrate around drift events rather than stable periods. The same analysis shows that the probing budget $\alpha$ controls a cost--reliability trade-off, while moderate and strict counterpart-matching granularities achieve the same average reward with similarly low verification overhead.

\subsection{Real-Robot Evaluation on JetRover}
\label{sec:real_robot}
\begin{wrapfigure}{r}{0.5\linewidth}
\vspace{-10pt}
    \centering
    \includegraphics[width=1\linewidth]{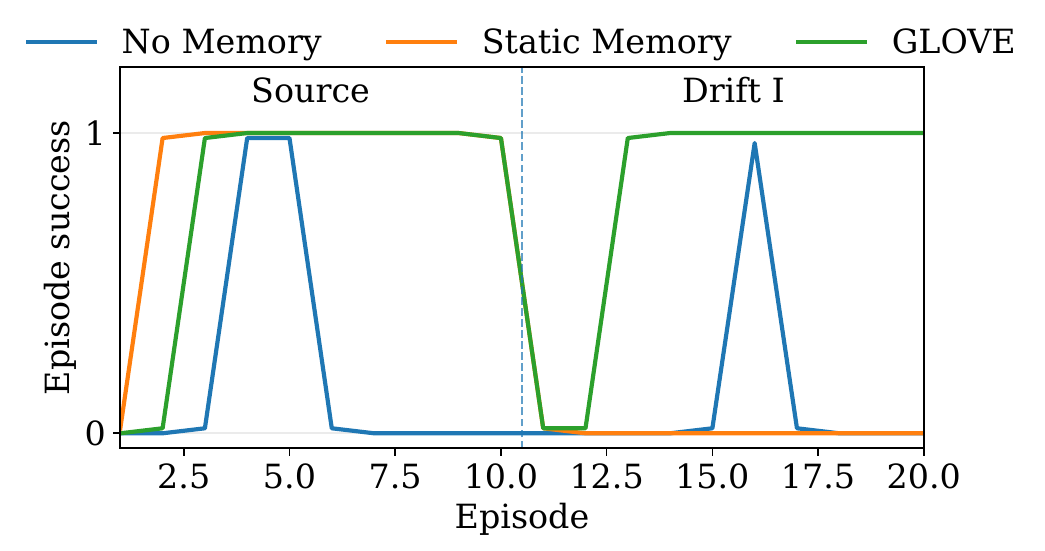}
    \caption{Real-robot evaluation on JetRover.}
    \label{fig:jetrover_adaptation}
\end{wrapfigure}
We evaluate GLOVE in a real-robot setting on a HiWonder JetRover~\cite{hiwonder_jetrover_docs} robot.
Experimental details are provided in Appendix~\ref{app:jetrover_setup}.
The setup includes physical nuisance variation, such as irregular cardboard walls, small target displacement, and added path obstacles.
The robot searches for the blue block.
After source-memory formation, we induce semantic drift by swapping the blue and green blocks, testing whether GLOVE distinguishes persistent target-location drift from incidental physical noise in deployment.
As shown in Fig.~\ref{fig:jetrover_adaptation}, No Memory fails to find the target reliably, while Static Memory succeeds before drift but collapses after the swap because it keeps revisiting the stale target location.
In contrast, GLOVE detects the semantic mismatch, updates the stale memory, and recovers correct search behavior while remaining stable under incidental physical variation.

\section{Conclusion}
\label{sec-conclusion}

We introduced GLOVE, a plug-and-play memory-validation module for memory-augmented LLM agents in dynamic environments. GLOVE detects cognitive dissonance between stored experience and current interaction evidence, constructs relative truth from fresh observations, and realigns the memory bank by replacing obsolete entries. 
Across web navigation, discrete planning, continuous control, and real-robot navigation, GLOVE improves adaptation under both explicit and implicit drift while remaining compatible with diverse memory architectures and LLM backbones. 
These results identify memory-environment misalignment as one of the core reliability issues for long-horizon agents and show that deployment-time memory realignment can improve robustness.

\begin{ack}
We thank Yaoyutong Shen for helpful hardware robotic experimental setup during her internship at the University of Hong Kong.

\textbf{Funding.} This work received no specific external funding.

\textbf{Competing interests.}
The authors declare no competing interests related to this work.
\end{ack}

\bibliographystyle{plainnat}
\bibliography{ref}

\appendix

\clearpage
\appendix
\onecolumn
\raggedbottom

\section*{Appendix Contents}

\begin{itemize}
    \setlength{\itemsep}{0.2em}

    \item[\textbf{A.}] \textbf{Limitation and Broader Impact} \dotfill Page \pageref{limitation_and_broader_impact}
    \begin{itemize}
        \setlength{\itemsep}{0em}
        \item[A.1] Limitation \dotfill Page \pageref{limitation}
        \item[A.2] Broader Impact \dotfill Page \pageref{broader_impact}
    \end{itemize}

    \item[\textbf{B.}] \textbf{Theoretical Analysis} \dotfill Page \pageref{app:proofs}
    \begin{itemize}
        \setlength{\itemsep}{0em}
        \item[B.1] Proof of Theorem~\ref{thm:detection}: Finite-Sample Detection Bound \dotfill Page \pageref{app:proof_thm_4_1}
        \item[B.2] Proof of Theorem~\ref{thm:convergence}: Verification Budget Requirement \dotfill Page \pageref{app:proof_thm_4_2}
    \end{itemize}

    \item[\textbf{C.}] \textbf{Experimental Details} \dotfill Page \pageref{sec:experimental_details}
    \begin{itemize}
        \setlength{\itemsep}{0em}
        \item[C.1] Baseline Setup \dotfill Page \pageref{appendix-baseline_setup}
        \item[C.2] Environment Setup \dotfill Page \pageref{appendix-environment_setup}
        \item[C.3] State Formation and Matching Rules \dotfill Page \pageref{app:state_formation}
        \item[C.4] Explicit Environmental Drifts \dotfill Page \pageref{subsec:explicit_env_drifts}
        \item[C.5] Implicit Environmental Drifts \dotfill Page \pageref{subsec:implicit_env_drifts}
        \item[C.6] JetRover Real-Robot Setup \dotfill Page \pageref{app:jetrover_setup}
        \item[C.7] Case Study: FrozenLake Implicit Environmental Drifts \dotfill Page \pageref{subsec:case_study}
    \end{itemize}

    \item[\textbf{D.}] \textbf{Additional Experiment Results} \dotfill Page \pageref{appendix-experiment_results}
    \begin{itemize}
        \setlength{\itemsep}{0em}
        \item[D.1] Experiment Results: Backbone Sweep Overview \dotfill Page \pageref{appendix-backbone_sweep_overview}
        \item[D.2] Experiment Results: Explicit Environmental Drifts \dotfill Page \pageref{appendix-exp_result_explicit}
        \item[D.3] Experiment Results: Implicit Environmental Drifts \dotfill Page \pageref{appendix-exp_result_implicit}
        \item[D.4] Positioning against Adaptive Model-Based RL \dotfill Page \pageref{app:dynaq_positioning}
        \item[D.5] Cost Analysis \dotfill Page \pageref{app:cost_analysis}
        \item[D.6] JetRover Real-World Robotic Result \dotfill Page \pageref{app:jetrover_result}
        \item[D.7] Result Analysis \dotfill Page \pageref{app:result_analysis}
    \end{itemize}

\end{itemize}

\section{Limitation and Broader Impact}
\label{limitation_and_broader_impact}

\subsection{Limitation}
\label{limitation}

GLOVE requires observable interaction evidence to detect memory-environment misalignment. Explicit drift can be detected from changed state-action-outcome transitions, while implicit drift requires downstream feedback such as reward, score, user response, or delayed task feedback. If neither the interaction process nor any feedback changes, the drift is observationally indistinguishable from stationarity for any experience-based method. GLOVE also depends on the cost and safety of evidence collection. In replay-feasible settings, active probing can directly estimate the current local response pattern. For costly, irreversible, or safety-critical actions, GLOVE uses replay-relaxed asynchronous realignment by suppressing contradicted memories and waiting for safe evidence from future encounters, proxy checks, or delayed feedback. This avoids unsafe repetition, but recovery can be slower and replacement memories may remain incomplete when safe feedback is sparse. Finally, the JetRover experiment provides a physical proof of concept under irregular walls, target displacement, and path-level obstacles, but it still uses a pre-constructed navigation map and waypoints. Extending GLOVE to fully open-world embodied agents, including 3D navigation platforms such as drones or autonomous vehicles, remains future work.

\subsection{Broader Impact}
\label{broader_impact}

GLOVE improves the reliability of memory-augmented LLM agents in dynamic environments by verifying retrieved memories against fresh interaction evidence. This may benefit long-horizon applications such as web automation, assistive agents, and embodied robotics, where stale experience can otherwise cause persistent failures. However, active probing also introduces risks where probes may trigger unwanted actions or extra interaction costs, especially in safety-critical or user-facing settings. Moreover, GLOVE constructs relative truth from observable consequences, which should not be treated as a substitute for safety, fairness, or human preference alignment. We therefore view GLOVE as a reliability layer rather than a standalone safety mechanism. Deployments should use task-level safety constraints, human approval for irreversible actions, and fixed policies that override learned experience. GLOVE should not be used for malicious automation, surveillance, deception, or other harmful applications.

\section{Theoretical Analysis}
\label{app:proofs}

This section provides a formal analysis of the two statistical components underlying GLOVE. The analysis mirrors the two-stage structure of the framework. First, we study the reliability of cognitive dissonance detection under finite historical samples, establishing a false-alarm bound based on Hoeffding's inequality. Second, we analyze the active verification process and derive a sample-complexity requirement for constructing a reliable relative truth, using concentration results for discrete distributions due to Weissman et al.~\cite{weissman2003inequalities}.

\subsection{Proof of Theorem~\ref{thm:detection}: Finite-Sample Detection Bound}
\label{app:proof_thm_4_1}

We prove Theorem~\ref{thm:detection}. Consider a fixed state--action pair $(s,a)$ and assume that, over a local time window, the environment response distribution $\mathcal{Q}_t(\cdot\mid s,a)$ remains unchanged. Let $s'$ be any specific outcome.

Each historical execution of $(s,a)$ produces an outcome $S'_i$ drawn from $\mathcal{Q}_t(\cdot\mid s,a)$. Define the indicator random variables
\[
X_i = \mathbf{1}\{S'_i = s'\}, \qquad i=1,\ldots,n.
\]
Under the locally stationary assumption, $\{X_i\}_{i=1}^n$ are independent and identically distributed Bernoulli random variables with
\[
\mathbb{E}[X_i] = \mathcal{Q}_t(s'\mid s,a).
\]

The empirical probability of observing outcome $s'$ is
\[
\hat{\mathcal{Q}}_{\mathrm{hist}}(s'\mid s,a)
= \frac{1}{n}\sum_{i=1}^n X_i.
\]

By Hoeffding's inequality, for any $\lambda > 0$,
\[
\Pr\!\left(
\left| \hat{\mathcal{Q}}_{\mathrm{hist}}(s'\mid s,a)
- \mathcal{Q}_t(s'\mid s,a) \right| \ge \lambda
\right)
\;\le\;
2\exp(-2n\lambda^2).
\]

Setting the right-hand side equal to $\delta$ and solving for $\lambda$ yields
\[
\lambda = \sqrt{\frac{\ln(2/\delta)}{2n}}.
\]

Therefore, with probability at least $1-\delta$,
\[
\left| \hat{\mathcal{Q}}_{\mathrm{hist}}(s'\mid s,a)
- \mathcal{Q}_t(s'\mid s,a) \right|
\le
\sqrt{\frac{\ln(2/\delta)}{2n}},
\]
which proves Eq.~\eqref{eq:detection_bound} and completes the proof.

\hfill $\square$

\subsection{Proof of Theorem~\ref{thm:convergence}: Verification Budget Requirement}
\label{app:proof_thm_4_2}

We prove Theorem~\ref{thm:convergence}. Fix a state--action pair $(s,a)$ at time $t$ and assume that the environment response distribution $\mathcal{Q}_t(\cdot\mid s,a)$ is supported on a finite set of $K$ distinct outcome modes. Let
\[
\mathcal{Q}_t = (q_1, \ldots, q_K)
\]
denote this categorical distribution.

Active verification re-executes $(s,a)$ for $\alpha$ trials, producing outcomes
\[
S'_1, \ldots, S'_\alpha \sim \mathcal{Q}_t(\cdot\mid s,a),
\]
which are independent and identically distributed by construction. The relative truth $\hat{\mathcal{Q}}_t$ is defined as the empirical distribution of these samples.

The problem therefore reduces to estimating a discrete distribution from $\alpha$ i.i.d. samples. By Theorem 2 of Weissman et al.~\cite{weissman2003inequalities}, for any $\varepsilon > 0$,
\[
\Pr\!\left(
\| \hat{\mathcal{Q}}_t - \mathcal{Q}_t \|_1 > \varepsilon
\right)
\;\le\;
2^{K} \exp\!\left(-\frac{\alpha \varepsilon^2}{2}\right).
\]

Setting the right-hand side to $\delta$ and solving for $\alpha$ gives
\[
\alpha \;\ge\; \frac{2\bigl(K \ln 2 + \ln(1/\delta)\bigr)}{\varepsilon^2}.
\]

Under this condition, with probability at least $1-\delta$,
\[
\| \hat{\mathcal{Q}}_t - \mathcal{Q}_t \|_1 \le \varepsilon,
\]
which proves Eqs.~\eqref{eq:alpha_requirement} and~\eqref{eq:verification_bound}, completing the proof.

\hfill $\square$

\section{Experimental Details}
\label{sec:experimental_details}

\subsection{Baseline Setup}
\label{appendix-baseline_setup}

In this section, we provide detailed descriptions of each baseline for comparison in our experiments: 

\paragraph{Vanilla Agent.} 
The Vanilla agent implements a fundamental RAG architecture. Given the current state $s$, the system queries the external memory to retrieve the top-$k$ historical experiences with the same state $s$, utilizing them directly as in-context exemplars for action generation

\paragraph{Voyager.} 
Derived from the Voyager agent~\cite{wang2024voyager}, the Voyager memory employs an iterative prompting mechanism, continuously acquiring and refining actions. In our experiment, we utilize its feedback-based refinement capability to evaluate experience entries.

\paragraph{MemoryBank.} 
MemoryBank~\cite{zhong2024memorybank} implements a memory updating mechanism inspired by the Ebbinghaus forgetting curve, which passively decays the retrieval weights of older experiences to simulate human forgetting. We include this baseline to represent passive adaptation strategies, providing a rigorous contrast to the active verification paradigm introduced in GLOVE.

\paragraph{Generative Agents.}
Generative Agents~\cite{park2023generative} synthesize high-level insights from experience entries through reflection during the retrieval phase. We evaluate whether introspection alone is sufficient to detect logical inconsistencies in dynamic environments without active probing.

\subsection{Environment Setup}
\label{appendix-environment_setup}

In this section, we provide the environments used in our experiments:

\paragraph{WebShop.}
WebShop~\cite{yao2022webshop} is a scalable simulated e-commerce environment that requires agents to navigate webpages and process semantic instructions to locate desired products. 
We adapt this environment to evaluate \textbf{semantic reasoning} under structural drifts and implicit drifts. 
We employ this environment to challenge the agent's ability to ground natural language instructions in an evolving website setting.
Details of the drift settings are in Section~\ref{subsec:explicit_env_drifts} and Section~\ref{subsec:implicit_env_drifts}.

\paragraph{FrozenLake.}
FrozenLake~\cite{brockman2016openai} is a classic grid-map environment, which we employ as a benchmark for \textbf{discrete planning} and spatial navigation.  Described in detail in Section~\ref{subsec:explicit_env_drifts} and Section~\ref{subsec:implicit_env_drifts}, we introduce topological shifts where the map layout and obstacle placement change explicitly, as well as implicit reward reversals where previously safe goals become traps, testing agents' ability to adapt to changing topological environments.

\paragraph{MountainCar.}
MountainCar~\cite{brockman2016openai} represents a standard continuous control problem where a vehicle in a valley must build momentum to reach a target hilltop. We introduce dynamic shifts by altering the engine force magnitude, thereby rigorously testing the agent's ability to adapt to dynamic physical feedback. Details of the drift settings are provided in Section~\ref{subsec:explicit_env_drifts}.

\subsection{State Formation and Matching Rules}
\label{app:state_formation}

For all environments, the agent does not observe the underlying environment state directly. Instead, it operates on an observation-derived state representation that is passed to the LLM, stored in memory, and reused for counterpart matching during verification. Importantly, this matching interface should not be confused with the overall memory format used for action generation: in trajectory-based memory backends, matched states are linked into longer action paths and may be further summarized into higher-level natural-language memory, while GLOVE applies local matching only at the verification stage.

\paragraph{WebShop.}
In WebShop, the state is constructed from the current webpage content and navigation context. Specifically, we use denoised HTML text that removes scripts and irrelevant markup, together with the current URL. This representation captures the visible product information and page structure available to the agent at decision time.

For counterpart retrieval, we use a compact confirmation-state representation derived from decision-relevant observable attributes rather than requiring exact equality over the full raw page. In line with the plug-and-play nature of GLOVE, this matching can reuse the same representation and retrieval mechanism already used by the underlying memory system. We additionally study a controllable granularity parameter $\eta$: $\eta=0.0$ keeps no attribute content, $\eta=0.5$ keeps only partial attribute information such as \texttt{\{"color":"red"\}}, and $\eta=1.0$ keeps the full attribute string such as \texttt{\{"color":"red","size":"x-large"\}}. Larger $\eta$ therefore yields stricter counterpart matching, while smaller $\eta$ merges more contexts and can trigger unnecessary verification. Appendix~\ref{app:cost_analysis} reports the corresponding sensitivity study.

\begin{quote}
\small\raggedright
\texttt{"state": \{"html": "Back to Search Page 1 (Total results: 50) Next > ...", "url": "http://127.0.0.1:3000/search\_results/<session\_id>/<query>/1"\}}
\end{quote}

\paragraph{FrozenLake.}
In FrozenLake, the state is obtained from the environment observation and includes the agent's current grid position and the corresponding tile type. This compact representation reflects the fully observable but discrete environment used for controlled analysis. Counterpart matching uses exact equality on the observation-derived state stored in memory.

\begin{quote}
\small\raggedright
\texttt{"state": \{"cur\_pos": [1,1], "tile\_type": "F", "gold\_collected": 1\}}
\end{quote}

Under implicit scenarios, destinations may contain fractional reward:

\begin{quote}
\small\raggedright
\texttt{"state": \{"cur\_pos": [1,1], "tile\_type": "G", "gold\_collected": 0.5\}}
\end{quote}

\paragraph{MountainCar.}
In MountainCar, the state consists of the continuous position and velocity returned by the environment. This setting evaluates GLOVE under continuous control dynamics where small environment changes can invalidate prior experience. Counterpart retrieval follows the same observation-derived state representation stored in memory.

\begin{quote}
\small\raggedright
\texttt{"state": \{"position": -0.477, "velocity": 0.0068\}}
\end{quote}

\paragraph{JetRover Maze.}
In JetRover, the agent operates on a semantic inspection state rather than raw-image equality. Each memory entry stores the queried target object together with its last verified location. During verification, a mismatch is triggered when the robot revisits the memorized target location and Qwen2.5-VL identifies a different object there. Counterpart matching is therefore performed at the level of target-conditioned inspection context, which tolerates small viewpoint changes, centimeter-scale target displacement, and path-level obstacle variation while still allowing stale target-location memory to be falsified after drift.

\paragraph{Full-Trajectory Ablation} To test whether stale-memory failure is induced by local state-matched retrieval, we additionally evaluate a trajectory-level context baseline in WebShop explicit drift. We first collect successful trajectories in the source environment. During drifted evaluation, \textsc{FullTraj-Static} selects a successful source trajectory at the beginning of each episode using only pre-drift information and keeps it fixed for the entire episode. At every LLM decision step, the prompt includes the same complete step-by-step source trajectory, together with the current observation and the current episode history. It does not include current-state matched one-step memory snippets for action generation. This protocol tests whether broader episodic context alone can make the agent discount stale source-environment experience. In the GLOVE condition, local counterpart matching is still used only for verification, as in Eq.~\eqref{eqcollect}, because GLOVE aims to identify which specific state-action transition has become stale rather than treating an entire trajectory as an indivisible verification unit.

\subsection{Explicit Environmental Drifts}
\label{subsec:explicit_env_drifts}

We construct explicit environmental drifts by changing the environment structures that are observable through interaction. These drifts modify the mapping from actions to outcomes, including navigation structure in WebShop, obstacle topology in FrozenLake, and physical dynamics in MountainCar. The purpose is to test whether an agent can detect that a previously valid memory no longer matches the current environment.

\begin{figure}[!htbp]
    \centering
    \includegraphics[width=0.95\textwidth]{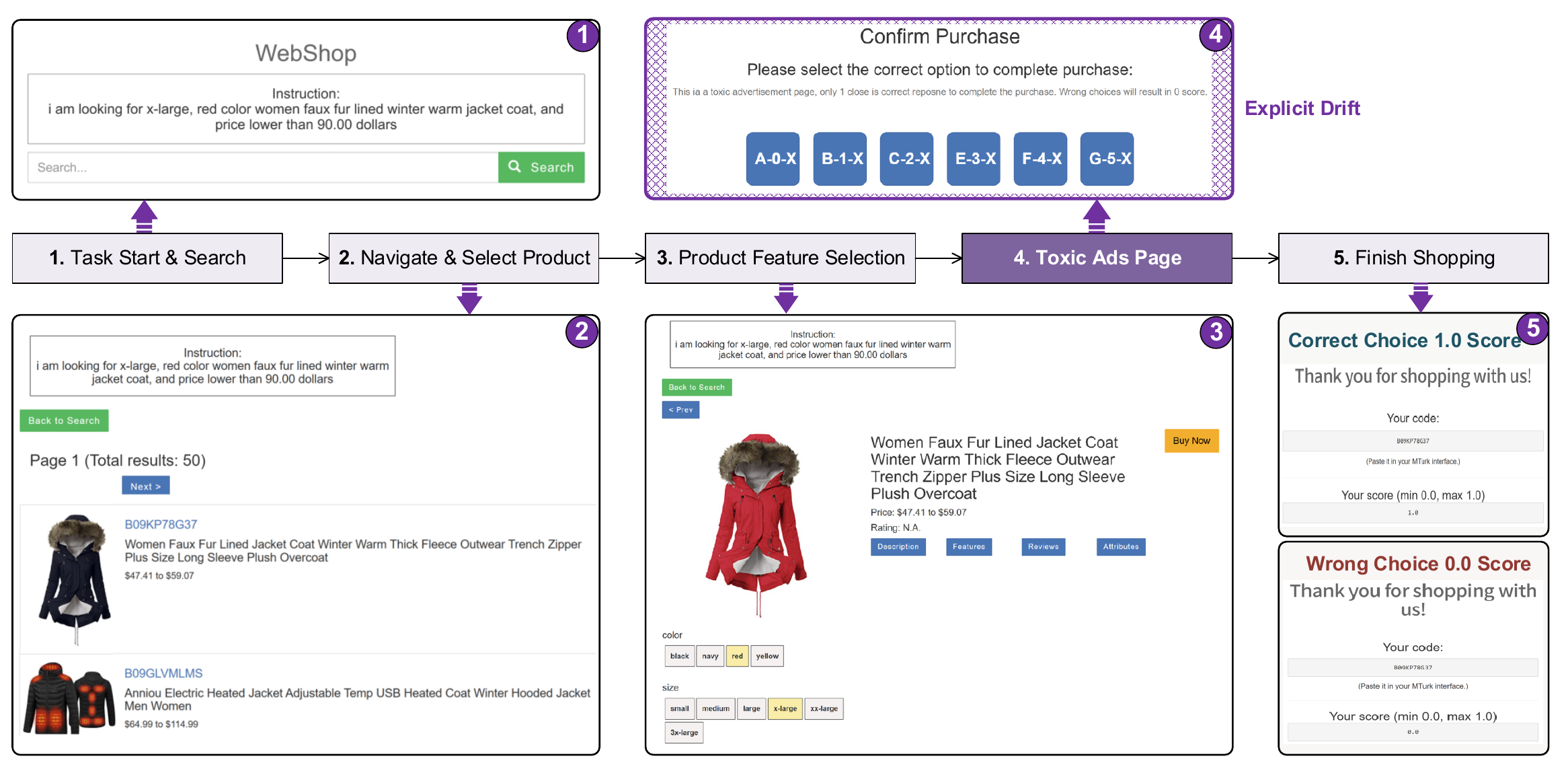}
    \caption{WebShop explicit drift.}
    \label{fig:webshop_drift_flow}
\end{figure}

\paragraph{WebShop.}
As shown in Fig.~\ref{fig:webshop_drift_flow}, the WebShop task proceeds normally until the agent selects the ``Buy Now'' action. After this action, the environment redirects the agent to an unexpected advertisement page. The page contains multiple candidate buttons and decoy elements, and only one button leads back to successful task completion. The explicit drift is implemented by changing which button is correct across drift phases. This setting tests whether the agent can revise stale navigational memory after the page structure changes.

\begin{figure}[!htbp]
    \centering
    \begin{subfigure}[b]{0.30\textwidth}
        \centering
        \includegraphics[width=\linewidth]{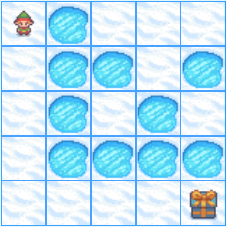}
        \caption{Source}
        \label{fig:frozen_source}
    \end{subfigure}
    \hfill
    \begin{subfigure}[b]{0.30\textwidth}
        \centering
        \includegraphics[width=\linewidth]{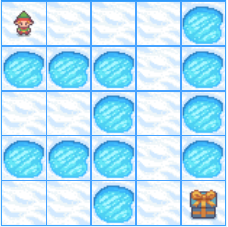}
        \caption{Drift I}
        \label{fig:frozen_drift1}
    \end{subfigure}
    \hfill
    \begin{subfigure}[b]{0.30\textwidth}
        \centering
        \includegraphics[width=\linewidth]{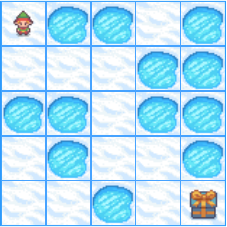}
        \caption{Drift II}
        \label{fig:frozen_drift2}
    \end{subfigure}    
    \caption{FrozenLake explicit drift.}
    \label{fig:frozenlake_drifts}
\end{figure}

\paragraph{FrozenLake.}
As shown in Fig.~\ref{fig:frozenlake_drifts}, the explicit drift in FrozenLake is introduced by modifying the obstacle layout. The source environment and the two drifted environments share the same task objective, but the feasible paths differ because the positions of holes change. This setting tests whether the agent can invalidate outdated topological memory and recover a valid path in the new map.

\begin{figure}[!htbp]
    \centering
    \includegraphics[width=0.50\textwidth]{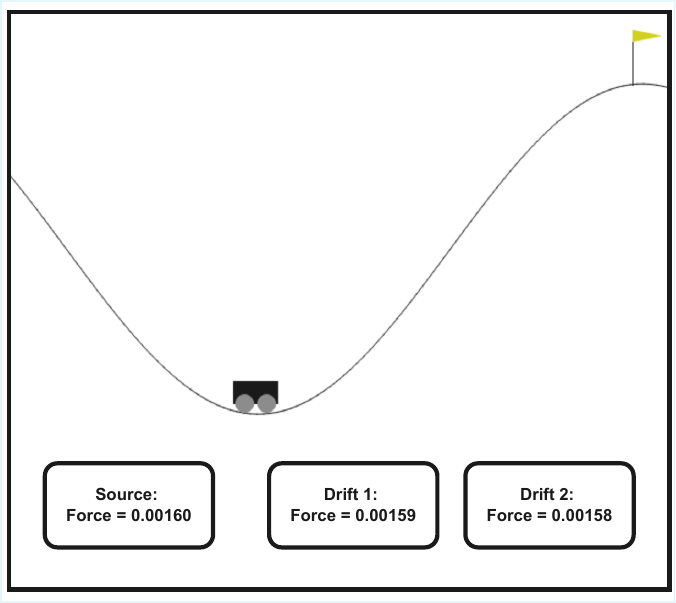}
    \caption{MountainCar explicit drift.}
    \label{fig:mountaincar_drift}
\end{figure}

\paragraph{MountainCar.}
As shown in Fig.~\ref{fig:mountaincar_drift}, the explicit drift in MountainCar is implemented by decreasing the engine force parameter from the source environment to the drift phases. This change alters the transition dynamics while preserving the observation and action spaces. The agent must therefore adapt its control behavior rather than reuse a policy learned under the original dynamics.

\subsection{Implicit Environmental Drifts}
\label{subsec:implicit_env_drifts}

We construct implicit environmental drifts by changing hidden task logic while keeping the surface observation largely unchanged. Unlike explicit drift, these changes are not directly visible from the state description or interface. The agent can identify them only through interaction feedback, such as reward or task success. These settings test whether an agent can stop trusting stale memory when the environment appears unchanged but the optimal behavior has shifted.

\begin{figure}[!htbp]
    \centering
    \includegraphics[width=0.95\textwidth]{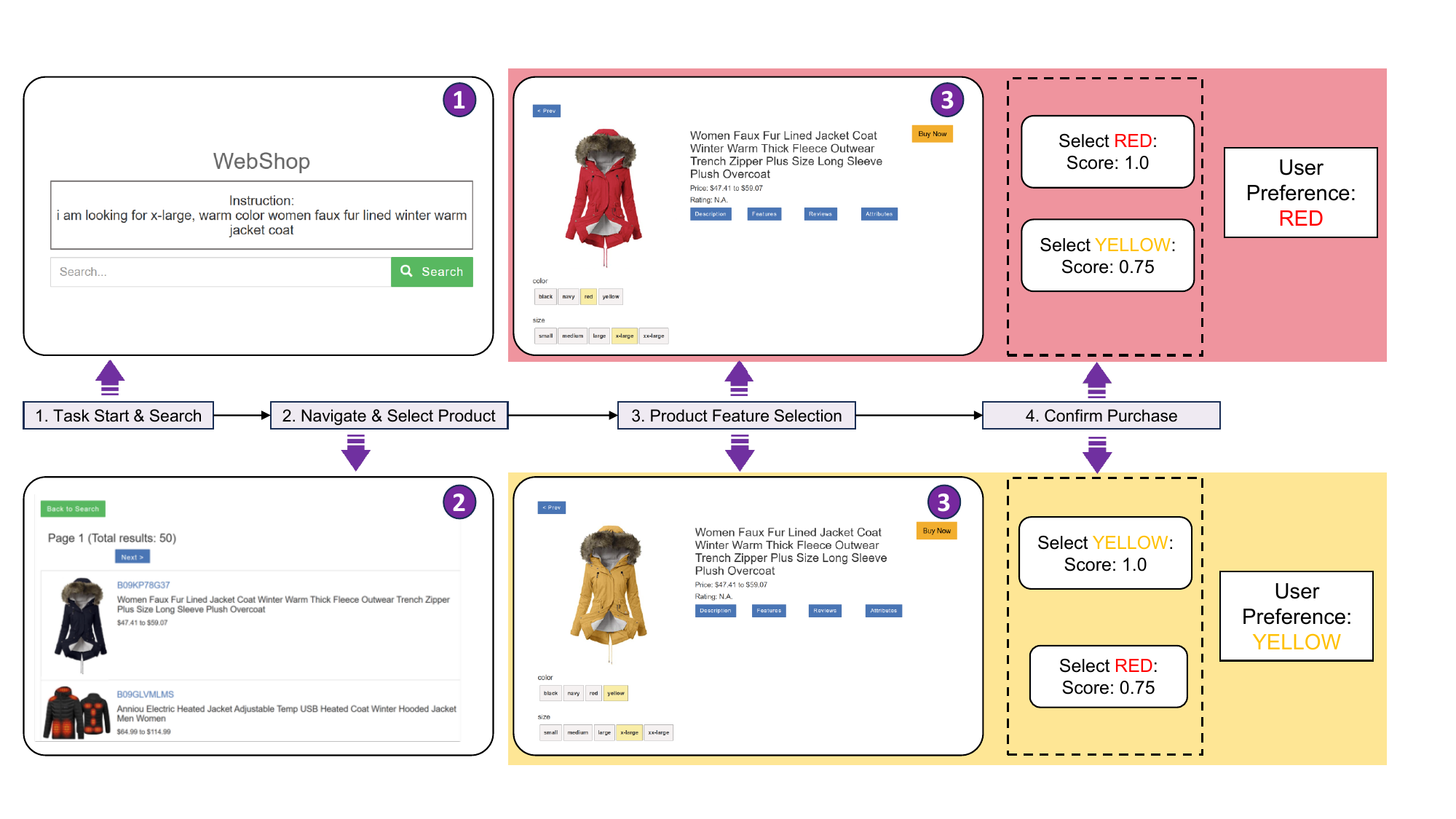}
    \caption{WebShop implicit drift.}
    \label{fig:webshop_implicit}
\end{figure}

\paragraph{WebShop.}
As shown in Fig.~\ref{fig:webshop_implicit}, the implicit WebShop drift changes the semantic grounding of the attribute ``warm color.'' In the source environment, the correct grounding is Yellow, while in the drifted environment it becomes Red. The instruction and interface remain the same, but the reward feedback changes according to the new semantic rule. This setting tests whether the agent can update semantic memory based on feedback rather than relying on the previous attribute mapping.

\begin{figure}[!htbp]
    \centering
    \includegraphics[width=0.7\textwidth]{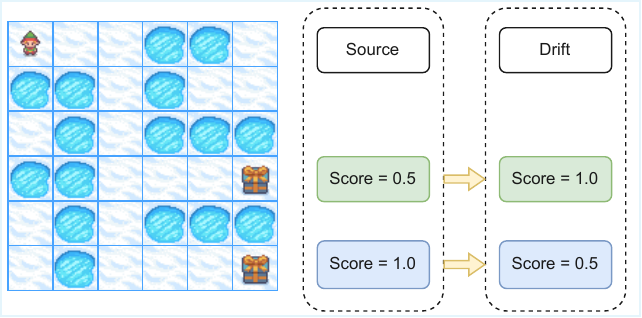}
    \caption{FrozenLake implicit drift.}
    \label{fig:frozenlake_implicit}
\end{figure}

\paragraph{FrozenLake.}
As shown in Fig.~\ref{fig:frozenlake_implicit}, the implicit FrozenLake drift keeps the map structure fixed but changes the reward values associated with goal tiles. The goal that was optimal in the source environment becomes a suboptimal trap after drift, while another goal becomes the new global optimum. This setting tests whether the agent can suppress a retrieved policy that still appears feasible but no longer maximizes reward.

\subsection{JetRover Real-Robot Setup}
\label{app:jetrover_setup}

We provide additional details for the real-robot evaluation in Section~\ref{sec:real_robot}. The goal of this experiment is not to train a new robot policy, but to test whether GLOVE can detect and correct stale memory in a replay-relaxed physical environment. We therefore keep the task deliberately simple and use the same deployed robot stack throughout.

\paragraph{Platform and models.}
We use a JetRover mobile robot in a small indoor maze built from cardboard boxes. The language reasoning module is Qwen2.5-Max, and visual grounding is provided by Qwen2.5-VL. No task-specific training or finetuning is performed. The robot is given a simple instruction, \emph{find the blue block}, and must navigate to the target region and verify the object identity from onboard visual observations.

\begin{figure}[t]
    \centering
    \begin{subfigure}[b]{0.24\textwidth}
        \centering
        \includegraphics[width=\linewidth,height=0.26\textheight,keepaspectratio]{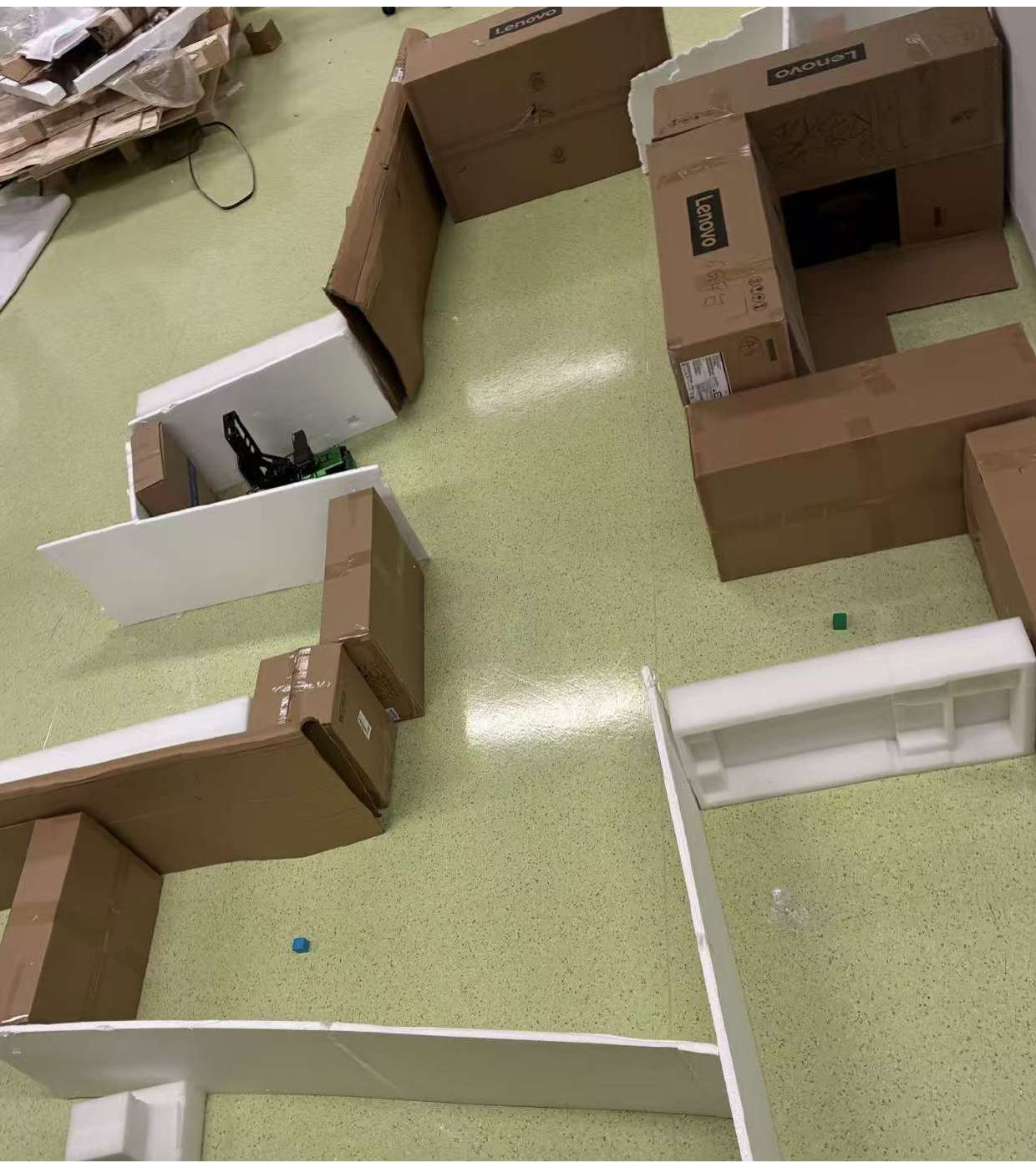}
        \caption{Real-world setup}
        \label{fig:jetrover_real_map}
    \end{subfigure}
    \hfill
    \begin{subfigure}[b]{0.24\textwidth}
        \centering
        \includegraphics[width=\linewidth,height=0.26\textheight,keepaspectratio]{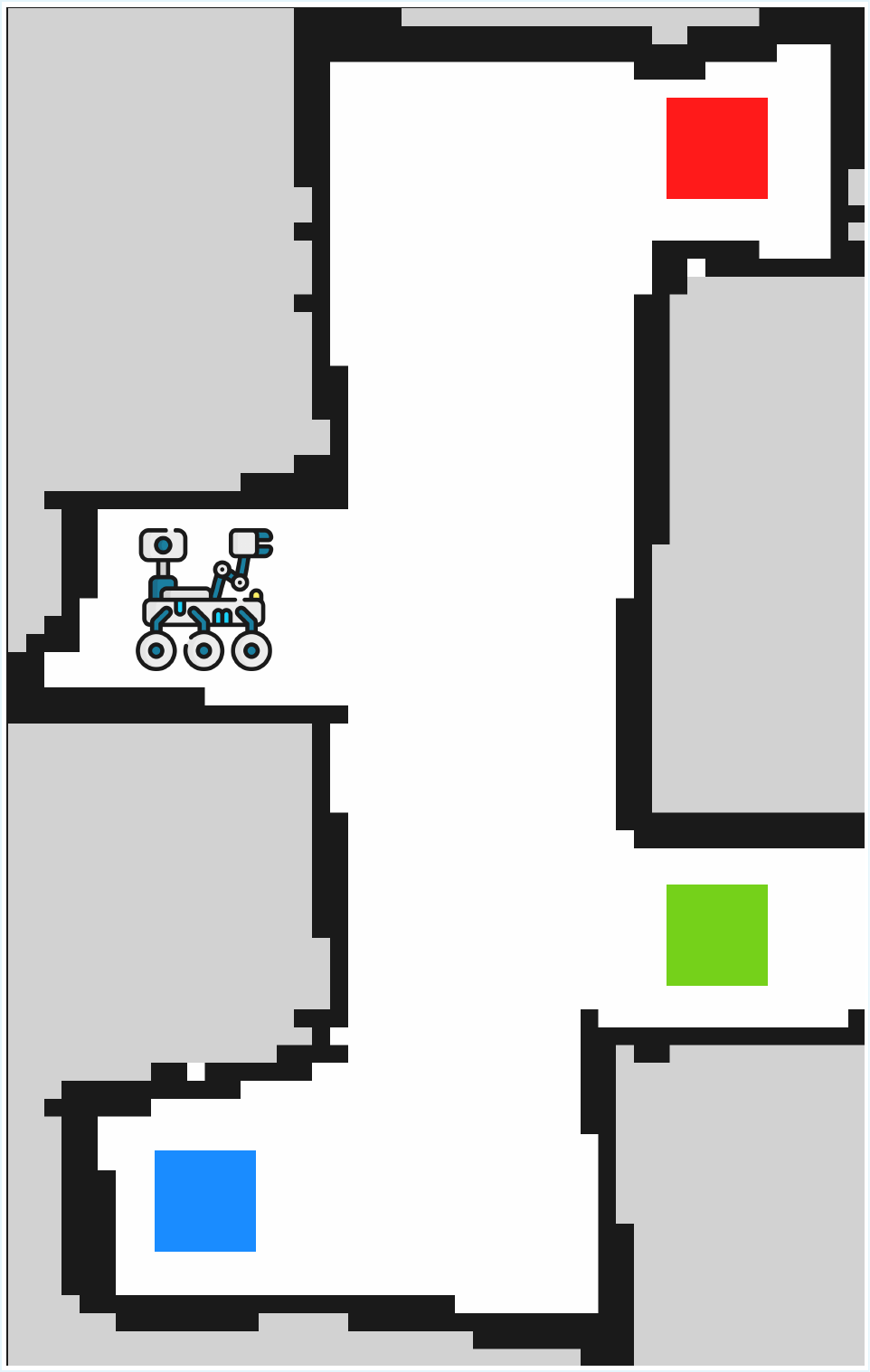}
        \caption{Source environment}
        \label{fig:jetrover_map_source}
    \end{subfigure}
    \hfill
    \begin{subfigure}[b]{0.24\textwidth}
        \centering
        \includegraphics[width=\linewidth,height=0.26\textheight,keepaspectratio]{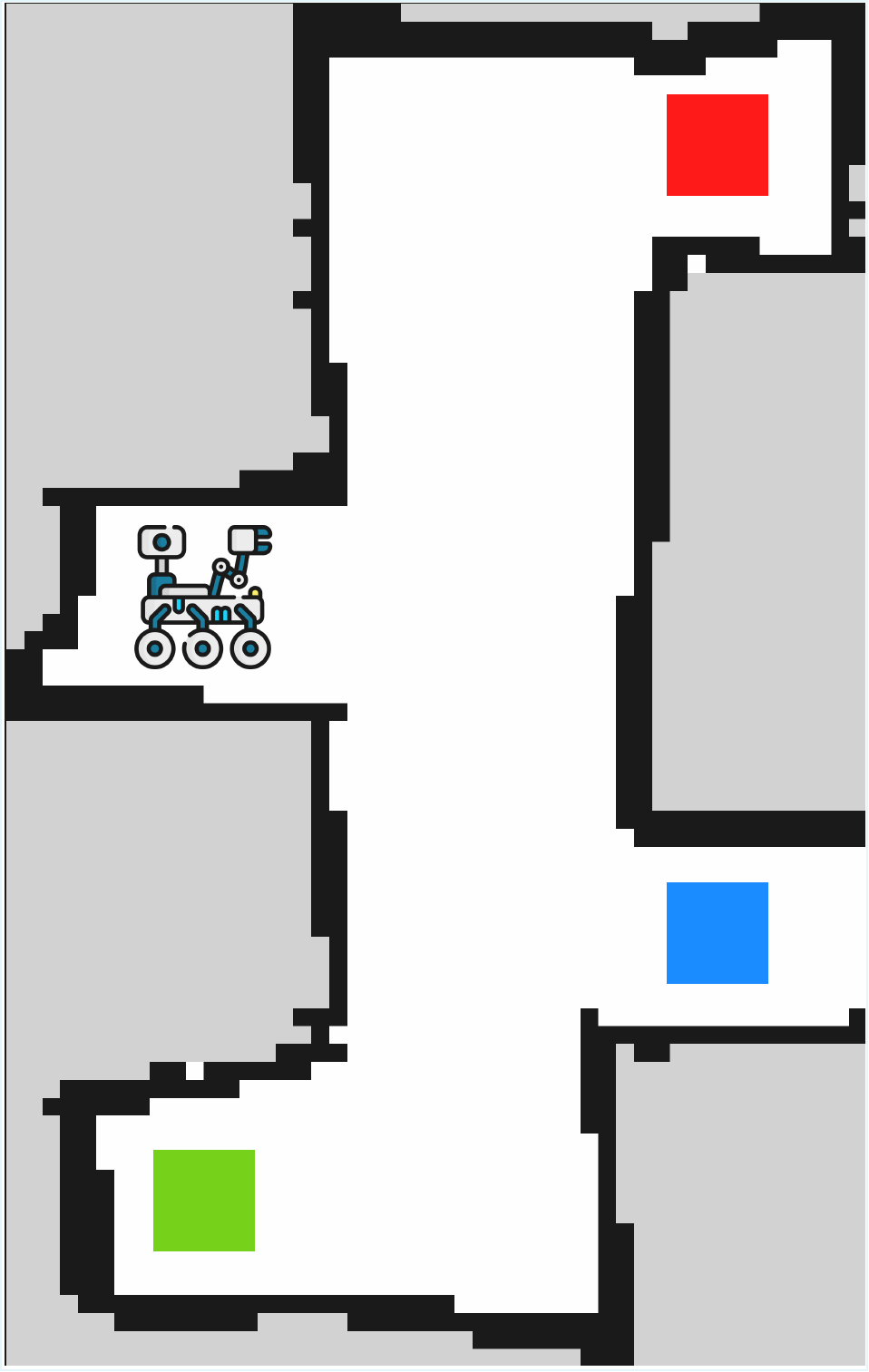}
        \caption{Drift 1: Cube swap}
        \label{fig:jetrover_map_drift1}
    \end{subfigure}
    \caption{\textbf{JetRover real-world layout and map drift.} Left: the physical cardboard-maze setup used in the real-robot experiment. Middle: the source environment with three colored targets. Right: Drift~1, where the \texttt{blue} and \texttt{green} blocks exchange positions while the rest of the layout remains unchanged.}
    \label{fig:jetrover_maps}
\end{figure}

\paragraph{Task and drift protocol.}
The environment contains three semantic targets: \texttt{red}, \texttt{green}, and \texttt{blue}. In the source environment, the robot first builds memory for the location of the \texttt{blue} block. We then introduce drift by swapping the positions of the \texttt{blue} and \texttt{green} blocks, while keeping the rest of the layout unchanged. 
Figure~\ref{fig:jetrover_maps} shows the real-world layout together with the abstract source map and the drifted map used in evaluation.
This setup creates a minimal but concrete stale-memory failure mode: after drift, a policy that trusts static memory will continue navigating to the old \texttt{blue} location and encounter the \texttt{green} block instead.

\paragraph{Compared methods.}
We compare three variants built on the same platform and navigation stack. \emph{No Memory} searches without retrieved target-location memory. \emph{Static Memory} always revisits the previously memorized location of the \texttt{blue} block. \emph{GLOVE} uses the same memory but verifies the retrieved target against current visual evidence and asynchronously updates stale memory after a mismatch. This comparison isolates the effect of memory verification and realignment, rather than changes in low-level navigation or perception.

\begin{figure}[t]
    \centering
    \begin{subfigure}[b]{0.47\textwidth}
        \centering
        \includegraphics[width=\linewidth]{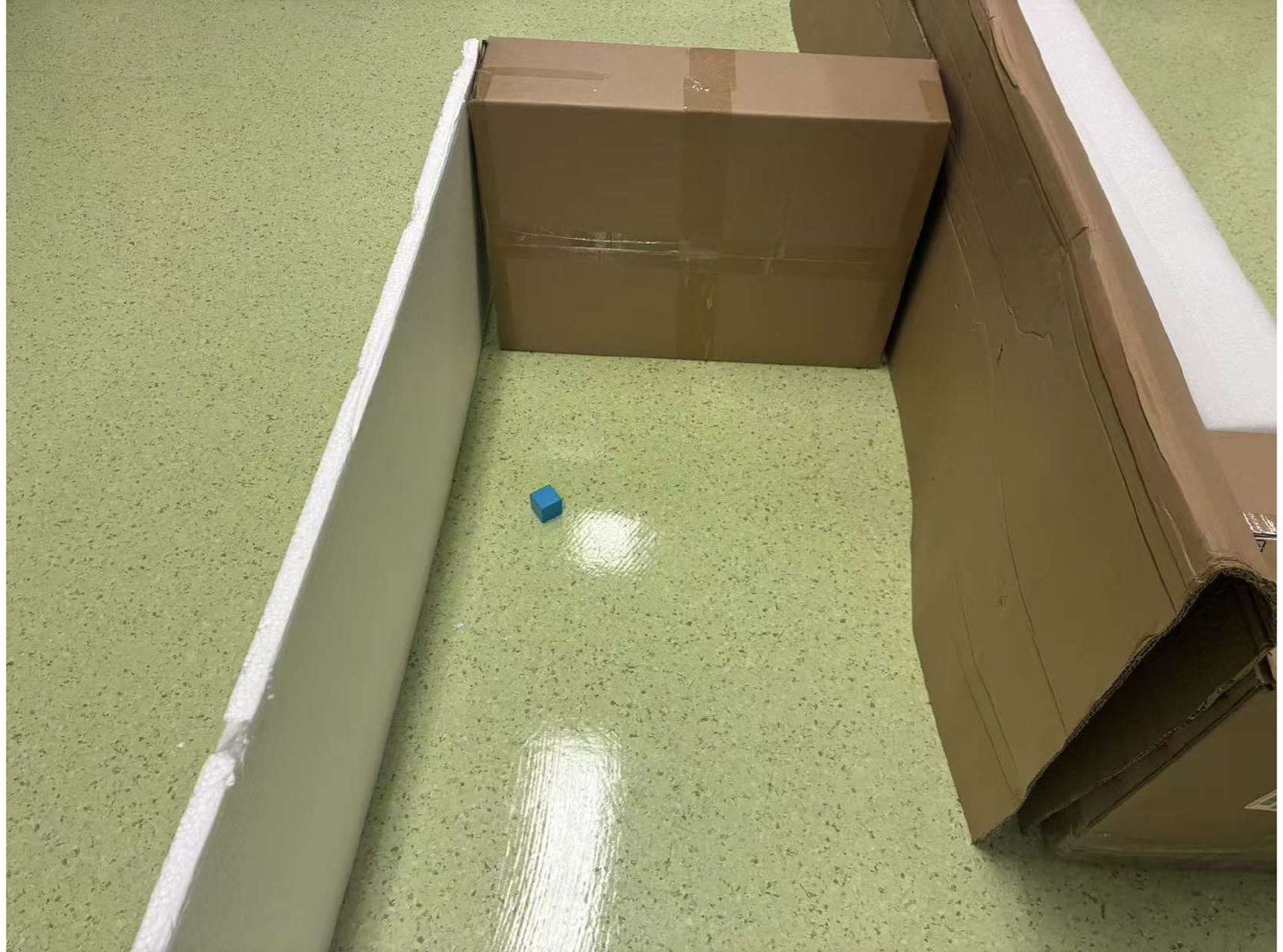}
        \caption{Blue block shifted left}
        \label{fig:jetrover_blue_left}
    \end{subfigure}
    \hfill
    \begin{subfigure}[b]{0.47\textwidth}
        \centering
        \includegraphics[width=\linewidth]{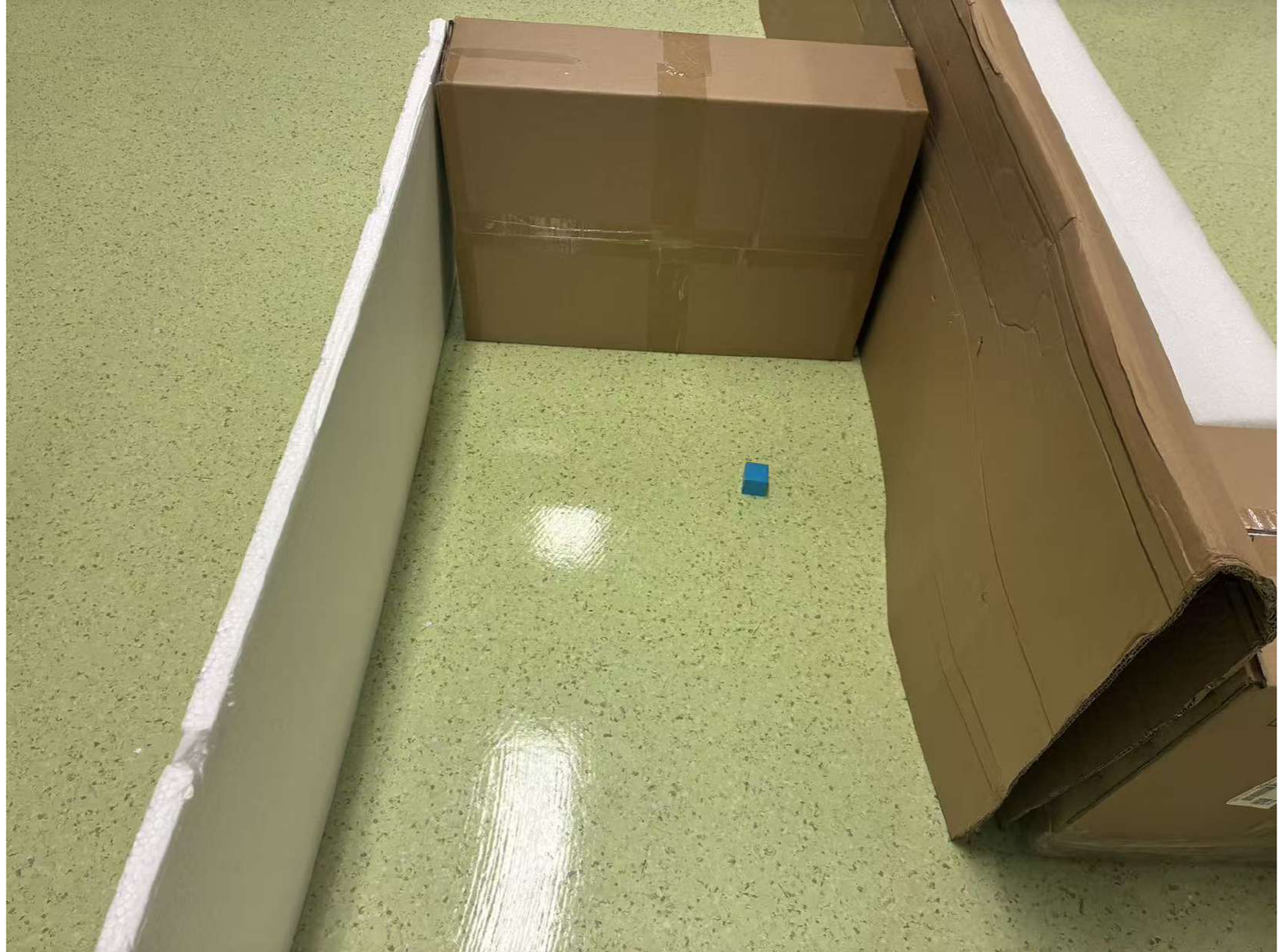}
        \caption{Blue block shifted right}
        \label{fig:jetrover_blue_right}
    \end{subfigure}
    \caption{\textbf{Robustness to small target displacement.} Small position changes of the \texttt{blue} block within the same local region do not change the semantic target and do not prevent recognition.}
    \label{fig:jetrover_blue_shift}
\end{figure}

\begin{figure}[t]
    \centering
    \begin{subfigure}[b]{0.47\textwidth}
        \centering
        \includegraphics[width=\linewidth]{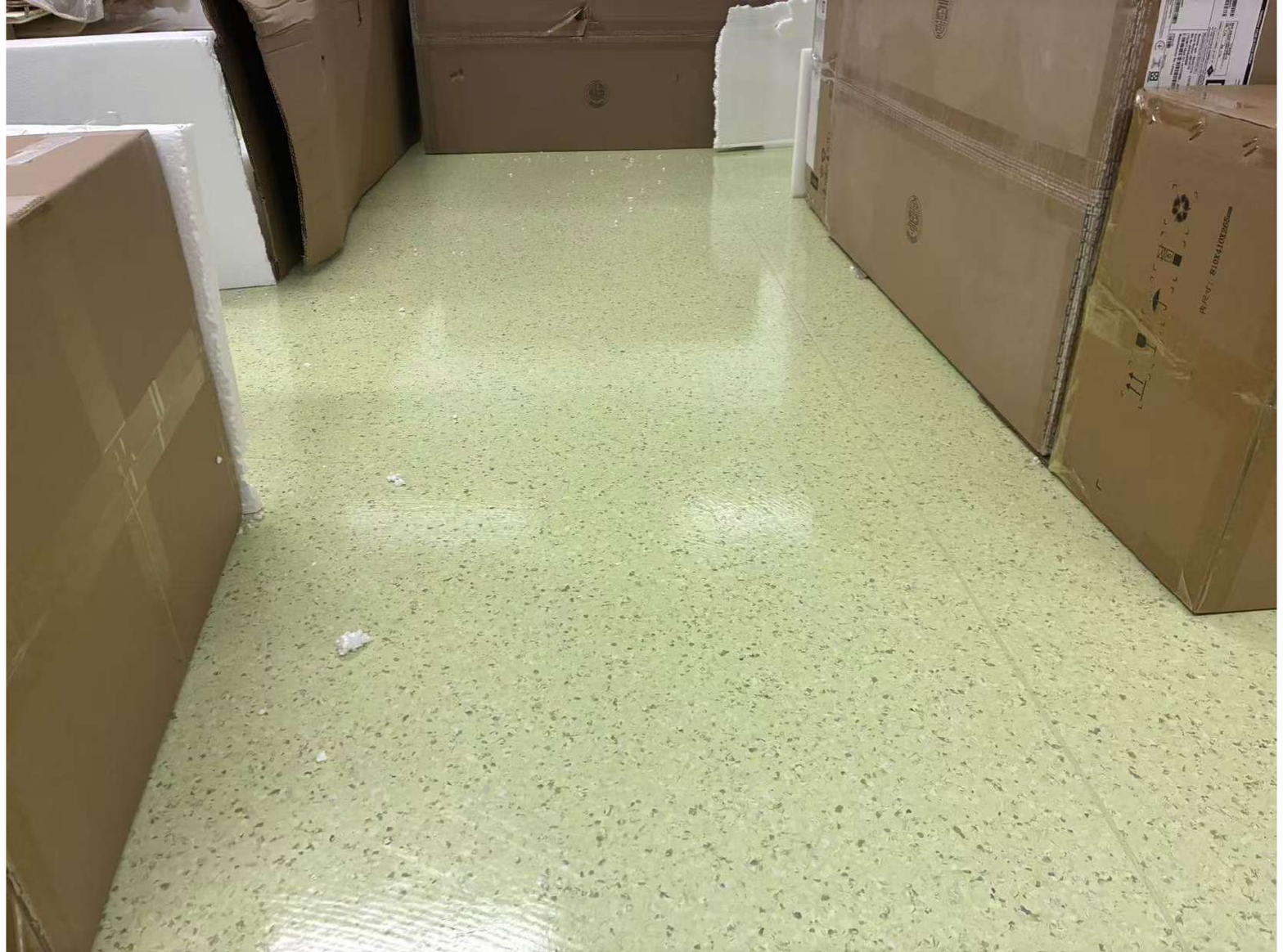}
        \caption{Corridor without extra obstacle}
        \label{fig:jetrover_corridor}
    \end{subfigure}
    \hfill
    \begin{subfigure}[b]{0.47\textwidth}
        \centering
        \includegraphics[width=\linewidth]{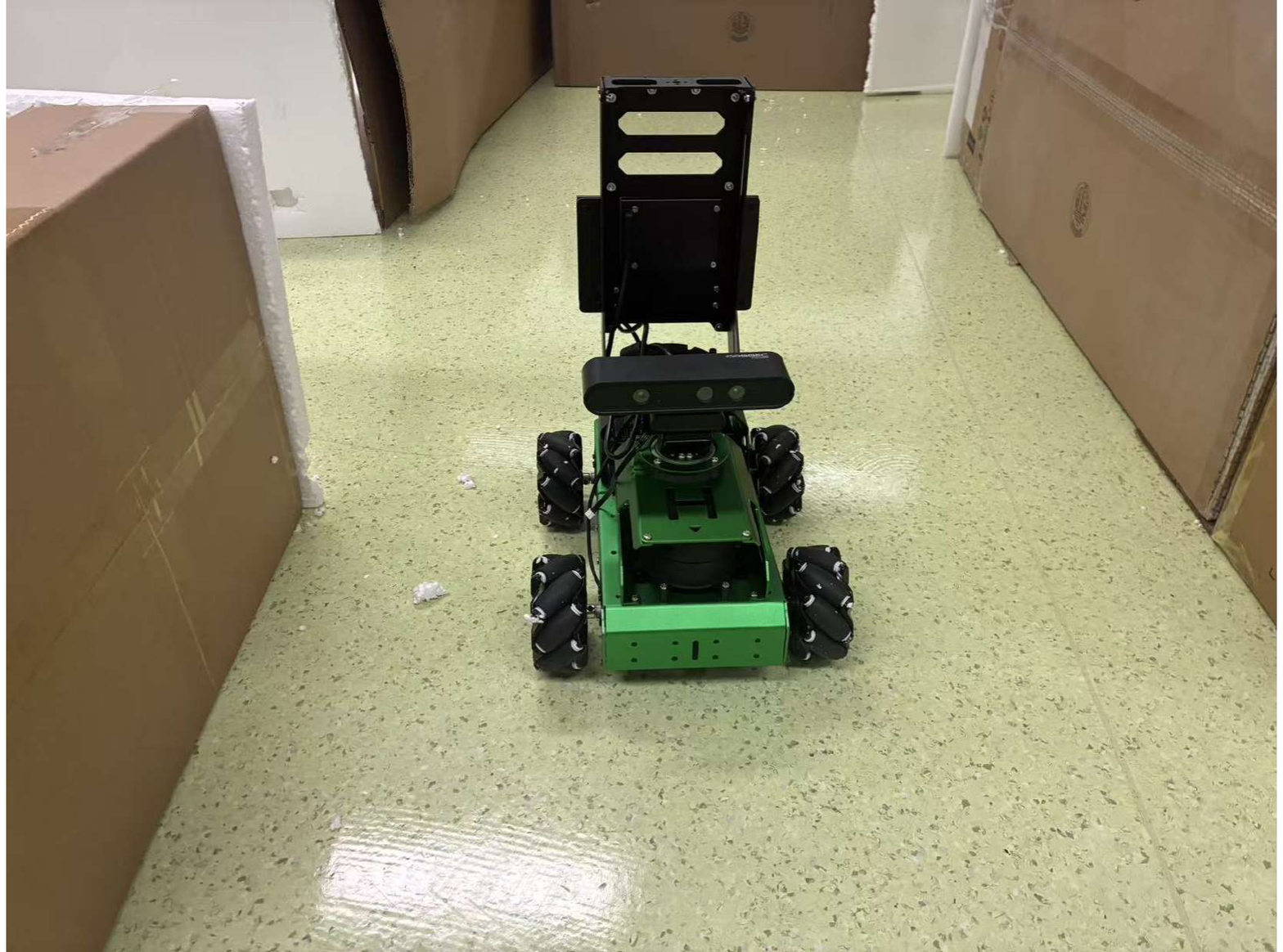}
        \caption{Successful passing without obstacle}
        \label{fig:jetrover_pass_noblock}
    \end{subfigure}

    \vspace{0.5em}

    \begin{subfigure}[b]{0.47\textwidth}
        \centering
        \includegraphics[width=\linewidth]{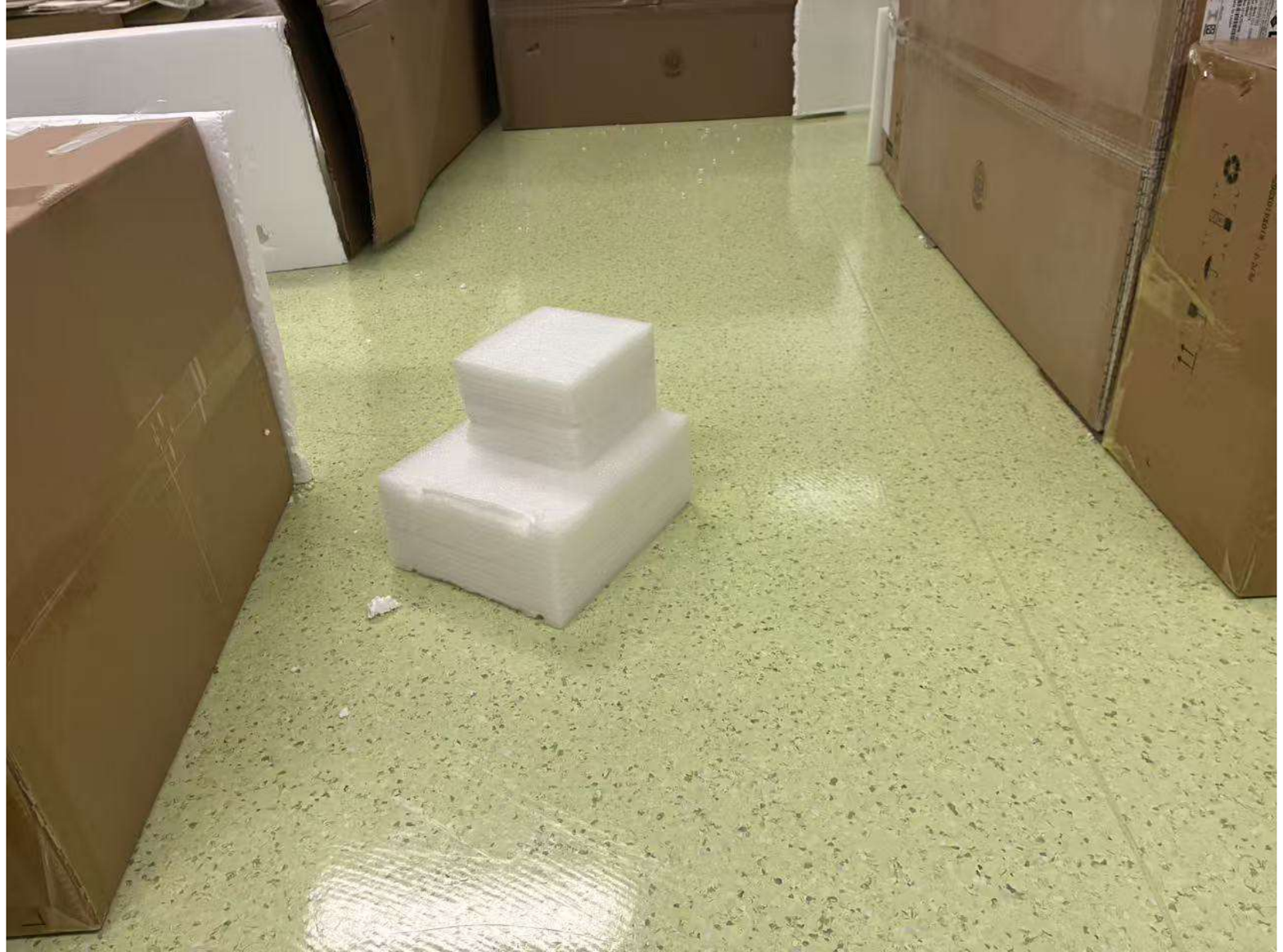}
        \caption{Corridor with an added obstacle}
        \label{fig:jetrover_corridor_block}
    \end{subfigure}
    \hfill
    \begin{subfigure}[b]{0.47\textwidth}
        \centering
        \includegraphics[width=\linewidth]{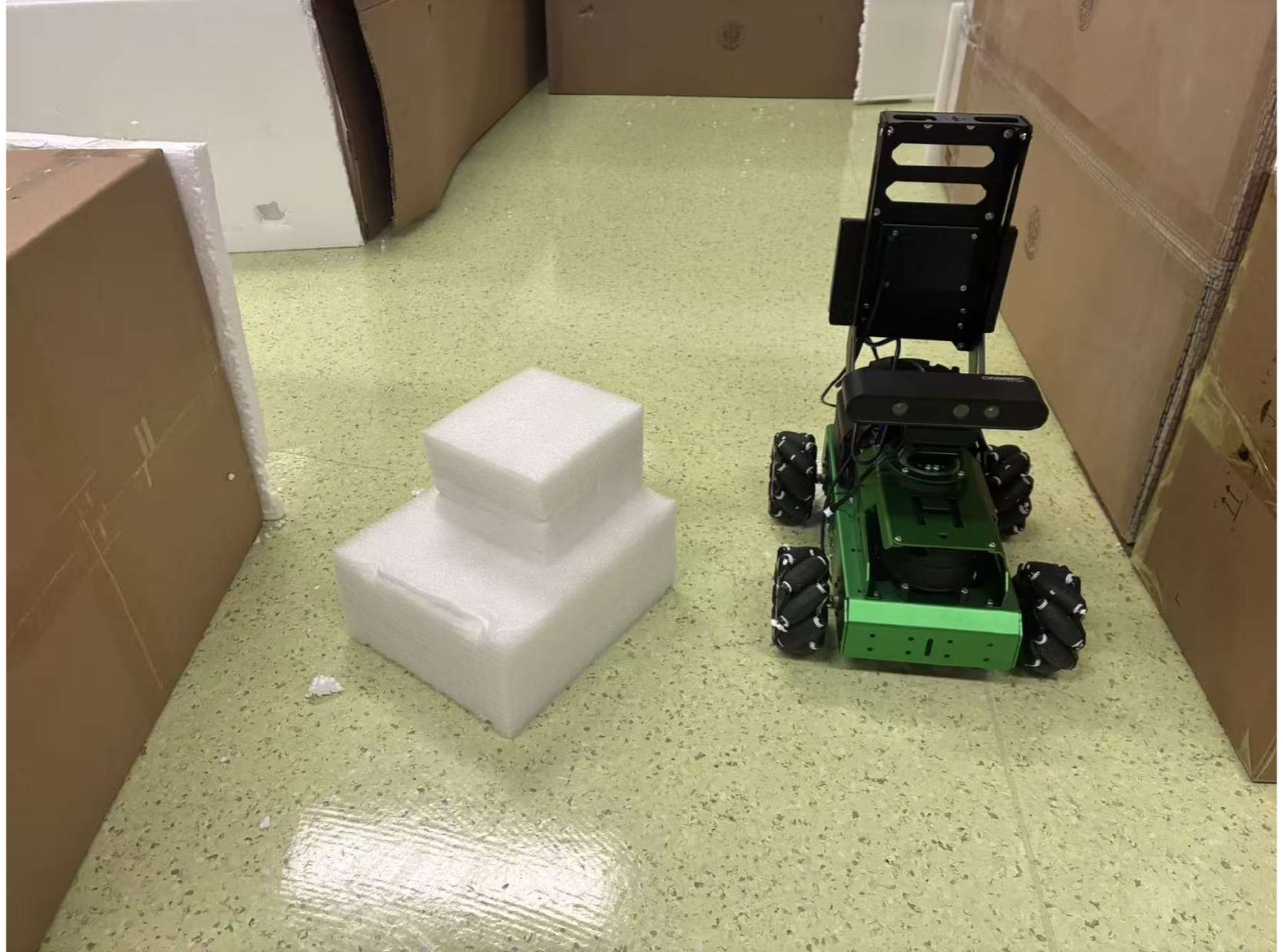}
        \caption{Successful passing with obstacle}
        \label{fig:jetrover_pass_block}
    \end{subfigure}
    \caption{\textbf{Robustness to path-level noise.} We add an extra obstacle in the corridor to perturb the path geometry. The robot can still traverse the corridor and reach the inspection region, indicating that local navigation noise does not by itself induce stale-memory failure.}
    \label{fig:jetrover_corridor_noise}
\end{figure}

\paragraph{Physical noise and robustness.}
The maze is intentionally imperfect: the cardboard walls are irregular, local paths may be partially blocked, and the target block may shift by a few centimeters between runs. These changes do not alter the semantic target identity, but they introduce realistic deployment-time noise. Figure~\ref{fig:jetrover_blue_shift} shows that small left-right target displacement does not prevent the robot from recognizing the \texttt{blue} block. Figure~\ref{fig:jetrover_corridor_noise} shows that the robot can still pass through the corridor and reach the target region even when an additional obstacle is inserted.

\paragraph{Interpretation.}
This setup produces a clean separation between true semantic drift and incidental physical noise. The swap between the \texttt{blue} and \texttt{green} blocks changes the correct target location and therefore invalidates previously stored memory. By contrast, centimeter-scale target displacement or temporary path obstruction changes only the local geometry of the run, not the semantic identity of the target. In our experiments, \emph{Static Memory} fails under the former case because it continues to trust stale target-location memory, whereas \emph{GLOVE} uses asynchronous verification to revise memory and recover the correct target.

\subsection{Case Study: FrozenLake Implicit Environmental Drifts}
\label{subsec:case_study}
When the agent enters block (3,2):
\begin{figure}[H]
    \centering
    \includegraphics[width=0.5\textwidth]{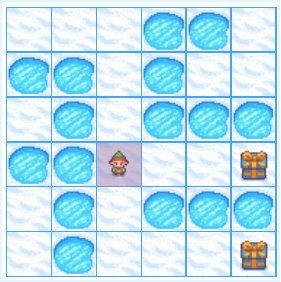}
    \label{fig:frozenlake_implicit_decision_point}
\end{figure}

Right after the drift, the agent still has experience from the source. The agent's prompt at the time looks like:

\begin{verbatim}
<|begin_of_sentence|>
You are an intelligent exploration agent navigating a frozen lake. 
Your goal is to reach the destination with highest possible score while 
avoiding holes. Analyze the current position and decide the next move. 
Respond with only the action number (0, 1, 2, or 3) without any additional 
explanation or formatting.

Map Size: 6 rows x 6 columns
Current Position: [3, 2]

There are a total of 2 destination(s).
Destinations: (3, 5), (5, 5)

......

--- Historical Experience from Similar States ---
You have been at this position before. Here are 2 previous experience(s):

  Action taken: 2
  Result Position: [3, 3]
  Result Tile: F
  Max score achievable from this path: 0.5

  Action taken: 1
  Result Position: [4, 2]
  Result Tile: F
  Max score achievable from this path: 1.0

......

Which action effectively moves you closer to the HIGHEST SCORE goal? 
Respond with only the action number (0, 1, 2, or 3).
\end{verbatim}

Through concatenating experiences' $s'$ with other experiences whose $s$ is the same, the agent constructs its internal world model and therefore realizes that the max score action 2 and action 1 can lead to. 

The agent will proceed to move to the destination (3,5).GLOVE starts probing when a memory inconsistency is detected.

After the agent arrives at the (3,2) again with updated experience:

\begin{verbatim}
<|begin_of_sentence|>
You are an intelligent exploration agent navigating a frozen lake. 
Your goal is to reach the destination with highest possible score while 
avoiding holes. Analyze the current position and decide the next move. 
Respond with only the action number (0, 1, 2, or 3) without any additional 
explanation or formatting.

Map Size: 6 rows x 6 columns
Current Position: [3, 2]

There are a total of 2 destination(s).
Destinations: (3, 5), (5, 5)

......

--- Historical Experience from Similar States ---
You have been at this position before. Here are 2 previous experience(s):

  Action taken: 2
  Result Position: [3, 3]
  Result Tile: F
  Max score achievable from this path: 0.5

  Action taken: 1
  Result Position: [4, 2]
  Result Tile: F
  Max score achievable from this path: 0.5

......

Which action effectively moves you closer to the HIGHEST SCORE goal? 
Respond with only the action number (0, 1, 2, or 3).
\end{verbatim}

The agent now notices that neither path leads to a full score and starts exploring again. 

After some time, it reaches (5,5). GLOVE realizes the memory conflicts and starts probing again, re-align the internal world model in the experience base with the environment.

\begin{verbatim}
<|begin_of_sentence|>
You are an intelligent exploration agent navigating a frozen lake. 
Your goal is to reach the destination with highest possible score while 
avoiding holes. Analyze the current position and decide the next move. 
Respond with only the action number (0, 1, 2, or 3) without any additional 
explanation or formatting.

Map Size: 6 rows x 6 columns
Current Position: [3, 2]

There are a total of 2 destination(s).
Destinations: (3, 5), (5, 5)

......

--- Historical Experience from Similar States ---
You have been at this position before. Here are 2 previous experience(s):

  Action taken: 2
  Result Position: [3, 3]
  Result Tile: F
  Max score achievable from this path: 0.5

  Action taken: 1
  Result Position: [4, 2]
  Result Tile: F
  Max score achievable from this path: 1.0

......

Which action effectively moves you closer to the HIGHEST SCORE goal? 
Respond with only the action number (0, 1, 2, or 3).
\end{verbatim}

\section{Additional Experiment Results}
\label{appendix-experiment_results}

\subsection{Backbone Sweep Overview}
\label{appendix-backbone_sweep_overview}
\begin{table}[H]
\centering
\caption{Backbone sweep overview. Each cell reports the change in percentage caused by a drift.
The shaded rows show the performance of GLOVE-augmented agents. We annotate the performance gap relative to the baseline (e.g., \improv{56.3} indicates improvement).}
\label{tab:appendix_backbone_sweep}
\small
\setlength{\tabcolsep}{6pt}
\resizebox{\textwidth}{!}{%
\begin{tabular}{l c c c c c}
\toprule
& \multicolumn{3}{c}{\textbf{Explicit Structural Drift (Drift II)}} & \multicolumn{2}{c}{\textbf{Implicit Hidden Drift}} \\
\cmidrule(lr){2-4} \cmidrule(lr){5-6}
\textbf{Backbone}
& \textbf{WebShop}
& \textbf{FrozenLake}
& \textbf{MountainCar}
& \textbf{WebShop}
& \textbf{FrozenLake} \\
\midrule
Llama3.1-8B       & 7.5 $\rightarrow$ 5   & 0   $\rightarrow$ 1.2 & 93.8 $\rightarrow$ 65   & 34.4 $\rightarrow$ 36.2 & 13.1 $\rightarrow$ 52.5 \\
\rowcolor{gray!10}
\textbf{+GLOVE}    & 67.5 $\rightarrow$ 75   \improv{70  } & 36.2 $\rightarrow$ 67.5 \improv{66.3} & 93.8 $\rightarrow$ 68.8 \improv{3.8} & 42.5 $\rightarrow$ 61.6 \improv{25.4} & 12.5 $\rightarrow$ 87.5 \improv{35  } \\
\midrule
Llama3.3-70B      & 8.8 $\rightarrow$ 7.5 & 83.7 $\rightarrow$ 0   & 0   $\rightarrow$ 0   & 95.3 $\rightarrow$ 76.6 & 39.4 $\rightarrow$ 96.2 \\
\rowcolor{gray!10}
\textbf{+GLOVE}    & 95   $\rightarrow$ 88.8 \improv{81.3} & 82.5 $\rightarrow$ 73.8 \improv{73.8} & 0   $\rightarrow$ 0   & 100   $\rightarrow$ 83.4 \improv{6.8} & 40   $\rightarrow$ 97.5 \improv{1.3} \\
\midrule
Qwen2.5-7B        & 7.5 $\rightarrow$ 3.8 & 6.2 $\rightarrow$ 0   & 0   $\rightarrow$ 0   & 35.6 $\rightarrow$ 69.7 & 40.6 $\rightarrow$ 97.5 \\
\rowcolor{gray!10}
\textbf{+GLOVE}    & 63.7 $\rightarrow$ 71.2 \improv{67.4} & 78.8 $\rightarrow$ 68.8 \improv{68.8} & 0   $\rightarrow$ 0   & 46.9 $\rightarrow$ 75   \improv{5.3} & 41.9 $\rightarrow$ 100   \improv{2.5} \\
\midrule
Qwen3-30B         & 8.8 $\rightarrow$ 7.5 & 80   $\rightarrow$ 13.8 & 100   $\rightarrow$ 100   & 75   $\rightarrow$ 100   & 88.8 $\rightarrow$ 46.9 \\
\rowcolor{gray!10}
\textbf{+GLOVE}    & 95   $\rightarrow$ 86.2 \improv{78.7} & 80   $\rightarrow$ 75   \improv{61.2} & 100   $\rightarrow$ 100   & 75   $\rightarrow$ 100   & 85   $\rightarrow$ 47.5 \improv{0.6} \\
\midrule
GPT-4o            & 5   $\rightarrow$ 5   & 3.8 $\rightarrow$ 11.2 & 96.2 $\rightarrow$ 90   & 99.4 $\rightarrow$ 76.6 & 60   $\rightarrow$ 51.9 \\
\rowcolor{gray!10}
\textbf{+GLOVE}    & 90   $\rightarrow$ 92.5 \improv{87.5} & 77.5 $\rightarrow$ 76.2 \improv{65  } & 100   $\rightarrow$ 100   \improv{10  } & 98.4 $\rightarrow$ 91.6 \improv{15  } & 51.2 $\rightarrow$ 96.2 \improv{44.3} \\
\midrule
Grok-3            & 8.8 $\rightarrow$ 7.5 & 60   $\rightarrow$ 11.2 & 100   $\rightarrow$ 100   & 96.2 $\rightarrow$ 78.4 & 35.6 $\rightarrow$ 68.1 \\
\rowcolor{gray!10}
\textbf{+GLOVE}    & 87.5 $\rightarrow$ 93.8 \improv{86.3} & 75   $\rightarrow$ 71.2 \improv{60  } & 100   $\rightarrow$ 100   & 95   $\rightarrow$ 98.4 \improv{20  } & 33.8 $\rightarrow$ 98.8 \improv{30.7} \\
\midrule
DeepSeek-V3.2     & 5   $\rightarrow$ 6.2 & 0   $\rightarrow$ 0   & 0   $\rightarrow$ 0   & 95.3 $\rightarrow$ 76.9 & 26.9 $\rightarrow$ 75.6 \\
\rowcolor{gray!10}
\textbf{+GLOVE}    & 85   $\rightarrow$ 93.8 \improv{87.6} & 68.8 $\rightarrow$ 70   \improv{70  } & 0   $\rightarrow$ 0   & 94.4 $\rightarrow$ 97.5 \improv{20.6} & 27.5 $\rightarrow$ 95.6 \improv{20  } \\
\bottomrule
\end{tabular}%
}
\end{table}

\subsection{Experiment Results: Explicit Environmental Drifts}
\label{appendix-exp_result_explicit}

\begin{table}[H]
\centering
\caption{\textbf{GPT-4o: GLOVE's robustness to environmental drift (Explicit).} Success rate in percent. The shaded rows show the performance of GLOVE-augmented agents. We annotate the performance gap relative to the baseline (e.g., \textcolor{cyan!60!green}{\scriptsize($\uparrow$56.3)} indicates improvement).}
\label{tab:plugplay_memdec_style_GPT_4o_explicit}
\footnotesize
\setlength{\tabcolsep}{3.5pt} 
\renewcommand{\arraystretch}{1.1}

\resizebox{1.0\textwidth}{!}{%
\begin{tabular}{l|ccc|ccc|ccc}
\toprule
\multicolumn{10}{c}{\textbf{\textsc{Verifier-visible structural drift}}} \\
\midrule
& \multicolumn{3}{c|}{\textbf{WebShop} (Semantic Shift)}
& \multicolumn{3}{c|}{\textbf{FrozenLake} (topology shift)}
& \multicolumn{3}{c}{\textbf{MountainCar} (dynamics shift)} \\
\cmidrule(lr){2-4}\cmidrule(lr){5-7}\cmidrule(lr){8-10}
\textbf{Method} & \textbf{Source} & \textbf{$\to$ Drift I} & \textbf{$\to$ Drift II}
& \textbf{Source} & \textbf{$\to$ Drift I} & \textbf{$\to$ Drift II}
& \textbf{Source} & \textbf{$\to$ Drift I} & \textbf{$\to$ Drift II} \\
\midrule

\text{No Memory (Plain)} & 15 & 5 & 0 & 0 & 5 & 0 & 85 & 75 & 55 \\
\midrule
\text{Vanilla} & 85 & 0 & 0 & 85 & 0 & 0 & 100 & 100 & 90 \\
\rowcolor{gray!10}
\hspace{0.9em}\textbf{+GLOVE} & 85 & 85 \improv{85} & 95 \improv{95} & 80 & 75 \improv{75} & 80 \improv{80} & 100 & 100 & 100 \improv{10} \\
\addlinespace[2pt]
\text{MemoryBank} & 85 & 20 & 20 & 80 & 0 & 45 & 95 & 100 & 100 \\
\rowcolor{gray!10}
\hspace{0.9em}\textbf{+GLOVE} & 100 & 90 \improv{70} & 85 \improv{65} & 80 & 70 \improv{70} & 65 \improv{20} & 95 & 100 & 100 \\
\addlinespace[2pt]
\text{Voyager} & 85 & 0 & 0 & 85 & 0 & 0 & 100 & 90 & 100 \\
\rowcolor{gray!10}
\hspace{0.9em}\textbf{+GLOVE} & 80 & 90 \improv{90} & 95 \improv{95} & 85 & 85 \improv{85} & 75 \improv{75} & 100 & 100 \improv{10} & 100 \\
\addlinespace[2pt]
\text{Generative Agent} & 80 & 0 & 0 & 80 & 15 & 0 & 100 & 95 & 70 \\
\rowcolor{gray!10}
\hspace{0.9em}\textbf{+GLOVE} & 75 & 95 \improv{95} & 95 \improv{95} & 80 & 80 \improv{65} & 85 \improv{85} & 95 & 100 \improv{5} & 100 \improv{30} \\

\bottomrule
\end{tabular}%
}
\end{table}

\begin{table}[H]
\centering
\caption{\textbf{Grok-3: GLOVE's robustness to environmental drift (Explicit).} Success rate in percent. The shaded rows show the performance of GLOVE-augmented agents. We annotate the performance gap relative to the baseline (e.g., \textcolor{cyan!60!green}{\scriptsize($\uparrow$56.3)} indicates improvement).}
\label{tab:plugplay_memdec_style_Grok_3_explicit}
\footnotesize
\setlength{\tabcolsep}{3.5pt} 
\renewcommand{\arraystretch}{1.1}

\resizebox{1.0\textwidth}{!}{%
\begin{tabular}{l|ccc|ccc|ccc}
\toprule
\multicolumn{10}{c}{\textbf{\textsc{Verifier-visible structural drift}}} \\
\midrule
& \multicolumn{3}{c|}{\textbf{WebShop} (Semantic Shift)}
& \multicolumn{3}{c|}{\textbf{FrozenLake} (topology shift)}
& \multicolumn{3}{c}{\textbf{MountainCar} (dynamics shift)} \\
\cmidrule(lr){2-4}\cmidrule(lr){5-7}\cmidrule(lr){8-10}
\textbf{Method} & \textbf{Source} & \textbf{$\to$ Drift I} & \textbf{$\to$ Drift II}
& \textbf{Source} & \textbf{$\to$ Drift I} & \textbf{$\to$ Drift II}
& \textbf{Source} & \textbf{$\to$ Drift I} & \textbf{$\to$ Drift II} \\
\midrule

\text{No Memory (Plain)} & 20 & 5 & 15 & 25 & 0 & 0 & 100 & 90 & 95 \\
\midrule
\text{Vanilla} & 85 & 0 & 0 & 100 & 75 & 0 & 100 & 100 & 100 \\
\rowcolor{gray!10}
\hspace{0.9em}\textbf{+GLOVE} & 90 & 80 \improv{80} & 95 \improv{95} & 90 & 80 \improv{5} & 70 \improv{70} & 100 & 100 & 100 \\
\addlinespace[2pt]
\text{MemoryBank} & 90 & 35 & 30 & 95 & 70 & 0 & 100 & 100 & 100 \\
\rowcolor{gray!10}
\hspace{0.9em}\textbf{+GLOVE} & 85 & 90 \improv{55} & 95 \improv{65} & 95 & 65 \decline{5} & 80 \improv{80} & 100 & 100 & 100 \\
\addlinespace[2pt]
\text{Voyager} & 85 & 0 & 0 & 95 & 95 & 0 & 100 & 100 & 100 \\
\rowcolor{gray!10}
\hspace{0.9em}\textbf{+GLOVE} & 85 & 90 \improv{90} & 95 \improv{95} & 85 & 70 \decline{25} & 65 \improv{65} & 100 & 100 & 100 \\
\addlinespace[2pt]
\text{Generative Agent} & 85 & 0 & 0 & 95 & 0 & 45 & 100 & 100 & 100 \\
\rowcolor{gray!10}
\hspace{0.9em}\textbf{+GLOVE} & 85 & 90 \improv{90} & 90 \improv{90} & 95 & 85 \improv{85} & 70 \improv{25} & 100 & 100 & 100 \\

\bottomrule
\end{tabular}%
}
\end{table}

\begin{table}[H]
\centering
\caption{\textbf{DeepSeek-V3.2: GLOVE's robustness to environmental drift (Explicit).} Success rate in percent. The shaded rows show the performance of GLOVE-augmented agents. We annotate the performance gap relative to the baseline (e.g., \textcolor{cyan!60!green}{\scriptsize($\uparrow$56.3)} indicates improvement).}
\label{tab:plugplay_memdec_style_DeepSeek_V3_2_explicit}
\footnotesize
\setlength{\tabcolsep}{3.5pt} 
\renewcommand{\arraystretch}{1.1}

\resizebox{1.0\textwidth}{!}{%
\begin{tabular}{l|ccc|ccc|ccc}
\toprule
\multicolumn{10}{c}{\textbf{\textsc{Verifier-visible structural drift}}} \\
\midrule
& \multicolumn{3}{c|}{\textbf{WebShop} (Semantic Shift)}
& \multicolumn{3}{c|}{\textbf{FrozenLake} (topology shift)}
& \multicolumn{3}{c}{\textbf{MountainCar} (dynamics shift)} \\
\cmidrule(lr){2-4}\cmidrule(lr){5-7}\cmidrule(lr){8-10}
\textbf{Method} & \textbf{Source} & \textbf{$\to$ Drift I} & \textbf{$\to$ Drift II}
& \textbf{Source} & \textbf{$\to$ Drift I} & \textbf{$\to$ Drift II}
& \textbf{Source} & \textbf{$\to$ Drift I} & \textbf{$\to$ Drift II} \\
\midrule

\text{No Memory (Plain)} & 10 & 20 & 5 & 0 & 0 & 0 & 0 & 0 & 0 \\
\midrule
\text{Vanilla} & 85 & 0 & 0 & 95 & 0 & 0 & 0 & 0 & 0 \\
\rowcolor{gray!10}
\hspace{0.9em}\textbf{+GLOVE} & 100 & 70 \improv{70} & 95 \improv{95} & 85 & 75 \improv{75} & 75 \improv{75} & 0 & 0 & 0 \\
\addlinespace[2pt]
\text{MemoryBank} & 85 & 20 & 25 & 80 & 0 & 0 & 0 & 0 & 0 \\
\rowcolor{gray!10}
\hspace{0.9em}\textbf{+GLOVE} & 85 & 90 \improv{70} & 95 \improv{70} & 80 & 70 \improv{70} & 65 \improv{65} & 0 & 0 & 0 \\
\addlinespace[2pt]
\text{Voyager} & 85 & 0 & 0 & 95 & 0 & 0 & 0 & 0 & 0 \\
\rowcolor{gray!10}
\hspace{0.9em}\textbf{+GLOVE} & 75 & 90 \improv{90} & 95 \improv{95} & 90 & 65 \improv{65} & 70 \improv{70} & 0 & 0 & 0 \\
\addlinespace[2pt]
\text{Generative Agent} & 85 & 0 & 0 & 80 & 0 & 0 & 0 & 0 & 0 \\
\rowcolor{gray!10}
\hspace{0.9em}\textbf{+GLOVE} & 75 & 90 \improv{90} & 90 \improv{90} & 100 & 65 \improv{65} & 70 \improv{70} & 0 & 0 & 0 \\

\bottomrule
\end{tabular}%
}
\end{table}

\begin{table}[H]
\centering
\caption{\textbf{Llama3.3-70B: GLOVE's robustness to environmental drift (Explicit).} Success rate in percent. The shaded rows show the performance of GLOVE-augmented agents. We annotate the performance gap relative to the baseline (e.g., \textcolor{cyan!60!green}{\scriptsize($\uparrow$56.3)} indicates improvement).}
\label{tab:plugplay_memdec_style_Llama3_3_70B_explicit}
\footnotesize
\setlength{\tabcolsep}{3.5pt} 
\renewcommand{\arraystretch}{1.1}

\resizebox{1.0\textwidth}{!}{%
\begin{tabular}{l|ccc|ccc|ccc}
\toprule
\multicolumn{10}{c}{\textbf{\textsc{Verifier-visible structural drift}}} \\
\midrule
& \multicolumn{3}{c|}{\textbf{WebShop} (Semantic Shift)}
& \multicolumn{3}{c|}{\textbf{FrozenLake} (topology shift)}
& \multicolumn{3}{c}{\textbf{MountainCar} (dynamics shift)} \\
\cmidrule(lr){2-4}\cmidrule(lr){5-7}\cmidrule(lr){8-10}
\textbf{Method} & \textbf{Source} & \textbf{$\to$ Drift I} & \textbf{$\to$ Drift II}
& \textbf{Source} & \textbf{$\to$ Drift I} & \textbf{$\to$ Drift II}
& \textbf{Source} & \textbf{$\to$ Drift I} & \textbf{$\to$ Drift II} \\
\midrule

\text{No Memory (Plain)} & 50 & 50 & 0 & 0 & 0 & 0 & 0 & 0 & 0 \\
\midrule
\text{Vanilla} & 90 & 0 & 0 & 85 & 85 & 0 & 0 & 0 & 0 \\
\rowcolor{gray!10}
\hspace{0.9em}\textbf{+GLOVE} & 95 & 95 \improv{95} & 85 \improv{85} & 85 & 80 \decline{5} & 75 \improv{75} & 0 & 0 & 0 \\
\addlinespace[2pt]
\text{MemoryBank} & 100 & 35 & 30 & 90 & 85 & 0 & 0 & 0 & 0 \\
\rowcolor{gray!10}
\hspace{0.9em}\textbf{+GLOVE} & 100 & 95 \improv{60} & 85 \improv{55} & 85 & 85 & 75 \improv{75} & 0 & 0 & 0 \\
\addlinespace[2pt]
\text{Voyager} & 90 & 0 & 0 & 85 & 85 & 0 & 0 & 0 & 0 \\
\rowcolor{gray!10}
\hspace{0.9em}\textbf{+GLOVE} & 90 & 95 \improv{95} & 95 \improv{95} & 85 & 85 & 75 \improv{75} & 0 & 0 & 0 \\
\addlinespace[2pt]
\text{Generative Agent} & 95 & 0 & 0 & 85 & 80 & 0 & 0 & 0 & 0 \\
\rowcolor{gray!10}
\hspace{0.9em}\textbf{+GLOVE} & 90 & 95 \improv{95} & 90 \improv{90} & 85 & 80 & 70 \improv{70} & 0 & 0 & 0 \\

\bottomrule
\end{tabular}%
}
\end{table}

\begin{table}[H]
\centering
\caption{\textbf{Llama3.1-8B: GLOVE's robustness to environmental drift (Explicit).} Success rate in percent. The shaded rows show the performance of GLOVE-augmented agents. We annotate the performance gap relative to the baseline (e.g., \textcolor{cyan!60!green}{\scriptsize($\uparrow$56.3)} indicates improvement).}
\label{tab:plugplay_memdec_style_Llama3_1_8B_explicit}
\footnotesize
\setlength{\tabcolsep}{3.5pt} 
\renewcommand{\arraystretch}{1.1}

\resizebox{1.0\textwidth}{!}{%
\begin{tabular}{l|ccc|ccc|ccc}
\toprule
\multicolumn{10}{c}{\textbf{\textsc{Verifier-visible structural drift}}} \\
\midrule
& \multicolumn{3}{c|}{\textbf{WebShop} (Semantic Shift)}
& \multicolumn{3}{c|}{\textbf{FrozenLake} (topology shift)}
& \multicolumn{3}{c}{\textbf{MountainCar} (dynamics shift)} \\
\cmidrule(lr){2-4}\cmidrule(lr){5-7}\cmidrule(lr){8-10}
\textbf{Method} & \textbf{Source} & \textbf{$\to$ Drift I} & \textbf{$\to$ Drift II}
& \textbf{Source} & \textbf{$\to$ Drift I} & \textbf{$\to$ Drift II}
& \textbf{Source} & \textbf{$\to$ Drift I} & \textbf{$\to$ Drift II} \\
\midrule

\text{No Memory (Plain)} & 30 & 0 & 0 & 0 & 0 & 0 & 50 & 60 & 30 \\
\midrule
\text{Vanilla} & 80 & 0 & 0 & 80 & 0 & 0 & 95 & 95 & 95 \\
\rowcolor{gray!10}
\hspace{0.9em}\textbf{+GLOVE} & 100 & 60 \improv{60} & 95 \improv{95} & 80 & 75 \improv{75} & 50 \improv{50} & 85 & 95 & 95 \\
\addlinespace[2pt]
\text{MemoryBank} & 95 & 30 & 20 & 75 & 0 & 5 & 75 & 90 & 0 \\
\rowcolor{gray!10}
\hspace{0.9em}\textbf{+GLOVE} & 55 & 75 \improv{45} & 25 \improv{5} & 35 & 70 \improv{70} & 75 \improv{70} & 60 & 100 \improv{10} & 0 \\
\addlinespace[2pt]
\text{Voyager} & 75 & 0 & 0 & 10 & 0 & 0 & 95 & 90 & 95 \\
\rowcolor{gray!10}
\hspace{0.9em}\textbf{+GLOVE} & 85 & 75 \improv{75} & 90 \improv{90} & 65 & 0 & 75 \improv{75} & 95 & 85 \decline{5} & 90 \decline{5} \\
\addlinespace[2pt]
\text{Generative Agent} & 100 & 0 & 0 & 75 & 0 & 0 & 100 & 100 & 70 \\
\rowcolor{gray!10}
\hspace{0.9em}\textbf{+GLOVE} & 90 & 60 \improv{60} & 90 \improv{90} & 80 & 0 & 70 \improv{70} & 80 & 95 \decline{5} & 90 \improv{20} \\

\bottomrule
\end{tabular}%
}
\end{table}

\begin{table}[H]
\centering
\caption{\textbf{Qwen3-30B: GLOVE's robustness to environmental drift (Explicit).} Success rate in percent. The shaded rows show the performance of GLOVE-augmented agents. We annotate the performance gap relative to the baseline (e.g., \textcolor{cyan!60!green}{\scriptsize($\uparrow$56.3)} indicates improvement).}
\label{tab:plugplay_memdec_style_Qwen3_30B_explicit}
\footnotesize
\setlength{\tabcolsep}{3.5pt} 
\renewcommand{\arraystretch}{1.1}

\resizebox{1.0\textwidth}{!}{%
\begin{tabular}{l|ccc|ccc|ccc}
\toprule
\multicolumn{10}{c}{\textbf{\textsc{Verifier-visible structural drift}}} \\
\midrule
& \multicolumn{3}{c|}{\textbf{WebShop} (Semantic Shift)}
& \multicolumn{3}{c|}{\textbf{FrozenLake} (topology shift)}
& \multicolumn{3}{c}{\textbf{MountainCar} (dynamics shift)} \\
\cmidrule(lr){2-4}\cmidrule(lr){5-7}\cmidrule(lr){8-10}
\textbf{Method} & \textbf{Source} & \textbf{$\to$ Drift I} & \textbf{$\to$ Drift II}
& \textbf{Source} & \textbf{$\to$ Drift I} & \textbf{$\to$ Drift II}
& \textbf{Source} & \textbf{$\to$ Drift I} & \textbf{$\to$ Drift II} \\
\midrule

\text{No Memory (Plain)} & 0 & 30 & 0 & 100 & 0 & 0 & 100 & 100 & 100 \\
\midrule
\text{Vanilla} & 65 & 0 & 0 & 100 & 80 & 0 & 100 & 100 & 100 \\
\rowcolor{gray!10}
\hspace{0.9em}\textbf{+GLOVE} & 85 & 95 \improv{95} & 85 \improv{85} & 100 & 80 & 75 \improv{75} & 100 & 100 & 100 \\
\addlinespace[2pt]
\text{MemoryBank} & 95 & 35 & 30 & 100 & 80 & 55 & 100 & 100 & 100 \\
\rowcolor{gray!10}
\hspace{0.9em}\textbf{+GLOVE} & 95 & 95 \improv{60} & 80 \improv{50} & 100 & 80 & 75 \improv{20} & 100 & 100 & 100 \\
\addlinespace[2pt]
\text{Voyager} & 90 & 0 & 0 & 100 & 80 & 0 & 100 & 100 & 100 \\
\rowcolor{gray!10}
\hspace{0.9em}\textbf{+GLOVE} & 85 & 95 \improv{95} & 90 \improv{90} & 100 & 80 & 75 \improv{75} & 100 & 100 & 100 \\
\addlinespace[2pt]
\text{Generative Agent} & 75 & 0 & 0 & 100 & 80 & 0 & 100 & 100 & 100 \\
\rowcolor{gray!10}
\hspace{0.9em}\textbf{+GLOVE} & 95 & 95 \improv{95} & 90 \improv{90} & 100 & 80 & 75 \improv{75} & 100 & 100 & 100 \\

\bottomrule
\end{tabular}%
}
\end{table}

\begin{table}[H]
\centering
\caption{\textbf{Qwen2.5-7B: GLOVE's robustness to environmental drift (Explicit).} Success rate in percent. The shaded rows show the performance of GLOVE-augmented agents. We annotate the performance gap relative to the baseline (e.g., \textcolor{cyan!60!green}{\scriptsize($\uparrow$56.3)} indicates improvement).}
\label{tab:plugplay_memdec_style_Qwen2_5_7B_explicit}
\footnotesize
\setlength{\tabcolsep}{3.5pt} 
\renewcommand{\arraystretch}{1.1}

\resizebox{1.0\textwidth}{!}{%
\begin{tabular}{l|ccc|ccc|ccc}
\toprule
\multicolumn{10}{c}{\textbf{\textsc{Verifier-visible structural drift}}} \\
\midrule
& \multicolumn{3}{c|}{\textbf{WebShop} (Semantic Shift)}
& \multicolumn{3}{c|}{\textbf{FrozenLake} (topology shift)}
& \multicolumn{3}{c}{\textbf{MountainCar} (dynamics shift)} \\
\cmidrule(lr){2-4}\cmidrule(lr){5-7}\cmidrule(lr){8-10}
\textbf{Method} & \textbf{Source} & \textbf{$\to$ Drift I} & \textbf{$\to$ Drift II}
& \textbf{Source} & \textbf{$\to$ Drift I} & \textbf{$\to$ Drift II}
& \textbf{Source} & \textbf{$\to$ Drift I} & \textbf{$\to$ Drift II} \\
\midrule

\text{No Memory (Plain)} & 70 & 5 & 0 & 100 & 0 & 0 & 0 & 0 & 0 \\
\midrule
\text{Vanilla} & 90 & 0 & 0 & 100 & 0 & 0 & 0 & 0 & 0 \\
\rowcolor{gray!10}
\hspace{0.9em}\textbf{+GLOVE} & 95 & 75 \improv{75} & 75 \improv{75} & 100 & 80 \improv{80} & 75 \improv{75} & 0 & 0 & 0 \\
\addlinespace[2pt]
\text{MemoryBank} & 100 & 30 & 15 & 100 & 25 & 0 & 0 & 0 & 0 \\
\rowcolor{gray!10}
\hspace{0.9em}\textbf{+GLOVE} & 100 & 15 \decline{15} & 50 \improv{35} & 100 & 80 \improv{55} & 75 \improv{75} & 0 & 0 & 0 \\
\addlinespace[2pt]
\text{Voyager} & 100 & 0 & 0 & 100 & 0 & 0 & 0 & 0 & 0 \\
\rowcolor{gray!10}
\hspace{0.9em}\textbf{+GLOVE} & 100 & 85 \improv{85} & 85 \improv{85} & 100 & 75 \improv{75} & 50 \improv{50} & 0 & 0 & 0 \\
\addlinespace[2pt]
\text{Generative Agent} & 100 & 0 & 0 & 100 & 0 & 0 & 0 & 0 & 0 \\
\rowcolor{gray!10}
\hspace{0.9em}\textbf{+GLOVE} & 100 & 80 \improv{80} & 75 \improv{75} & 100 & 80 \improv{80} & 75 \improv{75} & 0 & 0 & 0 \\

\bottomrule
\end{tabular}%
}
\end{table}

\subsubsection{Direct Adaptive Baselines and Statistical Reliability}
\label{app:adaptive_statistical_analysis}

\paragraph{Adaptive baseline protocols.}
We evaluate six direct online update rules on Qwen3-30B FrozenLake
explicit drift.
\textsc{LatestOverwrite} retains only the latest observed outcome for each
$(s,a)$ pair;
\textsc{RecentK-K3} retains the three most recent transitions;
\textsc{SlidingWindow-W5} estimates the empirical outcome distribution
from the five most recent transitions;
\textsc{RecencyWeighted-H5} exponentially weights the pair-local history
with an observation half-life of five;
\textsc{PageHinkleyReset} applies a pair-specific Page--Hinkley change
detector and resets the corresponding local history after a detected
outcome change; and
\textsc{PeriodicRefresh-S30} clears the memory table every 30 natural
environment steps.
All methods use the same deterministic maps, prompt, action parser,
model settings, 20-step episode cap, and 20 Source episodes.

In the natural protocol, adaptive baselines update only from their own
trajectories, whereas GLOVE uses its standard verification procedure.
In the fixed-budget protocol, every method receives exactly $B=310$
post-drift environment interactions; GLOVE uses 280 task interactions
and 30 verification interactions, while each baseline uses all 310
interactions for task execution.
If the budget expires during an episode, its interactions and memory
updates consume budget, but the unfinished episode is not counted as
completed.

\paragraph{Paired multi-run analysis.}
To assess run-to-run variability, we additionally evaluate Vanilla,
MemoryBank, and their GLOVE-augmented counterparts on the same
Qwen3-30B FrozenLake explicit-drift setting over 10 complete paired runs,
with 20 evaluation episodes per phase.
For each run and backend, the baseline and its GLOVE variant are initialized
from byte-equivalent source-memory snapshots and evaluated on identical
task instances.
Because memory is updated sequentially within each run, we use paired
run-level differences as the inferential unit.
We report pooled success/failure counts, run-level mean and sample standard
deviation, 95\% $t$-intervals for paired percentage-point differences,
Bonferroni-adjusted exact sign tests over four planned comparisons,
and Cohen's $h$ for binary outcomes.

\begin{table}[H]
\centering
\caption{\textbf{Paired multi-run statistical analysis on Qwen3-30B
FrozenLake explicit drift.}
Each condition contains 10 complete runs with 20 episodes per phase.
``S/F'' denotes pooled successes/failures.
Mean $\pm$ SD is computed over run-level success rates.
Confidence intervals are 95\% $t$-intervals over paired
percentage-point differences.
$p$-values are Bonferroni-adjusted exact sign tests over four planned
comparisons; $h$ denotes Cohen's $h$.}
\label{tab:paired_statistical_analysis}
\footnotesize
\setlength{\tabcolsep}{3.0pt}
\renewcommand{\arraystretch}{1.12}

\resizebox{\textwidth}{!}{%
\begin{tabular}{llcccccc}
\toprule
\textbf{Backend}
& \textbf{Phase}
& \textbf{Baseline S/F}
& \textbf{Baseline Mean $\pm$ SD}
& \textbf{+GLOVE S/F}
& \textbf{+GLOVE Mean $\pm$ SD}
& \textbf{Gap [95\% CI]}
& \textbf{Adj. $p$ / $h$} \\
\midrule

Vanilla
& Drift I
& 160/40
& $80.0 \pm 0.0\%$
& 160/40
& $80.0 \pm 0.0\%$
& $0.0$ pp $[0.0,\,0.0]$
& $1.000$ / $0.000$ \\

Vanilla
& Drift II
& 0/200
& $0.0 \pm 0.0\%$
& 150/50
& $75.0 \pm 0.0\%$
& $+75.0$ pp $[75.0,\,75.0]$
& $0.0078$ / $2.094$ \\

MemoryBank
& Drift I
& 131/69
& $65.5 \pm 30.8\%$
& 160/40
& $80.0 \pm 0.0\%$
& $+14.5$ pp $[-7.5,\,36.5]$
& $1.000$ / $0.328$ \\

MemoryBank
& Drift II
& 37/163
& $18.5 \pm 7.5\%$
& 150/50
& $75.0 \pm 0.0\%$
& $+56.5$ pp $[51.2,\,61.8]$
& $0.0078$ / $1.205$ \\

\bottomrule
\end{tabular}%
}
\end{table}

Source success is 200/200 for every condition and is omitted from
Table~\ref{tab:paired_statistical_analysis}.
Both Drift-II gains are positive in all 10 paired runs and remain
significant after Bonferroni correction.
The MemoryBank Drift-I comparison contains two positive differences and
eight ties and is inconclusive.
Zero observed standard deviation means that the phase-level success counts
are identical across these runs, rather than that provider-side stochasticity
is absent.
For settings without repeated paired runs, we report descriptive results
without making statistical-significance claims.

\subsection{Experiment Results: Implicit Environmental Drifts}
\label{appendix-exp_result_implicit}

\begin{table}[H]
\centering
\caption{\textbf{GPT-4o: GLOVE's robustness to environmental drift (Implicit).} Score obtained. The shaded rows show the performance of GLOVE-augmented agents. We annotate the performance gap relative to the baseline (e.g., \textcolor{cyan!60!green}{\scriptsize($\uparrow$56.3)} indicates improvement).}
\label{tab:plugplay_memdec_style_GPT_4o_implicit}
\footnotesize
\setlength{\tabcolsep}{12.5pt} 
\renewcommand{\arraystretch}{1.1}

\begin{tabular}{l|ccccc|cccc}
\toprule
\multicolumn{10}{c}{\textbf{\textsc{Verifier-hidden drift}}} \\
\midrule
& \multicolumn{5}{c|}{\textbf{WebShop} (semantic shift)}
& \multicolumn{4}{c}{\textbf{FrozenLake} (reward reversal)} \\
\cmidrule(lr){2-6}\cmidrule(lr){7-10}
\textbf{Method}
& \multicolumn{2}{c}{\textbf{Source}} & \multicolumn{3}{c|}{$\to$ \textbf{Hidden drift}}
& \multicolumn{2}{c}{\textbf{Source}} & \multicolumn{2}{c}{$\to$ \textbf{Hidden drift}} \\
\midrule

\text{No Memory (Plain)}
& \multicolumn{2}{c}{82.5} & \multicolumn{3}{c|}{87.5} & \multicolumn{2}{c}{0} & \multicolumn{2}{c}{0} \\
\midrule
\text{Vanilla}
& \multicolumn{2}{c}{100} & \multicolumn{3}{c|}{75} & \multicolumn{2}{c}{47.5} & \multicolumn{2}{c}{50} \\
\rowcolor{gray!10}
\hspace{0.9em}\textbf{+GLOVE}
& \multicolumn{2}{c}{97.5} & \multicolumn{3}{c|}{93.8 \improv{18.8}} & \multicolumn{2}{c}{35} & \multicolumn{2}{c}{97.5 \improv{47.5}} \\
\addlinespace[2pt]
\text{MemoryBank}
& \multicolumn{2}{c}{100} & \multicolumn{3}{c|}{81.2} & \multicolumn{2}{c}{62.5} & \multicolumn{2}{c}{57.5} \\
\rowcolor{gray!10}
\hspace{0.9em}\textbf{+GLOVE}
& \multicolumn{2}{c}{98.8} & \multicolumn{3}{c|}{98.8 \improv{17.5}} & \multicolumn{2}{c}{67.5} & \multicolumn{2}{c}{97.5 \improv{40}} \\
\addlinespace[2pt]
\text{Voyager}
& \multicolumn{2}{c}{98.8} & \multicolumn{3}{c|}{75} & \multicolumn{2}{c}{67.5} & \multicolumn{2}{c}{50} \\
\rowcolor{gray!10}
\hspace{0.9em}\textbf{+GLOVE}
& \multicolumn{2}{c}{98.8} & \multicolumn{3}{c|}{98.8 \improv{23.8}} & \multicolumn{2}{c}{40} & \multicolumn{2}{c}{97.5 \improv{47.5}} \\
\addlinespace[2pt]
\text{Generative Agent}
& \multicolumn{2}{c}{98.8} & \multicolumn{3}{c|}{75} & \multicolumn{2}{c}{62.5} & \multicolumn{2}{c}{50} \\
\rowcolor{gray!10}
\hspace{0.9em}\textbf{+GLOVE}
& \multicolumn{2}{c}{98.8} & \multicolumn{3}{c|}{75} & \multicolumn{2}{c}{62.5} & \multicolumn{2}{c}{92.5 \improv{42.5}} \\

\bottomrule
\end{tabular}
\end{table}

\begin{table}[H]
\centering
\caption{\textbf{Grok-3: GLOVE's robustness to environmental drift (Implicit).} Score obtained. The shaded rows show the performance of GLOVE-augmented agents. We annotate the performance gap relative to the baseline (e.g., \textcolor{cyan!60!green}{\scriptsize($\uparrow$56.3)} indicates improvement).}
\label{tab:plugplay_memdec_style_Grok_3_implicit}
\footnotesize
\setlength{\tabcolsep}{12.5pt} 
\renewcommand{\arraystretch}{1.1}

\begin{tabular}{l|ccccc|cccc}
\toprule
\multicolumn{10}{c}{\textbf{\textsc{Verifier-hidden drift}}} \\
\midrule
& \multicolumn{5}{c|}{\textbf{WebShop} (semantic shift)}
& \multicolumn{4}{c}{\textbf{FrozenLake} (reward reversal)} \\
\cmidrule(lr){2-6}\cmidrule(lr){7-10}
\textbf{Method}
& \multicolumn{2}{c}{\textbf{Source}} & \multicolumn{3}{c|}{$\to$ \textbf{Hidden drift}}
& \multicolumn{2}{c}{\textbf{Source}} & \multicolumn{2}{c}{$\to$ \textbf{Hidden drift}} \\
\midrule

\text{No Memory (Plain)}
& \multicolumn{2}{c}{76.2} & \multicolumn{3}{c|}{96.2} & \multicolumn{2}{c}{0} & \multicolumn{2}{c}{0} \\
\midrule
\text{Vanilla}
& \multicolumn{2}{c}{96.2} & \multicolumn{3}{c|}{75} & \multicolumn{2}{c}{30} & \multicolumn{2}{c}{90} \\
\rowcolor{gray!10}
\hspace{0.9em}\textbf{+GLOVE}
& \multicolumn{2}{c}{97.5} & \multicolumn{3}{c|}{98.8 \improv{23.8}} & \multicolumn{2}{c}{30} & \multicolumn{2}{c}{100 \improv{10}} \\
\addlinespace[2pt]
\text{MemoryBank}
& \multicolumn{2}{c}{97.5} & \multicolumn{3}{c|}{81.2} & \multicolumn{2}{c}{15} & \multicolumn{2}{c}{82.5} \\
\rowcolor{gray!10}
\hspace{0.9em}\textbf{+GLOVE}
& \multicolumn{2}{c}{88.8} & \multicolumn{3}{c|}{98.8 \improv{17.5}} & \multicolumn{2}{c}{20} & \multicolumn{2}{c}{97.5 \improv{15}} \\
\addlinespace[2pt]
\text{Voyager}
& \multicolumn{2}{c}{98.8} & \multicolumn{3}{c|}{75} & \multicolumn{2}{c}{62.5} & \multicolumn{2}{c}{50} \\
\rowcolor{gray!10}
\hspace{0.9em}\textbf{+GLOVE}
& \multicolumn{2}{c}{100} & \multicolumn{3}{c|}{98.8 \improv{23.8}} & \multicolumn{2}{c}{27.5} & \multicolumn{2}{c}{100 \improv{50}} \\
\addlinespace[2pt]
\text{Generative Agent}
& \multicolumn{2}{c}{92.5} & \multicolumn{3}{c|}{82.5} & \multicolumn{2}{c}{35} & \multicolumn{2}{c}{50} \\
\rowcolor{gray!10}
\hspace{0.9em}\textbf{+GLOVE}
& \multicolumn{2}{c}{93.8} & \multicolumn{3}{c|}{97.5 \improv{15}} & \multicolumn{2}{c}{57.5} & \multicolumn{2}{c}{97.5 \improv{47.5}} \\

\bottomrule
\end{tabular}
\end{table}

\begin{table}[H]
\centering
\caption{\textbf{DeepSeek-V3.2: GLOVE's robustness to environmental drift (Implicit).} Score obtained. The shaded rows show the performance of GLOVE-augmented agents. We annotate the performance gap relative to the baseline (e.g., \textcolor{cyan!60!green}{\scriptsize($\uparrow$56.3)} indicates improvement).}
\label{tab:plugplay_memdec_style_DeepSeek_V3_2_implicit}
\footnotesize
\setlength{\tabcolsep}{12.5pt} 
\renewcommand{\arraystretch}{1.1}

\begin{tabular}{l|ccccc|cccc}
\toprule
\multicolumn{10}{c}{\textbf{\textsc{Verifier-hidden drift}}} \\
\midrule
& \multicolumn{5}{c|}{\textbf{WebShop} (semantic shift)}
& \multicolumn{4}{c}{\textbf{FrozenLake} (reward reversal)} \\
\cmidrule(lr){2-6}\cmidrule(lr){7-10}
\textbf{Method}
& \multicolumn{2}{c}{\textbf{Source}} & \multicolumn{3}{c|}{$\to$ \textbf{Hidden drift}}
& \multicolumn{2}{c}{\textbf{Source}} & \multicolumn{2}{c}{$\to$ \textbf{Hidden drift}} \\
\midrule

\text{No Memory (Plain)}
& \multicolumn{2}{c}{80} & \multicolumn{3}{c|}{95} & \multicolumn{2}{c}{0} & \multicolumn{2}{c}{0} \\
\midrule
Vanilla
& \multicolumn{2}{c}{93.8} & \multicolumn{3}{c|}{75} & \multicolumn{2}{c}{52.5} & \multicolumn{2}{c}{50} \\
\rowcolor{gray!10}
\hspace{0.9em}\textbf{+GLOVE}
& \multicolumn{2}{c}{86.2} & \multicolumn{3}{c|}{97.5 \improv{22.5}} & \multicolumn{2}{c}{50} & \multicolumn{2}{c}{92.5 \improv{42.5}} \\
\addlinespace[2pt]
MemoryBank
& \multicolumn{2}{c}{100} & \multicolumn{3}{c|}{82.5} & \multicolumn{2}{c}{10} & \multicolumn{2}{c}{82.5} \\
\rowcolor{gray!10}
\hspace{0.9em}\textbf{+GLOVE}
& \multicolumn{2}{c}{100} & \multicolumn{3}{c|}{97.5 \improv{15}} & \multicolumn{2}{c}{10} & \multicolumn{2}{c}{95 \improv{12.5}} \\
\addlinespace[2pt]
Voyager
& \multicolumn{2}{c}{90} & \multicolumn{3}{c|}{75} & \multicolumn{2}{c}{22.5} & \multicolumn{2}{c}{82.5} \\
\rowcolor{gray!10}
\hspace{0.9em}\textbf{+GLOVE}
& \multicolumn{2}{c}{93.8} & \multicolumn{3}{c|}{97.5 \improv{22.5}} & \multicolumn{2}{c}{30} & \multicolumn{2}{c}{95 \improv{12.5}} \\
\addlinespace[2pt]
Generative Agent
& \multicolumn{2}{c}{97.5} & \multicolumn{3}{c|}{75} & \multicolumn{2}{c}{22.5} & \multicolumn{2}{c}{87.5} \\
\rowcolor{gray!10}
\hspace{0.9em}\textbf{+GLOVE}
& \multicolumn{2}{c}{97.5} & \multicolumn{3}{c|}{97.5 \improv{22.5}} & \multicolumn{2}{c}{20} & \multicolumn{2}{c}{100 \improv{12.5}} \\

\bottomrule
\end{tabular}
\end{table}

\begin{table}[H]
\centering
\caption{\textbf{Llama3.3-70B: GLOVE's robustness to environmental drift (Implicit).} Score obtained. The shaded rows show the performance of GLOVE-augmented agents. We annotate the performance gap relative to the baseline (e.g., \textcolor{cyan!60!green}{\scriptsize($\uparrow$56.3)} indicates improvement).}
\label{tab:plugplay_memdec_style_Llama3_3_70B_implicit}
\footnotesize
\setlength{\tabcolsep}{12.5pt} 
\renewcommand{\arraystretch}{1.1}

\begin{tabular}{l|ccccc|cccc}
\toprule
\multicolumn{10}{c}{\textbf{\textsc{Verifier-hidden drift}}} \\
\midrule
& \multicolumn{5}{c|}{\textbf{WebShop} (semantic shift)}
& \multicolumn{4}{c}{\textbf{FrozenLake} (reward reversal)} \\
\cmidrule(lr){2-6}\cmidrule(lr){7-10}
\textbf{Method}
& \multicolumn{2}{c}{\textbf{Source}} & \multicolumn{3}{c|}{$\to$ \textbf{Hidden drift}}
& \multicolumn{2}{c}{\textbf{Source}} & \multicolumn{2}{c}{$\to$ \textbf{Hidden drift}} \\
\midrule

\text{No Memory (Plain)}
& \multicolumn{2}{c}{91.2} & \multicolumn{3}{c|}{80} & \multicolumn{2}{c}{0} & \multicolumn{2}{c}{0} \\
\midrule
\text{Vanilla}
& \multicolumn{2}{c}{86.2} & \multicolumn{3}{c|}{75} & \multicolumn{2}{c}{42.5} & \multicolumn{2}{c}{95} \\
\rowcolor{gray!10}
\hspace{0.9em}\textbf{+GLOVE}
& \multicolumn{2}{c}{100} & \multicolumn{3}{c|}{75} & \multicolumn{2}{c}{42.5} & \multicolumn{2}{c}{100 \improv{5}} \\
\addlinespace[2pt]
\text{MemoryBank}
& \multicolumn{2}{c}{98.8} & \multicolumn{3}{c|}{81.2} & \multicolumn{2}{c}{35} & \multicolumn{2}{c}{90} \\
\rowcolor{gray!10}
\hspace{0.9em}\textbf{+GLOVE}
& \multicolumn{2}{c}{100} & \multicolumn{3}{c|}{88.8 \improv{7.5}} & \multicolumn{2}{c}{35} & \multicolumn{2}{c}{90} \\
\addlinespace[2pt]
\text{Voyager}
& \multicolumn{2}{c}{96.2} & \multicolumn{3}{c|}{75} & \multicolumn{2}{c}{40} & \multicolumn{2}{c}{100} \\
\rowcolor{gray!10}
\hspace{0.9em}\textbf{+GLOVE}
& \multicolumn{2}{c}{100} & \multicolumn{3}{c|}{95 \improv{20}} & \multicolumn{2}{c}{40} & \multicolumn{2}{c}{100} \\
\addlinespace[2pt]
\text{Generative Agent}
& \multicolumn{2}{c}{100} & \multicolumn{3}{c|}{75} & \multicolumn{2}{c}{40} & \multicolumn{2}{c}{100} \\
\rowcolor{gray!10}
\hspace{0.9em}\textbf{+GLOVE}
& \multicolumn{2}{c}{100} & \multicolumn{3}{c|}{75} & \multicolumn{2}{c}{42.5} & \multicolumn{2}{c}{100} \\

\bottomrule
\end{tabular}
\end{table}

\begin{table}[H]
\centering
\caption{\textbf{Llama3.1-8B: GLOVE's robustness to environmental drift (Implicit).} Score obtained. The shaded rows show the performance of GLOVE-augmented agents. We annotate the performance gap relative to the baseline (e.g., \textcolor{cyan!60!green}{\scriptsize($\uparrow$56.3)} indicates improvement).}
\label{tab:plugplay_memdec_style_Llama3_1_8B_implicit}
\footnotesize
\setlength{\tabcolsep}{12.5pt} 
\renewcommand{\arraystretch}{1.1}

\begin{tabular}{l|ccccc|cccc}
\toprule
\multicolumn{10}{c}{\textbf{\textsc{Verifier-hidden drift}}} \\
\midrule
& \multicolumn{5}{c|}{\textbf{WebShop} (semantic shift)}
& \multicolumn{4}{c}{\textbf{FrozenLake} (reward reversal)} \\
\cmidrule(lr){2-6}\cmidrule(lr){7-10}
\textbf{Method}
& \multicolumn{2}{c}{\textbf{Source}} & \multicolumn{3}{c|}{$\to$ \textbf{Hidden drift}}
& \multicolumn{2}{c}{\textbf{Source}} & \multicolumn{2}{c}{$\to$ \textbf{Hidden drift}} \\
\midrule

\text{No Memory (Plain)}
& \multicolumn{2}{c}{47.5} & \multicolumn{3}{c|}{60} & \multicolumn{2}{c}{0} & \multicolumn{2}{c}{0} \\
\midrule
\text{Vanilla}
& \multicolumn{2}{c}{30} & \multicolumn{3}{c|}{36.2} & \multicolumn{2}{c}{10} & \multicolumn{2}{c}{50} \\
\rowcolor{gray!10}
\hspace{0.9em}\textbf{+GLOVE}
& \multicolumn{2}{c}{42.5} & \multicolumn{3}{c|}{76.2 \improv{40}} & \multicolumn{2}{c}{12.5} & \multicolumn{2}{c}{95 \improv{45}} \\
\addlinespace[2pt]
\text{MemoryBank}
& \multicolumn{2}{c}{26.2} & \multicolumn{3}{c|}{27.5} & \multicolumn{2}{c}{7.5} & \multicolumn{2}{c}{65} \\
\rowcolor{gray!10}
\hspace{0.9em}\textbf{+GLOVE}
& \multicolumn{2}{c}{50} & \multicolumn{3}{c|}{67.5 \improv{40}} & \multicolumn{2}{c}{5} & \multicolumn{2}{c}{65} \\
\addlinespace[2pt]
\text{Voyager}
& \multicolumn{2}{c}{32.5} & \multicolumn{3}{c|}{26.2} & \multicolumn{2}{c}{20} & \multicolumn{2}{c}{35} \\
\rowcolor{gray!10}
\hspace{0.9em}\textbf{+GLOVE}
& \multicolumn{2}{c}{15} & \multicolumn{3}{c|}{42.5 \improv{16.2}} & \multicolumn{2}{c}{15} & \multicolumn{2}{c}{100 \improv{65}} \\
\addlinespace[2pt]
\text{Generative Agent}
& \multicolumn{2}{c}{48.8} & \multicolumn{3}{c|}{55} & \multicolumn{2}{c}{15} & \multicolumn{2}{c}{60} \\
\rowcolor{gray!10}
\hspace{0.9em}\textbf{+GLOVE}
& \multicolumn{2}{c}{62.5} & \multicolumn{3}{c|}{60 \improv{5}} & \multicolumn{2}{c}{17.5} & \multicolumn{2}{c}{90 \improv{30}} \\

\bottomrule
\end{tabular}
\end{table}

\begin{table}[H]
\centering
\caption{\textbf{Qwen3-30B: GLOVE's robustness to environmental drift (Implicit).} Score obtained. The shaded rows show the performance of GLOVE-augmented agents. We annotate the performance gap relative to the baseline (e.g., \textcolor{cyan!60!green}{\scriptsize($\uparrow$56.3)} indicates improvement).}
\label{tab:plugplay_memdec_style_Qwen3_30B_implicit}
\footnotesize
\setlength{\tabcolsep}{12.5pt} 
\renewcommand{\arraystretch}{1.1}

\begin{tabular}{l|ccccc|cccc}
\toprule
\multicolumn{10}{c}{\textbf{\textsc{Verifier-hidden drift}}} \\
\midrule
& \multicolumn{5}{c|}{\textbf{WebShop} (semantic shift)}
& \multicolumn{4}{c}{\textbf{FrozenLake} (reward reversal)} \\
\cmidrule(lr){2-6}\cmidrule(lr){7-10}
\textbf{Method}
& \multicolumn{2}{c}{\textbf{Source}} & \multicolumn{3}{c|}{$\to$ \textbf{Hidden drift}}
& \multicolumn{2}{c}{\textbf{Source}} & \multicolumn{2}{c}{$\to$ \textbf{Hidden drift}} \\
\midrule

\text{No Memory (Plain)}
& \multicolumn{2}{c}{75} & \multicolumn{3}{c|}{100} & \multicolumn{2}{c}{0} & \multicolumn{2}{c}{0} \\
\midrule
\text{Vanilla}
& \multicolumn{2}{c}{75} & \multicolumn{3}{c|}{100} & \multicolumn{2}{c}{90} & \multicolumn{2}{c}{50} \\
\rowcolor{gray!10}
\hspace{0.9em}\textbf{+GLOVE}
& \multicolumn{2}{c}{75} & \multicolumn{3}{c|}{100} & \multicolumn{2}{c}{85} & \multicolumn{2}{c}{50} \\
\addlinespace[2pt]
\text{MemoryBank}
& \multicolumn{2}{c}{75} & \multicolumn{3}{c|}{100} & \multicolumn{2}{c}{90} & \multicolumn{2}{c}{37.5} \\
\rowcolor{gray!10}
\hspace{0.9em}\textbf{+GLOVE}
& \multicolumn{2}{c}{75} & \multicolumn{3}{c|}{100} & \multicolumn{2}{c}{85} & \multicolumn{2}{c}{40 \improv{2.5}} \\
\addlinespace[2pt]
\text{Voyager}
& \multicolumn{2}{c}{75} & \multicolumn{3}{c|}{100} & \multicolumn{2}{c}{85} & \multicolumn{2}{c}{50} \\
\rowcolor{gray!10}
\hspace{0.9em}\textbf{+GLOVE}
& \multicolumn{2}{c}{75} & \multicolumn{3}{c|}{100} & \multicolumn{2}{c}{85} & \multicolumn{2}{c}{50} \\
\addlinespace[2pt]
\text{Generative Agent}
& \multicolumn{2}{c}{75} & \multicolumn{3}{c|}{100} & \multicolumn{2}{c}{90} & \multicolumn{2}{c}{50} \\
\rowcolor{gray!10}
\hspace{0.9em}\textbf{+GLOVE}
& \multicolumn{2}{c}{75} & \multicolumn{3}{c|}{100} & \multicolumn{2}{c}{85} & \multicolumn{2}{c}{50} \\

\bottomrule
\end{tabular}
\end{table}

\begin{table}[H]
\centering
\caption{\textbf{Qwen2.5-7B: GLOVE's robustness to environmental drift (Implicit).} Score obtained. The shaded rows show the performance of GLOVE-augmented agents. We annotate the performance gap relative to the baseline (e.g., \textcolor{cyan!60!green}{\scriptsize($\uparrow$56.3)} indicates improvement).}
\label{tab:plugplay_memdec_style_Qwen2_5_7B_implicit}
\footnotesize
\setlength{\tabcolsep}{12.5pt} 
\renewcommand{\arraystretch}{1.1}

\begin{tabular}{l|ccccc|cccc}
\toprule
\multicolumn{10}{c}{\textbf{\textsc{Verifier-hidden drift}}} \\
\midrule
& \multicolumn{5}{c|}{\textbf{WebShop} (semantic shift)}
& \multicolumn{4}{c}{\textbf{FrozenLake} (reward reversal)} \\
\cmidrule(lr){2-6}\cmidrule(lr){7-10}
\textbf{Method}
& \multicolumn{2}{c}{\textbf{Source}} & \multicolumn{3}{c|}{$\to$ \textbf{Hidden drift}}
& \multicolumn{2}{c}{\textbf{Source}} & \multicolumn{2}{c}{$\to$ \textbf{Hidden drift}} \\
\midrule

\text{No Memory (Plain)}
& \multicolumn{2}{c}{0} & \multicolumn{3}{c|}{15} & \multicolumn{2}{c}{0} & \multicolumn{2}{c}{0} \\
\midrule
\text{Vanilla}
& \multicolumn{2}{c}{30} & \multicolumn{3}{c|}{75} & \multicolumn{2}{c}{42.5} & \multicolumn{2}{c}{100} \\
\rowcolor{gray!10}
\hspace{0.9em}\textbf{+GLOVE}
& \multicolumn{2}{c}{47.5} & \multicolumn{3}{c|}{75} & \multicolumn{2}{c}{42.5} & \multicolumn{2}{c}{100} \\
\addlinespace[2pt]
\text{MemoryBank}
& \multicolumn{2}{c}{17.5} & \multicolumn{3}{c|}{53.8} & \multicolumn{2}{c}{35} & \multicolumn{2}{c}{95} \\
\rowcolor{gray!10}
\hspace{0.9em}\textbf{+GLOVE}
& \multicolumn{2}{c}{45} & \multicolumn{3}{c|}{75 \improv{21.3}} & \multicolumn{2}{c}{37.5} & \multicolumn{2}{c}{100 \improv{5}} \\
\addlinespace[2pt]
\text{Voyager}
& \multicolumn{2}{c}{50} & \multicolumn{3}{c|}{75} & \multicolumn{2}{c}{42.5} & \multicolumn{2}{c}{95} \\
\rowcolor{gray!10}
\hspace{0.9em}\textbf{+GLOVE}
& \multicolumn{2}{c}{47.5} & \multicolumn{3}{c|}{75} & \multicolumn{2}{c}{45} & \multicolumn{2}{c}{100 \improv{5}} \\
\addlinespace[2pt]
\text{Generative Agent}
& \multicolumn{2}{c}{45} & \multicolumn{3}{c|}{75} & \multicolumn{2}{c}{42.5} & \multicolumn{2}{c}{100} \\
\rowcolor{gray!10}
\hspace{0.9em}\textbf{+GLOVE}
& \multicolumn{2}{c}{47.5} & \multicolumn{3}{c|}{75} & \multicolumn{2}{c}{42.5} & \multicolumn{2}{c}{100} \\

\bottomrule
\end{tabular}
\end{table}

\subsubsection{Full-Trajectory Context Ablation}
\label{app:fulltraj_ablation}

We further test whether the failure of static memory under explicit drift is mainly caused by the narrowness of local state-matched retrieval. We consider two broader-context variants in WebShop. First, \textsc{FullTraj-Static} provides a complete successful source-environment trajectory at every decision step during drifted evaluation. Second, we include a trajectory-aware Voyager-style memory that uses both trajectory summaries and full-trajectory context, representing a richer episodic/procedural memory backend.

\begin{table}[H]
\centering
\caption{\textbf{Full-trajectory context ablation on WebShop explicit drift.} Success rate in percent over 20 episodes. \textsc{FullTraj-Static} provides a complete successful source-environment trajectory at each decision step. The Voyager-style variant uses both a full trajectory and a trajectory summary, representing a broader trajectory-aware memory backend.}
\label{tab:fulltraj_ablation_webshop}
\footnotesize
\setlength{\tabcolsep}{6pt}
\renewcommand{\arraystretch}{1.1}
\begin{tabular}{lccc}
\toprule
\textbf{Method} & \textbf{Source} & \textbf{$\rightarrow$ Drift I} & \textbf{$\rightarrow$ Drift II} \\
\midrule
Static Memory (StateMatched) & 85 & 0 & 0 \\
Static Memory (\textsc{FullTraj-Static}) & 95 & 0 & 0 \\
\rowcolor{gray!10}
StateMatched + GLOVE (Vanilla) & 85 & 85 & 95 \\
\rowcolor{gray!10}
Trajectory Memory + GLOVE (Voyager) & 80 & 90 & 95 \\
\bottomrule
\end{tabular}
\end{table}
Table~\ref{tab:fulltraj_ablation_webshop} shows that broader trajectory context alone does not remove stale-memory failure. \textsc{FullTraj-Static} improves source-environment success from $85\%$ to $95\%$, but still collapses to $0\%$ after both drifts. By contrast, once GLOVE is added, both backends remain robust after drift: state-matched Vanilla + GLOVE reaches $85\%$ and $95\%$, while Voyager + GLOVE reaches $90\%$ and $95\%$ on Drift I and Drift II. These results suggest that broader episodic context can improve performance in the stationary source environment, but it does not by itself enable the agent to discount stale source-environment experience after drift. The main benefit under drift comes from explicit memory verification and realignment.

\subsection{Positioning with Respect to Adaptive Model-Based RL}
\label{app:dynaq_positioning}

GLOVE share similarities with adaptive model-based RL as both use fresh interaction to correct beliefs about environment behavior. This appendix is included to clarify the boundary of this connection, rather than to treat RL algorithms as the primary baselines. Classical adaptive model-based RL typically learns or updates a reward model and uses it for reward-conditioned planning or policy improvement. GLOVE instead operates as a deployment-time memory-validation layer for memory-augmented LLM agents: it asks whether a retrieved memory entry, which may be a textual observation, a procedural trace, a trajectory summary, or a semantic association, should still be trusted before it is reused as prompt context.

This distinction is important in text-rich agent tasks. In WebShop, for example, the effective decision context includes natural-language instructions, noisy HTML, URLs, product attributes, and page affordances. A drift may be semantic, such as a changed grounding of a user preference, even when the surface interface remains similar. A tabular or low-dimensional model-based RL baseline would first require a stable Markov abstraction and a reward model over this textual state/action space. Learned latent world models are possible in principle, but then the comparison would conflate memory validation with representation learning, semantic grounding, and policy optimization. 
GLOVE addresses a narrower deployment-time problem in which, given the memory abstraction already consumed by the LLM agent, the system locally falsifies stale memory and realigns it with current observations or feedback.

Where adaptive model-based RL is naturally applicable, we include a small FrozenLake positioning check with Dyna-Q and Dyna-Q+~\cite{sutton1990integrated, sutton1991reinforcement}. As shown in Table~\ref{tab:dynaq}, these baselines adapt well under the smallest map drift, but degrade substantially under the larger structured drift. In the same large-drift setting, GLOVE preserves its pre-drift success rate. This comparison should be interpreted as a scope check rather than a fully symmetric RL benchmark that GLOVE is not a replacement for model-based RL, and adaptive RL is not the main target class of this paper. The intended takeaway is that stale decision memory is a distinct failure mode in LLM agents, especially when memory is textual or procedural, and that local memory verification can be complementary to policy learning or planning methods.

\begin{figure}[H]
    \centering
    \begin{subfigure}[t]{0.22\linewidth}
        \centering
        \includegraphics[width=\linewidth]{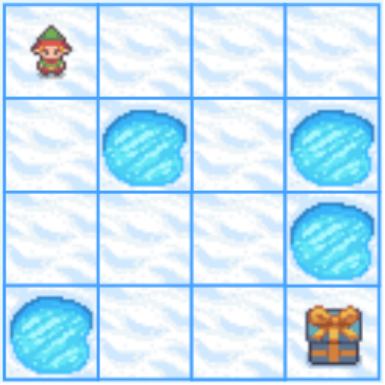}
        \caption{Small-drift setting: source map.}
    \end{subfigure}
    \hfill
    \begin{subfigure}[t]{0.22\linewidth}
        \centering
        \includegraphics[width=\linewidth]{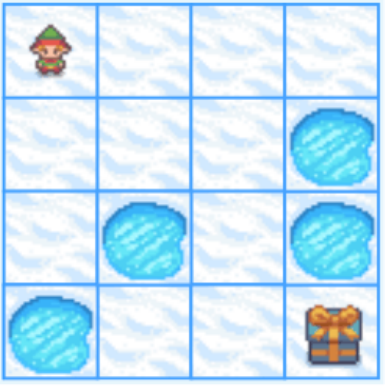}
        \caption{Small-drift setting: drifted map.}
    \end{subfigure}
    \hfill
    \begin{subfigure}[t]{0.22\linewidth}
        \centering
        \includegraphics[width=\linewidth]{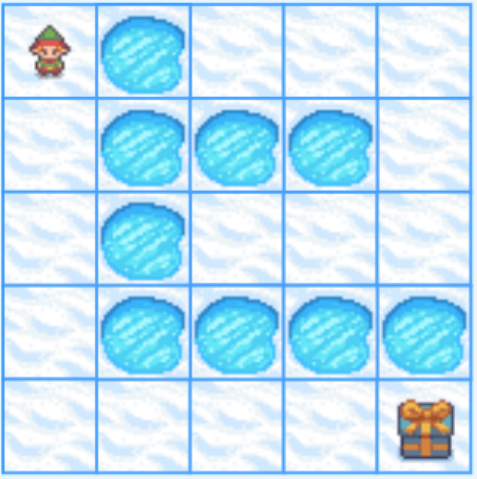}
        \caption{Large-drift setting: source map.}
    \end{subfigure}
    \hfill
    \begin{subfigure}[t]{0.22\linewidth}
        \centering
        \includegraphics[width=\linewidth]{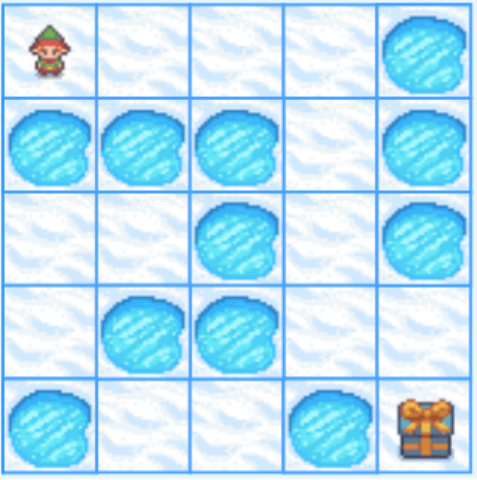}
        \caption{Large-drift setting: drifted map.}
    \end{subfigure}
    \caption{FrozenLake maps used for the experiments in Table~\ref{tab:dynaq}.}
    \label{fig:dynaq_maps}
\end{figure}

\begin{table}[h]
\centering
\small
\caption{Positioning check against classical adaptive model-based RL in FrozenLake.}
\label{tab:dynaq}
\begin{tabular}{lccc}
\toprule
Setting & Method & Success before $\uparrow$ & Success after $\uparrow$ \\
\midrule
Small drift & Dyna-Q & $0.74 \pm 0.15$ & $0.90 \pm 0.01$ \\
Small drift & Dyna-Q+ & $0.69 \pm 0.17$ & $0.80 \pm 0.11$ \\
Large drift & Dyna-Q & $0.20 \pm 0.24$ & $0.27 \pm 0.26$ \\
Large drift & Dyna-Q+ & $0.52 \pm 0.22$ & $0.21 \pm 0.22$ \\
Large drift & GLOVE & $0.80$ & $0.80$ \\
\bottomrule
\end{tabular}
\end{table}

\subsection{Cost Analysis}
\label{app:cost_analysis}

Active verification introduces additional environment interactions, so its benefit should be evaluated together with its cost. We analyze this cost from three perspectives: when verification is triggered, how the probing budget affects robustness, and how matching granularity changes the number of verification steps.

\begin{figure}[H]
    \centering
    \includegraphics[width=0.6\linewidth]{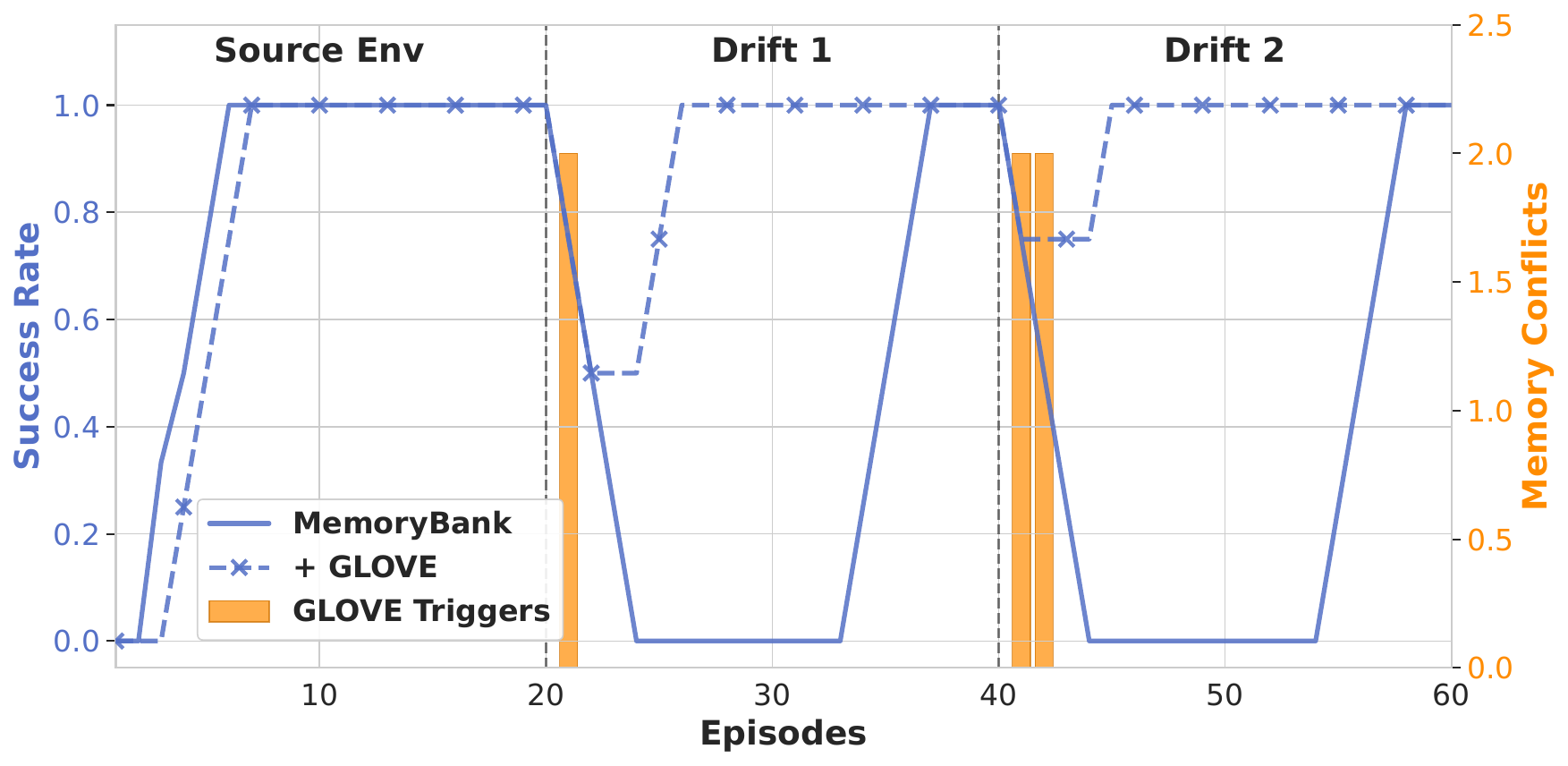}
    \caption{Verification dynamics under explicit drift in WebShop. GLOVE triggers verification sparsely, with conflict spikes concentrated around drift events, enabling rapid recovery without continuous probing.}
    \label{fig:adaptation_curve}
\end{figure}
As shown in Fig.~\ref{fig:adaptation_curve}, GLOVE does not increase interaction uniformly throughout deployment. Instead, verification is triggered sparsely and concentrates around drift events, where memory conflicts spike. This allows GLOVE to recover quickly after environmental change while maintaining low overhead during stable periods.

\begin{table}[h]
\centering
\small
\caption{Cost-normalized comparison on WebShop explicit drift. Total interactions include both task-execution and verification steps. Active-probe interactions count only direct GLOVE verification probes.}
\label{tab:cost_normalized}
\begin{tabular}{lccc}
\toprule
Method & Phase & Total env. interactions & Active-probe interactions \\
\midrule
MemoryBank & Source & 116 & 0 \\
MemoryBank & Drift I & 128 & 0 \\
MemoryBank & Drift II & 136 & 0 \\
MemoryBank + GLOVE & Source & 118 & 0 \\
MemoryBank + GLOVE & Drift I & 254 & 86 \\
MemoryBank + GLOVE & Drift II & 419 & 151 \\
\bottomrule
\end{tabular}
\end{table}
Table~\ref{tab:cost_normalized} provides a cost-normalized view of GLOVE's verification overhead. Total environment interactions include both normal task-execution steps and verification steps, while active-probe interactions count only the additional steps used for direct GLOVE probing. In the source environment, GLOVE adds almost no overhead, increasing total interactions from 116 to 118 with zero active probes. After drift, the additional cost is concentrated around realignment: total interactions increase from 128/136 to 254/419 in Drift I/II, with 86/151 active-probe interactions. This indicates that GLOVE's benefit does not come from uniformly using more interactions, but from selectively spending verification budget when memory-environment conflicts arise.

\begin{figure}[H]
    \centering
    \includegraphics[width=0.6\linewidth]{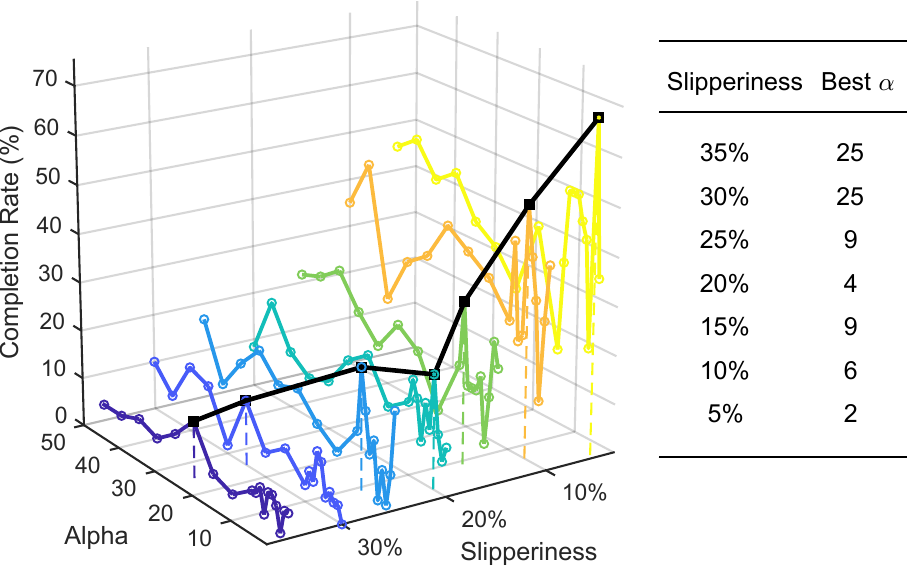}
    \caption{Effect of probing budget $\alpha$ in FrozenLake. Larger probing budgets improve robustness under higher stochasticity, while smaller budgets are sufficient in more stable environments.}
    \label{fig:probing_budget}
\end{figure}
Fig.~\ref{fig:probing_budget} shows that the probing budget $\alpha$ acts as a cost--reliability control parameter. Larger $\alpha$ improves robustness under more stochastic dynamics by providing more reliable transition estimates, while smaller budgets are sufficient in more stable settings. Thus, $\alpha$ can be adjusted according to the expected noise level of the environment.

\begin{table}[H]
\centering
\small
\caption{WebShop trigger-sensitivity study. $\eta$ controls how much attribute information is retained in the confirmation-state representation. Moderate matching granularity saturates performance while keeping verification cost low.}
\begin{tabular}{cccc}
\toprule
$\eta$ & Avg.~Reward & Avg.~Steps / Episode & Avg.~Verify Env Steps / Episode \\
\midrule
0.0 & 0.408 & 10.93 & 38.83 \\
0.5 & 0.683 & 8.85 & 1.48 \\
1.0 & 0.683 & 8.67 & 1.40 \\
\bottomrule
\end{tabular}
\label{tab:webshop_eta}
\end{table}
Finally, Table~\ref{tab:webshop_eta} shows that verification cost also depends on the granularity of counterpart matching. Overly coarse matching ($\eta=0.0$) triggers excessive verification, leading to 38.83 verification steps per episode. In contrast, moderate and strict matching ($\eta=0.5$ and $\eta=1.0$) attain the same average reward while keeping verification overhead comparably low, at 1.48 and 1.40 verification steps per episode respectively. Overall, these results show that GLOVE's overhead is selective and controllable rather than continuously incurred.

\subsection{JetRover Real-World Robotic Result}
\label{app:jetrover_result}
Figure~\ref{fig:jetrover_adaptation} reports the real-robot evaluation on JetRover. The robot searches for the \texttt{blue} block in a cardboard maze. After source-memory formation, we induce semantic drift by swapping the \texttt{blue} and \texttt{green} blocks. Static Memory keeps visiting the stale target location after the swap, while GLOVE detects the semantic mismatch, updates memory, and recovers correct search behavior under physical nuisance variation.

\subsection{Result Analysis}
\label{app:result_analysis}
While GLOVE yields substantial improvements in most settings, it does not uniformly dominate across all configurations. In particular, in {\textbf{FrozenLake Drift I}} (e.g., Table~\ref{tab:plugplay_memdec_style_Grok_3_explicit}), GLOVE occasionally underperforms the baseline. The reason is that the induced drift in this setting is relatively mild, such that high-capacity models can already maintain near-optimal performance without explicit verification. In these cases, additional probing may introduce unnecessary perturbations, whereas under more severe drift (e.g., {\textbf{Drift II}}), GLOVE consistently outperforms the baseline by actively correcting stale transitions.

Another source of performance degradation is the use of small backbone models. As shown in Tables~\ref{tab:plugplay_memdec_style_Llama3_1_8B_explicit} and~\ref{tab:plugplay_memdec_style_Qwen2_5_7B_explicit}, models such as \textbf{Llama3.1-8B} and \textbf{Qwen2.5-7B} exhibit limited ability to interpret and exploit the verified transition summaries produced by GLOVE. This suggests that effective utilization of corrected memory requires a certain level of instruction-following and reasoning capacity. When this condition is violated, verification may fail to translate into improved action selection.

These observations highlight that GLOVE is most beneficial when environmental drift is substantial and when the backbone model can reliably incorporate corrective feedback. They also indicate a cost-benefit trade-off. In low-drift regimes, the marginal gains of verification may not justify additional interaction. Nevertheless, across the full benchmark tests, GLOVE remains a positive augmentation, particularly in regimes where stale memory would otherwise lead to catastrophic failure.

\newpage
\section*{NeurIPS Paper Checklist}
\begin{enumerate}

\item {\bf Claims}
    \item[] Question: Do the main claims made in the abstract and introduction accurately reflect the paper's contributions and scope?
    \item[] Answer: \answerYes{}{} 
    \item[] Justification: The abstract and introduction reflect the paper's contributions and scope.

\item {\bf Limitations}
    \item[] Question: Does the paper discuss the limitations of the work performed by the authors?
    \item[] Answer: \answerYes{} 
    \item[] Justification: We discuss the limitations in Appendix~\ref{limitation}.

\item {\bf Theory assumptions and proofs}
    \item[] Question: For each theoretical result, does the paper provide the full set of assumptions and a complete (and correct) proof?
    \item[] Answer: \answerYes{}{} 
    \item[] Justification: Assumptions and proof are provided in the paper.

    \item {\bf Experimental result reproducibility}
    \item[] Question: Does the paper fully disclose all the information needed to reproduce the main experimental results of the paper to the extent that it affects the main claims and/or conclusions of the paper (regardless of whether the code and data are provided or not)?
    \item[] Answer: \answerYes{}{} 
    \item[] Justification: Source code and information needed to reproduce the main results are provided.

\item {\bf Open access to data and code}
    \item[] Question: Does the paper provide open access to the data and code, with sufficient instructions to faithfully reproduce the main experimental results, as described in supplemental material?
    \item[] Answer: \answerYes{}{} 
    \item[] Justification: We release the GLOVE implementation, the four memory backends (vanilla, MemoryBank, Voyager, and Generative Agents), and the augmented evaluation environments (the modified WebShop purchase flow, the FrozenLake map variants, and the MountainCar dynamics variants), together with installation and run instructions, at \url{https://github.com/NICE-HKU/GLOVE}. These reproduce the main memory-realignment experiments in Tables~\ref{table-gpt4o_explicit} and~\ref{table-gpt4o_implicit}. The rule-based adaptive memory baselines and the cost and paired-run analyses added during revision are not released as separate scripts, as they are short variants of the released memory-update interface that are fully specified in Appendix~\ref{app:adaptive_statistical_analysis} and Appendix~\ref{app:cost_analysis}. The real-robot results require the physical JetRover platform and cannot be reproduced in simulation.

\item {\bf Experimental setting/details}
    \item[] Question: Does the paper specify all the training and test details (e.g., data splits, hyperparameters, how they were chosen, type of optimizer) necessary to understand the results?
    \item[] Answer: \answerYes{} 
    \item[] Justification: We cover related details in the Appendix~\ref{appendix-experiment_results}.

\item {\bf Experiment statistical significance}
    \item[] Question: Does the paper report error bars suitably and correctly defined or other appropriate information about the statistical significance of the experiments?
    \item[] Answer: \answerYes{} 
    \item[] Justification:
We report a 10-run paired analysis (run-level mean $\pm$ sample SD, 95\% CI,
Bonferroni-adjusted exact sign tests, and Cohen's $h$) for the Qwen3-30B
FrozenLake explicit-drift setting in
Appendix~\ref{app:adaptive_statistical_analysis}. Settings without repeated
paired runs are reported descriptively, without significance claims.

\item {\bf Experiments compute resources}
    \item[] Question: For each experiment, does the paper provide sufficient information on the computer resources (type of compute workers, memory, time of execution) needed to reproduce the experiments?
    \item[] Answer: \answerYes{} 
    \item[] Justification:
All experiments are inference-only: no model is trained or fine-tuned, and no
local GPU is used. Every LLM backbone is queried through the provider's cloud
inference API; the memory backends and environments run on a single CPU machine
with negligible storage, and API inference dominates wall-clock time and cost.
The JetRover experiment additionally uses the robot's onboard compute; no
unreported runs required further compute.
    
\item {\bf Code of ethics}
    \item[] Question: Does the research conducted in the paper conform, in every respect, with the NeurIPS Code of Ethics \url{https://neurips.cc/public/EthicsGuidelines}?
    \item[] Answer: \answerYes{} 
    \item[] Justification: The research respects the NeurIPS Code of Ethics in every aspect.

\item {\bf Broader impacts}
    \item[] Question: Does the paper discuss both potential positive societal impacts and negative societal impacts of the work performed?
    \item[] Answer: \answerYes{} 
    \item[] Justification: We discuss the broader impacts of this work in Appendix~\ref{broader_impact}.
    
\item {\bf Safeguards}
    \item[] Question: Does the paper describe safeguards that have been put in place for responsible release of data or models that have a high risk for misuse (e.g., pre-trained language models, image generators, or scraped datasets)?
    \item[] Answer: \answerNA{}{} 
    \item[] Justification: Our works provide a plug-and-play memory-augment module that does not have related risks.

\item {\bf Licenses for existing assets}
    \item[] Question: Are the creators or original owners of assets (e.g., code, data, models), used in the paper, properly credited and are the license and terms of use explicitly mentioned and properly respected?
    \item[] Answer: \answerYes{}{} 
    \item[] Justification: Their contribution has already been properly acknowledged and credited.

\item {\bf New assets}
    \item[] Question: Are new assets introduced in the paper well documented and is the documentation provided alongside the assets?
    \item[] Answer: \answerNA{}{} 
    \item[] Justification: The paper does not introduce new assets other than source code of implementation.

\item {\bf Crowdsourcing and research with human subjects}
    \item[] Question: For crowdsourcing experiments and research with human subjects, does the paper include the full text of instructions given to participants and screenshots, if applicable, as well as details about compensation (if any)? 
    \item[] Answer: \answerNA{} 
    \item[] Justification: The paper does not include crowdsourcing and research with human subjects.

\item {\bf Institutional review board (IRB) approvals or equivalent for research with human subjects}
    \item[] Question: Does the paper describe potential risks incurred by study participants, whether such risks were disclosed to the subjects, and whether Institutional Review Board (IRB) approvals (or an equivalent approval/review based on the requirements of your country or institution) were obtained?
    \item[] Answer: \answerNA{} 
    \item[] Justification: N/A

\item {\bf Declaration of LLM usage}
    \item[] Question: Does the paper describe the usage of LLMs if it is an important, original, or non-standard component of the core methods in this research? Note that if the LLM is used only for writing, editing, or formatting purposes and does \emph{not} impact the core methodology, scientific rigor, or originality of the research, declaration is not required.
    \item[] Answer: \answerYes{} 
    \item[] Justification: LLMs are used as one of the main components of the proposed LLM memory system module.

\end{enumerate}

\end{document}